\documentclass{article}

\usepackage{microtype}
\usepackage{graphicx}
\usepackage{subfigure}
\usepackage{booktabs} 
\usepackage{algorithm}
\usepackage[noend]{algorithmic}
\usepackage{upgreek}

\usepackage{hyperref}

\usepackage[accepted]{icml2024}

\usepackage{amsmath}
\usepackage{amssymb}
\usepackage{mathtools}
\usepackage{amsthm}
\usepackage[color]{mychangebar}

\usepackage[capitalize,noabbrev]{cleveref}

\theoremstyle{plain}
\newtheorem{theorem}{Theorem}[section]

\newtheorem{corollary}[theorem]{Corollary}
\theoremstyle{definition}

\theoremstyle{remark}

\newtheorem{example}{Example}[section]

\usepackage[textsize=tiny]{todonotes}

\DeclareMathOperator*{\argmin}{arg\,min}

\newcommand{\TRNSPS}{^\mathrm{T}}
\newcommand{\LL}{\mathcal{L}}
\newcommand{\EE}{\mathbb{E}}
\newcommand{\VV}{\mathbb{V}}

\newcommand{\diff}{\mathrm{d}}
\newcommand{\BS}{\text{DT}}
\newcommand{\trg}{\text{Target}}

\newcommand{\vol}{\Omega}
\newcommand{\sur}{\partial\Omega}
\newcommand{\sphr}{\partial\vol_r^{}}

\newcommand{\xp}{x'_1,\cdots,x'_N}
\newcommand{\xpi}{\xp}

\newcommand{\gea}{\frac{1}{N}\sum_{i=1}^{N} g_\theta(x'_i)}
\newcommand{\geatrg}{\frac{1}{N}\sum_{i=1}^{N} g_{\theta^\trg}(x'_i)}

\newcommand{\NSPDF}{P(x'|x)}
\newcommand{\NSPDFI}{P(x'_{1:n}|x)}

\renewcommand{\SS}{K}
\newcommand{\UU}{\mathcal{U}}
\newcommand{\PI}{\Pi}
\newcommand{\SPC}{\;}
\newcommand{\SMK}{K}
\newcommand{\UPOIS}{U}
\newcommand{\EPOIS}{E}
\newcommand{\smrho}{n}
\newcommand{\rhopois}{\rho}

\newcommand{\A}{A}
\newcommand{\B}{B}
\newcommand{\J}{J}
\newcommand{\SSS}{S}
\newcommand{\I}{I}

\newcommand{\thetapu}{\theta^*_{\text{DT}}}
\newcommand\scalemath[2]{\scalebox{#1}{\mbox{\ensuremath{\displaystyle #2}}}}

\icmltitlerunning{Learning from Integral Losses in Physics Informed Neural Networks}

\begin{document}

\twocolumn[
\icmltitle{Learning from Integral Losses in Physics Informed Neural Networks}




\begin{icmlauthorlist}
\icmlauthor{Ehsan Saleh}{cs}
\icmlauthor{Saba Ghaffari}{cs}
\icmlauthor{Timothy Bretl}{ae}
\icmlauthor{Luke Olson}{cs}
\icmlauthor{Matthew West}{mse}
\end{icmlauthorlist}

\icmlaffiliation{cs}{Department of Computer Science}
\icmlaffiliation{ae}{Department of Aerospace Engineering}
\icmlaffiliation{mse}{Department of Mechanical Science and Engineering. University of Illinois Urbana-Champaign}
\icmlcorrespondingauthor{Ehsan Saleh}{ehsans2@illinois.edu}

\icmlkeywords{Machine Learning, ICML}

\vskip 0.3in
]



\printAffiliationsAndNotice{}  

\begin{abstract}
    This work proposes a solution for the problem of training physics-informed networks under partial integro-differential equations. These equations require an infinite or a large number of neural evaluations to construct a single residual for training.  As a result, accurate evaluation may be impractical, and we show that naive approximations at replacing these integrals with unbiased estimates lead to biased loss functions and solutions. To overcome this bias, we investigate three types of potential solutions: the deterministic sampling approaches, the double-sampling trick, and the delayed target method. We consider three classes of PDEs for benchmarking; one defining Poisson problems with singular charges and weak solutions of up to 10 dimensions, another involving weak solutions on electro-magnetic fields and a Maxwell equation, and a third one defining a Smoluchowski coagulation problem. Our numerical results confirm the existence of the aforementioned bias in practice and also show that our proposed delayed target approach can lead to accurate solutions with comparable quality to ones estimated with a large sample size integral. Our implementation is open-source and available at \href{https://github.com/ehsansaleh/btspinn}{https://github.com/ehsansaleh/btspinn}.
\end{abstract}

\section{Introduction}

Physics Informed Neural Networks (PINNs)~\citep{raissi2019physics} can be described as solvers of a particular Partial Differential Equation (PDE). Typically, these problems consist of three defining elements. A sampling procedure selects a number of points for learning. Automatic differentiation is then used to evaluate the PDE at these points and define a residual. Finally, a loss function, such as the Mean Squared Error (MSE), is applied to these residuals, and the network learns the true solution by minimizing this loss through back-propagation and stochastic approximation. These elements form the basis of many methods capable of learning high-dimensional parameters. A wealth of existing work demonstrated the utility of this approach to solving a wide array of applications and PDE forms~\citep{li2020neural,shukla2021parallel,li2019d3m}.

One particular problem in this area is the prevalent assumption around our ability to accurately evaluate the PDE residuals for learning. In particular, partial integro-differential forms include integrals or large summations within them. These forms appear in a broad range of scientific applications including quantum physics~\citep{laskin2000fractional}, aerosol modeling~\citep{wang2022learning}, and ecology~\citep{humphries2010environmental}. In such instances, an accurate evaluation of the PDE elements, even at a single point, can become impractical. Naive approximations, such as replacing integrals with unbiased estimates, can result in biased solutions, as we will show later. This work is dedicated to the problem of learning PINNs with loss functions containing a parametrized integral or summation.

One natural approach for learning PINNs with integral forms would be to use techniques such as importance sampling, numerical quadrature, or Quasi Monte Carlo (QMC) to estimate the integrals more accurately than a standard i.i.d.\ sampling approach. This follows the classical theory and such approaches have been investigated thoroughly in prior work~\citep{caflisch1998monte,evans1995methods}.

\begin{figure*}[t]
	\centering
	\includegraphics[width=0.59\linewidth]{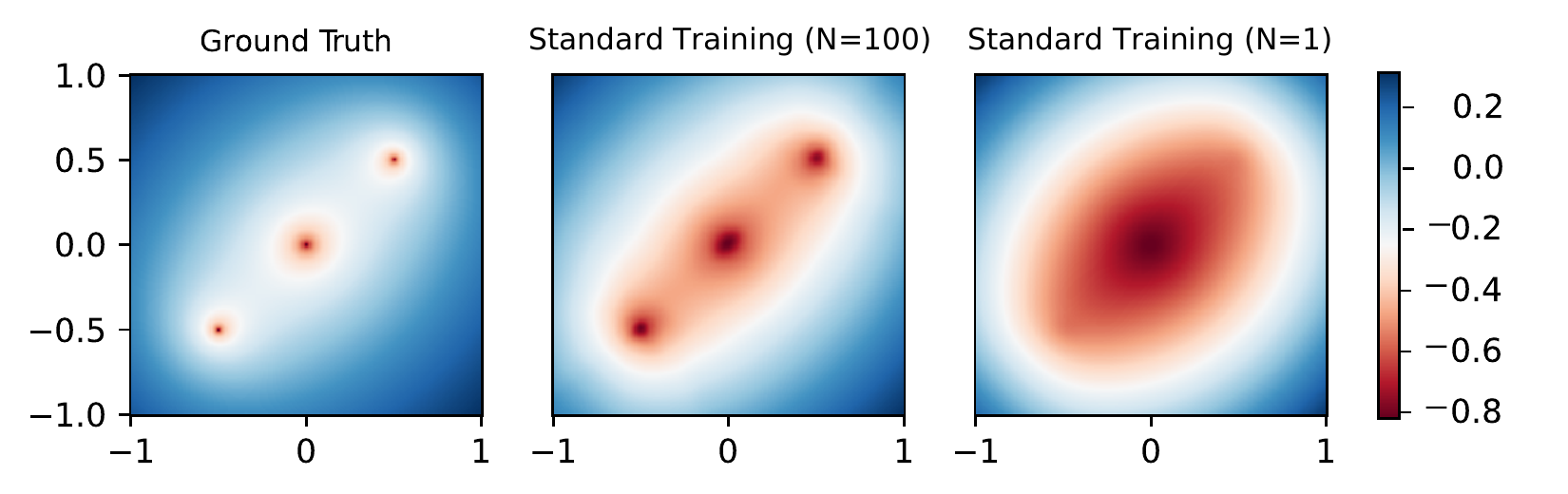}
	\raisebox{-0.05\height}{\includegraphics[width=0.4\linewidth]{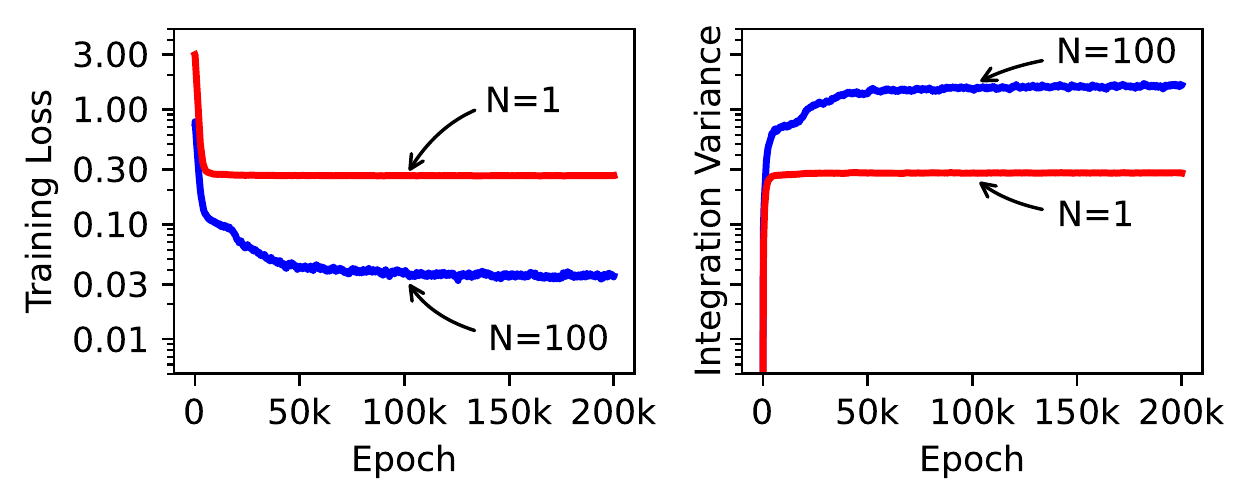}}
	\vspace{-0.5mm}\caption{Training with the MSE loss under different sample sizes per surface ($N$). The heatmaps show the analytical solution (left), the low-variance training with $N=100$ (middle), and the high-variance training with $N=1$ (right). The smaller the $N$, the more biased the training objective becomes towards finding smoother solutions. The right panel shows the training curves; the training loss and the integration variance represent $\hat{\LL}_{\theta}(x)$ and $\VV_{\NSPDF}[g_\theta(x')]$ in Equation~\eqref{eq:excessvar}, respectively. For $N=1$, the training loss seems to be floored at the same value as the integration variance (i.e., approximately $0.3$). However, with $N=100$, the model produces better solutions, lower training losses, and higher integration variances.}\vspace{-1mm}\label{fig:msegt}
\end{figure*}

In this work, we consider an alternative approach, which we will show can be more effective than reducing the variance of the integral estimation. The methods we investigate are based around the idea of reducing the bias and the variance in the parameter gradients so that we can train effectively even if our loss functions are not accurately estimated. We consider three potential approaches to do this; the deterministic sampling approach, the double-sampling trick, and the delayed target method. As we will see, the delayed target approach, which is based upon ideas from learning temporal differences~\citep{sutton1984temporal,mnih2015human,fujimoto2018addressing}, gives the best results, performing comparable or slightly better than accurate integral estimators (i.e., with $N=100$ samples) using just a single sample ($N=1$). Combining importance sampling and QMC methods with our techniques is a promising direction that we leave for future work. 

The main contributions of this work are: (1) we formulate the integral learning problem under a general framework and show the biased nature of standard approximated loss functions; (2) we present three techniques to solve such problems, namely the deterministic sampling approaches, the double-sampling trick, and the delayed target method; (3) we detail an effective way of implementation for the delayed target method compared to a naive one; (4) we compare the efficacy of the potential solutions using numerical examples on Poisson problems with singular charges and up to 10 dimensions, a Maxwell problem with magnetic fields, and a Smoluchowski coagulation problem; (5) provide a convergence guarantee, approximation error upper bound, and computational complexity analysis for the delayed target method under linear function approximation.

\section{Problem Formulation}\label{sec:prblmfrmltn}

Consider a typical partial integro-differential equation
\begin{equation}\label{eq:typicalipde}
f_\theta(x) := \mathbb{E}_{\NSPDF}[g_\theta(x')] + y(x).
\end{equation}
The $f_\theta(x)$ and $g_\theta(x')$ are parametrized, and $y(x)$ includes all the non-parametrized terms in the PDE\@. The right side of the equation serves as the target value for $f_\theta(x)$ (see Section~\ref{sec:pinnnotation} of the supplementary material for all the notation). Equation~\eqref{eq:typicalipde} is a general, yet concise, form for expressing partial integro-differential equations. To motivate this, we will express three examples in this form. 

\begin{example}\label{ex:poisson}
The Poisson problem is to solve the system $\nabla^2 \UPOIS = \rhopois$ for $\UPOIS$ given a charge function $\rhopois$. This is equivalent to finding a solution for a gradient and divergence system:
\begin{align}
\EPOIS &= \nabla \UPOIS, \label{eq:poissys1}\\
\rhopois &= \nabla \cdot \EPOIS. \label{eq:poissys2}
\end{align}
A weak solution can be obtained by enforcing the divergence theorem over many volumes:
\begin{equation}\label{eq:divthm}
\int_{\sur} \EPOIS\cdot \hat{n}\quad \diff S = \iint_{\vol} \nabla\cdot \EPOIS\quad \diff V,
\end{equation}
where $\hat{n}$ is the normal vector perpendicular to $\diff S$. The weak solutions can be preferable over the strong ones when dealing with singular or sparse $\rhopois$ charges.

To solve this system, we parametrize $\EPOIS$ as the gradient of a neural network predicting the $\UPOIS$ potentials. To convert this into the form of Equation~\eqref{eq:typicalipde}, we replace the left integral in Equation~\eqref{eq:divthm} with an arbitrarily large Riemann sum as
\begin{equation}\label{eq:reimansum}
\int_{\sur} \EPOIS\cdot \hat{n}~\diff S =\frac{A}{M} \sum_{i=1}^{M} E_\theta(x_i)\cdot \hat{n}_i,
\end{equation}
where $A=\int_{\sur} 1\text{ d}S$ is the surface area and the $x_i$ samples are uniform on the surface. To convert this system into the form of Equation~\eqref{eq:typicalipde}, we define the following elements:
\begin{align}
x&:=x_1,\label{eq:poissdefs1}\\
f_\theta(x)&:=\frac{A}{M} E_\theta(x)\cdot \hat{n}_1, \label{eq:poissdefs2}\\
g_\theta(x_i)&:=-\frac{A(M-1)}{M} E_\theta(x_i) \cdot \hat{n}_i, \label{eq:poissdefs3}\\
P(x'|x_1) &:= \text{Unif}(\{x_2, \cdots , x_M\}), \label{eq:poissdefs4}\\
y(x_1)&:=\iint_{\vol} \rhopois~\diff V. \label{eq:poissdefs5}
\end{align}
\end{example}

\begin{figure*}[t]
	\centering
	\includegraphics[width=0.305\linewidth]{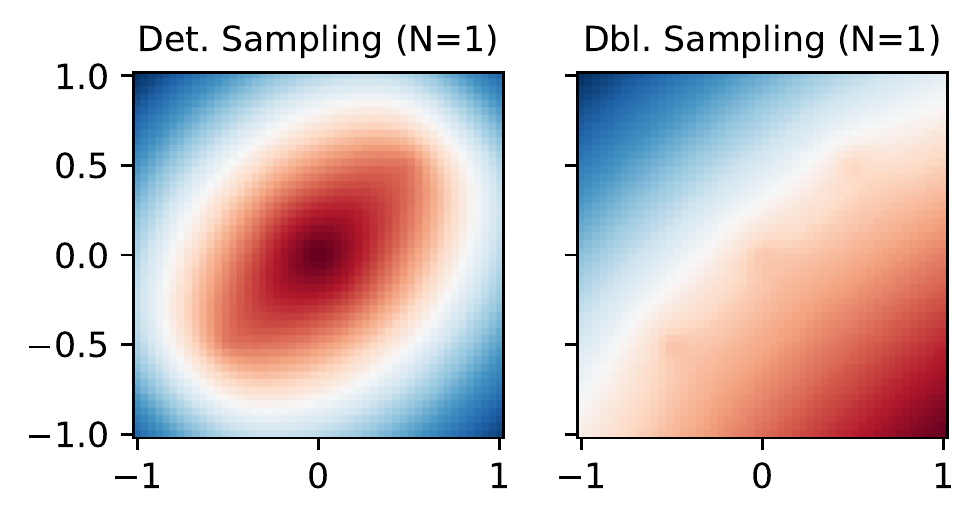}
	\raisebox{0.0\height}{\includegraphics[width=0.18\linewidth]{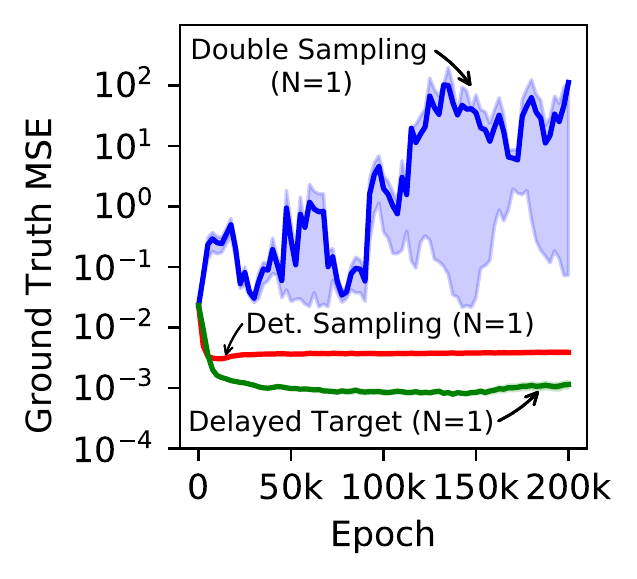}}
	\includegraphics[width=0.305\linewidth]{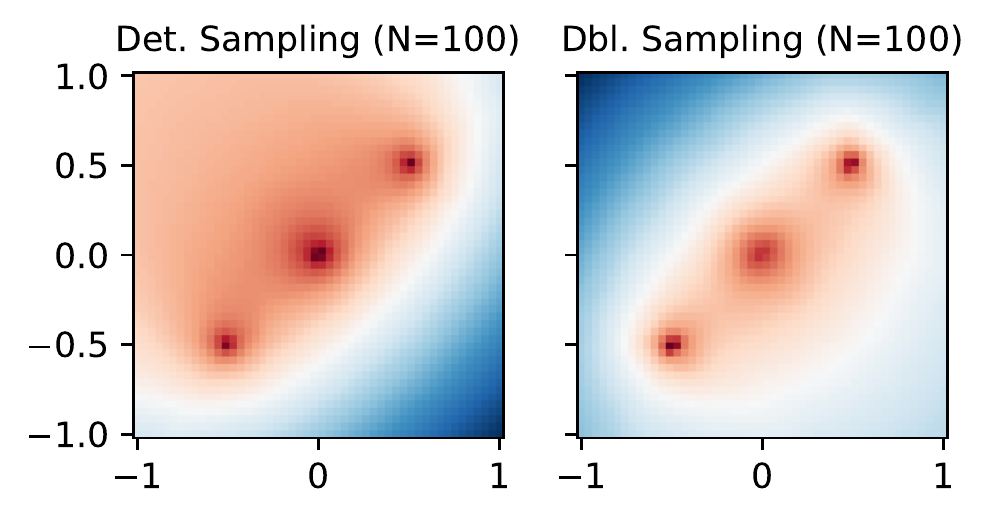}
	\raisebox{0.0\height}{\includegraphics[width=0.18\linewidth]{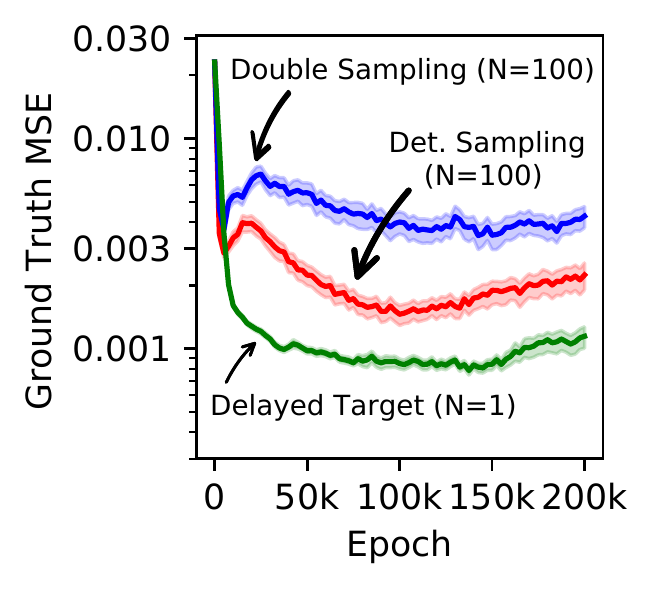}}
	\vspace{-4mm}\caption{The results of the deterministic and double sampling techniques on the Poisson problem. The left plots demonstrate the solutions with $N=1$, while the right plots show the solutions with $N=100$. The training curves represent the mean squared error to the analytical solution vs. the training epochs. With $N=1$, the double sampling trick exhibits divergence in training, and the deterministic sampling process yields overly smooth functions similar to the standard solution in Figure~\ref{fig:msegt}. However, with $N=100$, both the deterministic and double-sampling approaches exhibit improvements. According to the training curves, the delayed target method with $N=1$ yields the best solutions to this problem.}\vspace{-1mm}
	\label{fig:dtsdblpoisson}
\end{figure*}

\begin{example}\label{ex:maxwell}
In static electromagnetic conditions, one of the Maxwell Equations, the Ampere circuital law, is to solve the $\nabla\times \A = \B$ and $\nabla \times \B=\J$ system for $\A$ given the current density $\J$ in the 3D space (we assumed a unit physical coefficient for simplicity). Here, $\B$ represents the magnetic field and $\A$ denotes the magnetic potential vector. A weak solution for this system can be obtained by enforcing the Stokes theorem over many volumes:
\begin{equation}\label{eq:stksthm}
\int_{\sur} \nabla \times \A \cdot \diff l = \iint_{\vol} \J \cdot \diff \SSS,
\end{equation}
where $\diff l$ and $\diff \SSS$ are infinitesimal surface tangent and normal vectors, respectively. Just like the Poisson problem, the weak solutions can be preferable over the strong ones when dealing with singular inputs, and this equation can be converted into the form of Equation~\eqref{eq:typicalipde} similarly.
\end{example}

\begin{example}\label{ex:smoluchowski} 
The Smoluchowski coagulation equation simulates the evolution of particles into larger ones and is described as
\begin{align}\label{eq:smoll}
\frac{\partial \smrho(x,t)}{\partial t} = &\int_{0}^{x} \SMK(x-x', x') \smrho(x-x', t) \smrho(x',t) \diff x' \nonumber\\
&- \int_{0}^{\infty} \SMK(x,x')\smrho(x,t)\smrho(x',t)\diff x',
\end{align}
where $\SMK(x,x')$ is the coagulation kernel between two particles of size $x$ and $x'$. The particle sizes $x$ and $x'$ can be generalized into vectors, inducing a higher-dimensional PDE to solve. To solve this problem, we parametrize $\smrho(x,t)$ as the output of a neural model and write
\begin{align}
f_\theta(x)&:=\frac{\partial \smrho_\theta(x,t)}{\partial t}, \\
g_\theta^{(1)}(x')&:=A_1 \SMK(x-x', x')\smrho_\theta(x-x', t) \smrho_\theta(x',t), \\
g_\theta^{(2)}(x')&:=A_2 \SMK(x,x')\smrho_\theta(x,t)\smrho_\theta(x',t).\vspace{-3mm}
\end{align}
The $x'$ values in both $g_\theta^{(1)}$ and $g_\theta^{(2)}$ are sampled from their respective uniform distributions, and $A_1$ and $A_2$ are used to normalize the uniform integrals into expectations. Finally, $y(x):=0$ and we can define $g_\theta(x')$ in a way such that
\begin{equation}
\EE_{x'}[g_\theta(x')]:=\EE_{x'}[g_\theta^{(1)}(x')] + \EE_{x'}[g_\theta^{(2)}(x')].
\end{equation}
\end{example}
\vspace{-2mm}The standard way to solve systems such as Examples~\ref{ex:poisson},~\ref{ex:maxwell}, and~\ref{ex:smoluchowski} with PINNs, is to minimize the following mean squared error (MSE) loss~\citep{raissi2019physics,jagtap2020conservative}:
\begin{equation}\label{eq:origmseloss}
\LL_{\theta}(x) := \big( f_\theta(x) - \EE_{\NSPDF}[g_\theta(x')] - y(x)\big)^2.
\end{equation}
\begin{algorithm}[h]
	\caption{The regularized delayed target method}
	\begin{algorithmic}[1]
		\label{alg:brs}
		\REQUIRE The initial parameter values $\theta_0$, learning rate $\eta$, Polyak averaging rate $\tau$, target sample size $N$, and the target regularization weight $\lambda$.
		\STATE Initialize the main and target parameters:$\theta,\theta_\trg  \leftarrow \theta_0$.
		\FOR{$k = 1, 2, \ldots$}
        \STATE Sample $x$ from $P$ and obtain the $y(x)$ label.
		\STATE Compute the $f_\theta(x)$ term using the main parameters.
		\STATE Obtain the $\xpi$ i.i.d.\ samples from $P(x'|x)$.
		\STATE Compute the $\geatrg + y(x)$ target using the $\theta_\trg$ target parameters.
		\STATE Construct the main loss: 
        \vspace{-3.5mm}\begin{equation}
            \hat{\LL}_{\theta}^{\BS} = \big(f_\theta(x) - \geatrg - y(x)\big)^2.
		\vspace{-2.5mm}\end{equation}
		\STATE Construct the target regularization loss: 
        \vspace{-1mm}\begin{equation}
            \hat{\LL}_\theta^{\text{R}}=(f_\theta(x) - f_{\theta^\trg}(x))^2.
		\vspace{-1mm}\end{equation}
		\STATE Compute the total loss $\hat{\LL}_\theta^{\BS\text{R}}=\hat{\LL}^{\BS}_{\theta}(x) + \lambda \hat{\LL}_\theta^{\text{R}}$.
		\STATE Perform a gradient descent step on $\theta$: 
        \vspace{-2mm}\begin{equation}
            \theta \leftarrow \theta - \eta \nabla_\theta \hat{\LL}_\theta^{\BS\text{R}}.
		\vspace{-2mm}\end{equation}
		\STATE Update the target parameters using Polyak averaging: 
        \vspace{-3mm}\begin{equation}
            \theta^\trg \leftarrow \tau \theta^\trg + (1-\tau) \theta.
		\vspace{-1mm}\end{equation}
		\ENDFOR
	\end{algorithmic}
\end{algorithm}
Since computing exact integrals may be impractical, one may contemplate replacing the expectation in Equation~\eqref{eq:origmseloss} with an unbiased estimate, as implemented in NVIDIA's Modulus package~\citep{modulus2022pinn}. This prompts the following approximate objective:
\begin{equation}\label{eq:aprxmseloss}
	\scalemath{0.905}{\hat{\LL}_{\theta}(x) := \EE_{\NSPDFI}\bigg[\big( f_\theta(x) - \gea - y(x)\big)^2\bigg].}
\end{equation}
\begin{figure*}[t]
	\centering
	\includegraphics[width=0.34\linewidth]{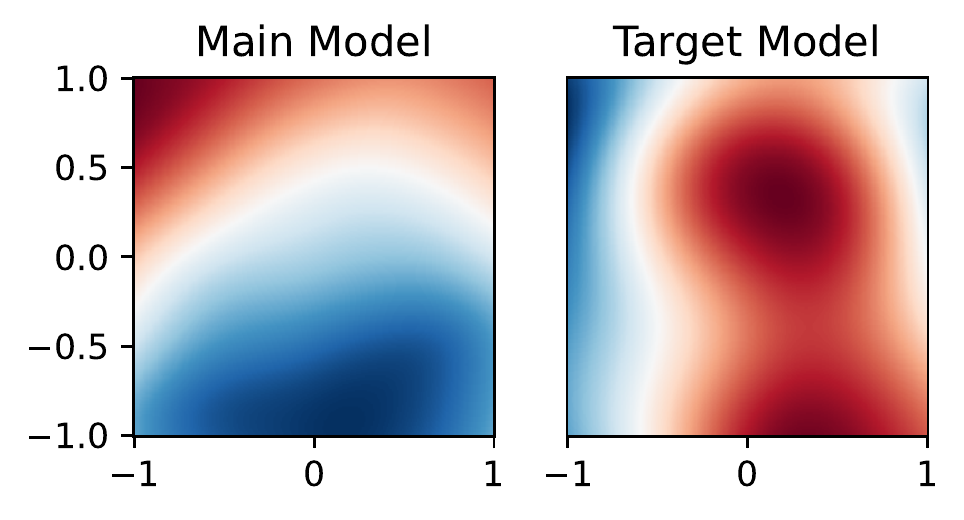}\hspace{-1mm}
	\includegraphics[width=0.155\linewidth]{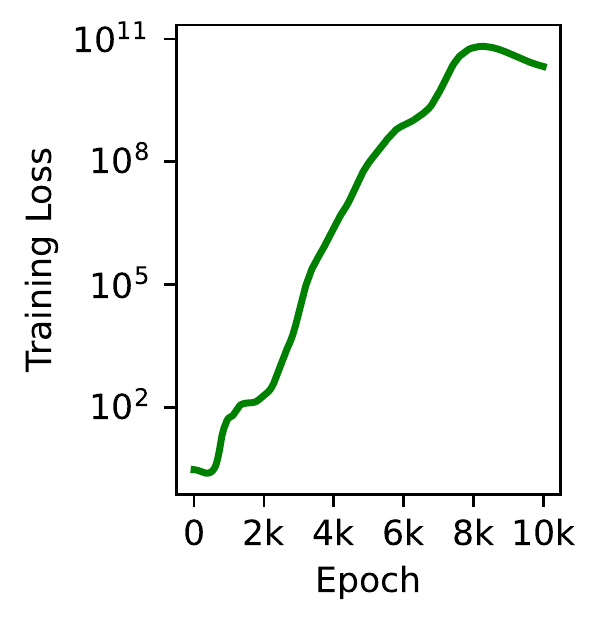}\hspace{9mm}
	\includegraphics[width=0.18\linewidth]{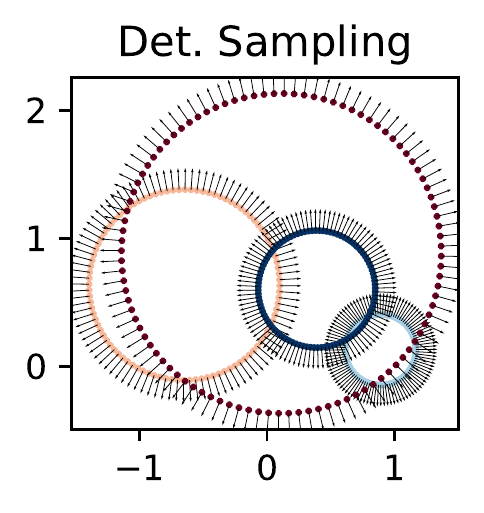}
	\includegraphics[width=0.18\linewidth]{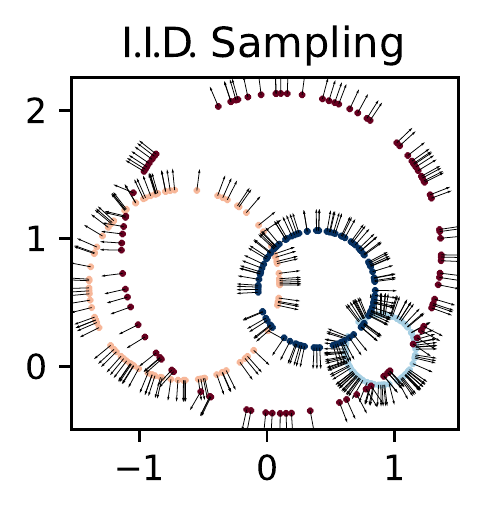}\phantom{aaaaa}\\
	\includegraphics[width=0.33\linewidth]{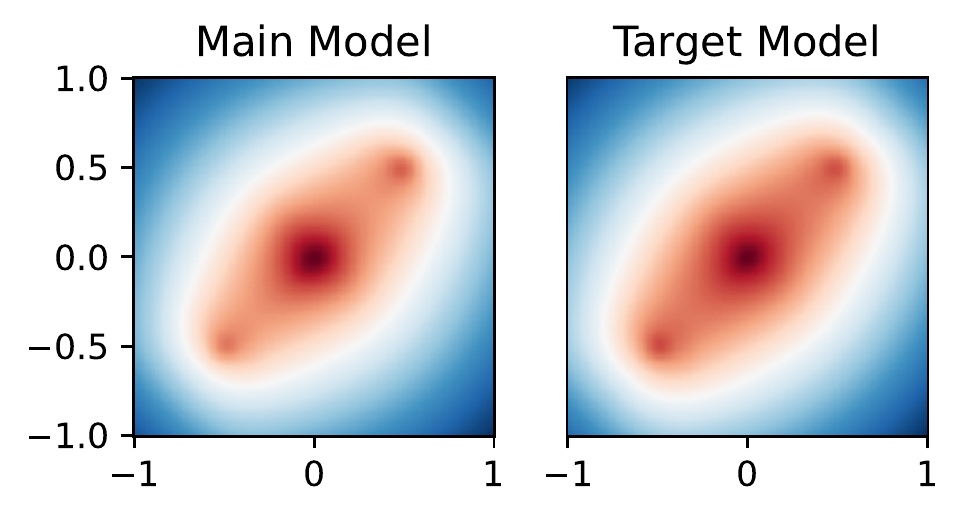}\hspace{-1mm}
	\includegraphics[width=0.155\linewidth]{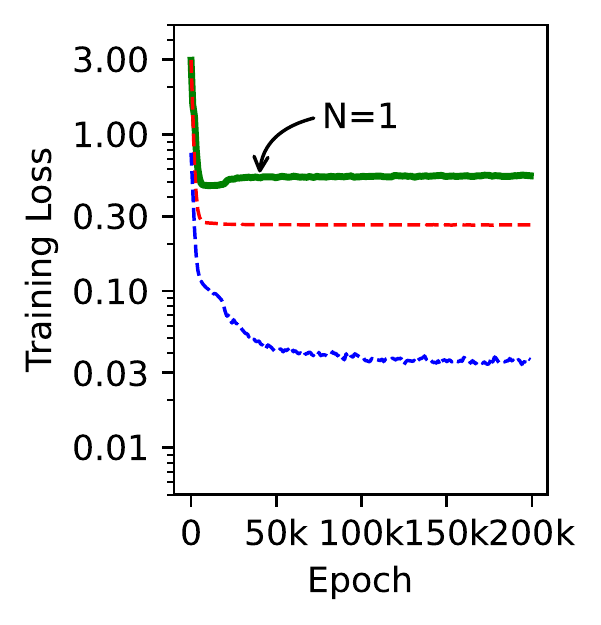}\hspace{-3mm}
	\raisebox{0.1\height}{\includegraphics[width=0.37\linewidth]{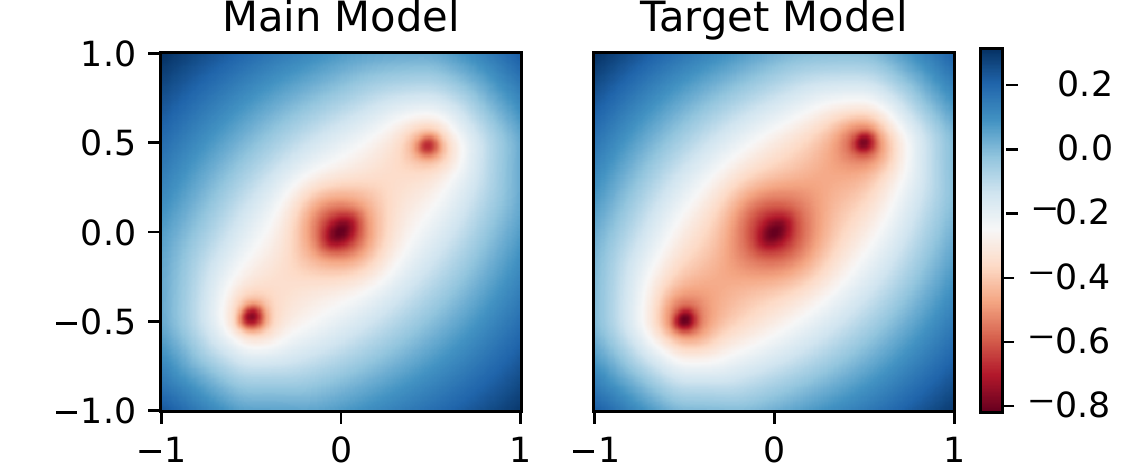}}\hspace{-2mm}
	\includegraphics[width=0.155\linewidth]{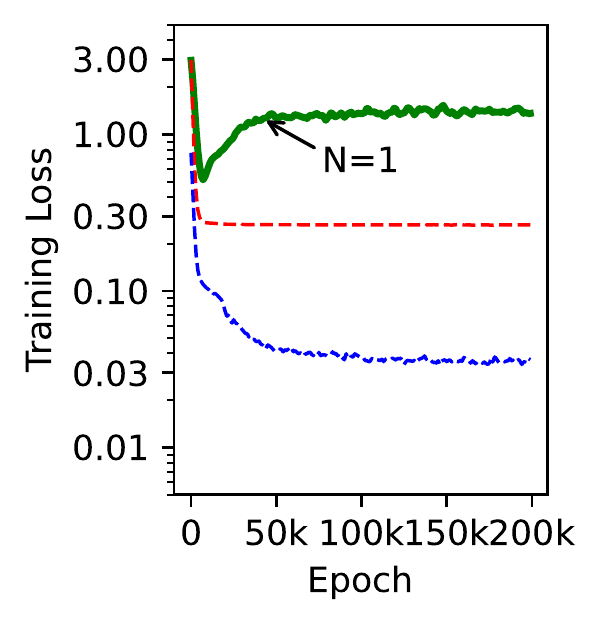}
	
	\vspace{-2mm}\caption{\label{fig:divergedbstrap}Training the same problem as in Figure~\ref{fig:msegt} with delayed targets and $N=1$. \textit{The top left panel} shows a diverged training with $M=100$ in Equation~\eqref{eq:divthmbsloss}. \textit{The lower left panel} corresponds to $M=10$, which has a converging training curve even though it produces an overly smooth solution. \label{fig:bstrapconv}In \textit{the lower right panel}, we set $\lambda=1$ which allowed setting $M=1000$ while maintaining a stable training loss. In each panel, the left and right heatmaps show the main and the target model predictions, respectively, and the right plots show the training curves. The green curves show the training loss for the delayed target method, and the standard training curves with $N=1$ and $100$ are also shown using dotted red and blue lines for comparison, respectively. \label{fig:detiidsampling}\textit{The top right panel} shows an example of deterministic vs.\ i.i.d.\ sampling of the surface points in the Poisson problem. For each sampled sphere, the surface points and their normal vectors are shown with $N=100$ samples. With deterministic sampling, the points are evenly spaced to cover the sampling domain.}\vspace{-2mm}
\end{figure*}
We therefore analyze the approximation error by adding and subtracting $\EE_{x''}[g_\theta(x'')]$:
\begin{align}
\hat{\LL}_{\theta}(x) = &\EE_{\NSPDFI}\bigg[\bigg( \big(f_\theta(x) - \EE_{x''}[g_\theta(x'')]  - y(x)\big) + \nonumber\\
&\big(\EE_{x''}[g_\theta(x'')] - \gea \big)\bigg)^2\bigg].
\end{align}
\vspace{-0.5mm}By decomposing the squared sum, we get
\begin{align}\label{eq:biasder1}
\hat{\LL}_{\theta}&(x) =\, \big(f_\theta(x) - \EE_{x''}[g_\theta(x'')]  - y(x)\big)^2  
+ \nonumber\\
&\EE_{\NSPDFI}\big[\big(\EE_{x''}[g_\theta(x'')] - \frac{1}{N}\sum_{i=1}^{N} g_\theta(x'_i)\big)^2\big] + \nonumber\\
&2\,\EE_{\NSPDFI}\big[ \big(f_\theta(x) - \EE_{x''}[g_\theta(x'')]  - y(x)\big) \nonumber\\
&\phantom{aaaaa} \big(\EE_{x''}[g_\theta(x'')] - \frac{1}{N}\sum_{i=1}^{N} g_\theta(x'_i)\big)\big].
\end{align}
Since $\EE_{x''}[g_\theta(x'')] = \EE_{\xp}[\frac{1}{N}\sum_{i=1}^{N} g_\theta(x'_i)]$, the last term in Equation~\eqref{eq:biasder1} is zero, and we have
\vspace{-1.0mm}\begin{equation}\label{eq:excessvargen}
\hat{\LL}_{\theta}(x)  = \LL_{\theta}(x) + \VV_{\NSPDFI}[\gea],
\vspace{-1mm}\end{equation}
where $\VV$ denotes the variance operator. If the $\xpi$ values are sampled in an i.i.d.\ manner, Equation~\eqref{eq:excessvargen} simplifies further to
\vspace{-1.0mm}\begin{equation}\label{eq:excessvar}
\hat{\LL}_{\theta}(x)  = \LL_{\theta}(x) + \frac{1}{N} \VV_{\NSPDF}[g_\theta(x')].
\vspace{-1.0mm}\end{equation}
The induced excess variance in Equation~\eqref{eq:excessvar} can bias the optimal solution.  As a result, optimizing the approximated loss will prefer smoother solutions over all $\xpi$ samples. It is worth noting that this bias is mostly harmful due to its \textit{parametrized nature}; the only link through which this bias can offset the optimal solution is its \textit{dependency on $\theta$}. This is in contrast to any non-parametrized stochasticity in the $y$ term of Equation~\eqref{eq:origmseloss}. Non-parameterized terms cannot offset the optimal solutions, since stochastic gradient descent methods are indifferent to them.

\vspace{-3.5mm}\section{Potential Solutions}\label{sec:ptntlsltns}

\vspace{-0.5mm}Based on Equation~\eqref{eq:excessvar}, the induced bias in the solution has a direct relationship with the stochasticity of the $P(x'|x)$ distribution. If we were to sample the $(x,x')$ pairs deterministically, the excess variance in Equation~\eqref{eq:excessvar} would disappear. However, this results in modifying the problem conditions. Next, we introduce three potential solutions to this problem: the \textit{deterministic sampling approaches}, the \textit{double-sampling trick}, and the \textit{delayed target method} which is based upon the method of learning from temporal differences~\citep{sutton1984temporal}.
\newcommand{\ax}{\mathrm{T}^x}
\newcommand{\axgea}{\frac{1}{N}\sum_{i=1}^{N}g_\theta(\ax_{i}) }
\newcommand{\dax}{\delta_{\ax}}
\newcommand{\dts}{\text{DET}}
\newcommand{\dbl}{\text{DBL}}

\begin{figure*}[t]
	\centering
	\includegraphics[width=0.713\linewidth]{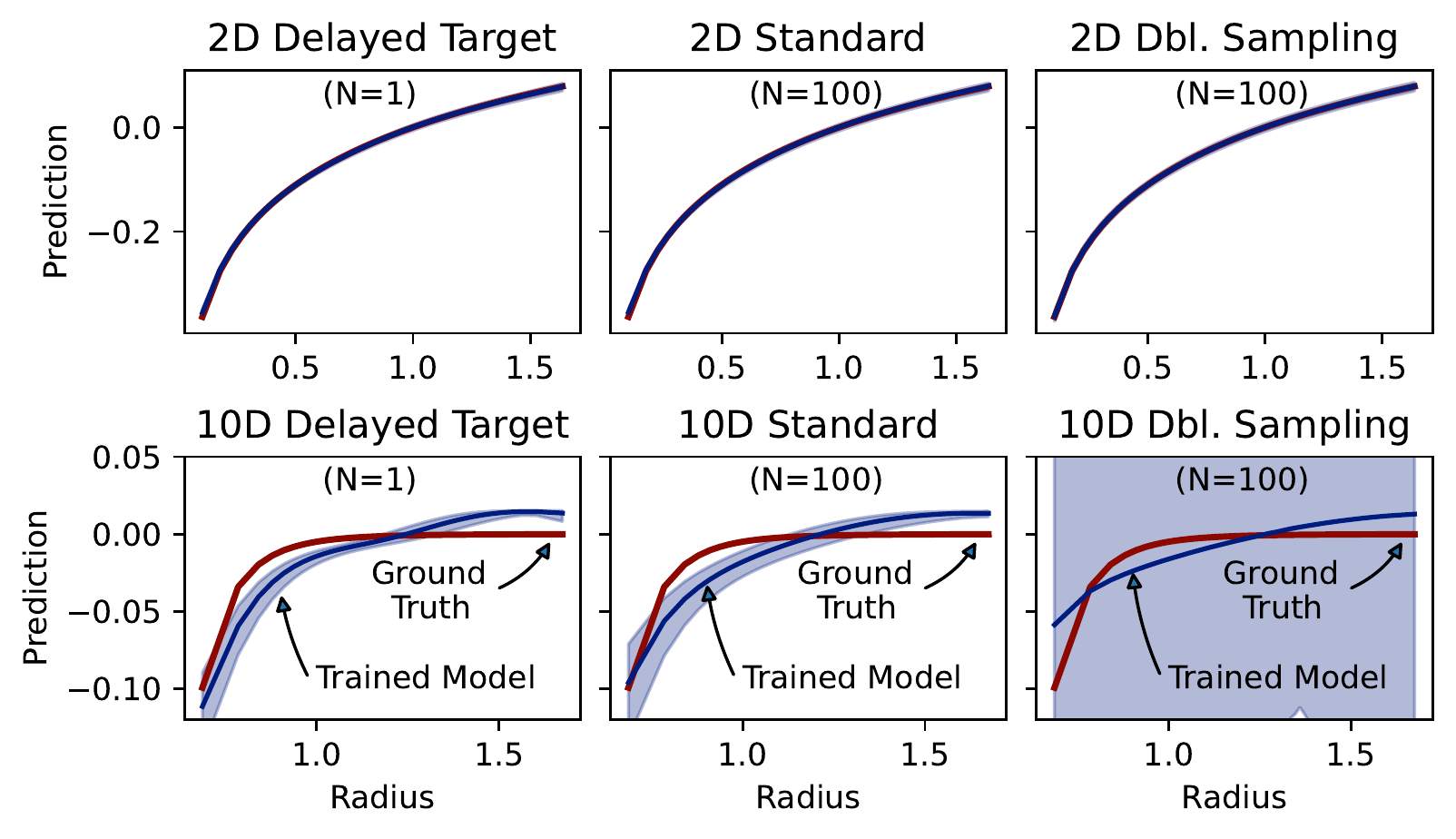}
	\includegraphics[width=0.277\linewidth]{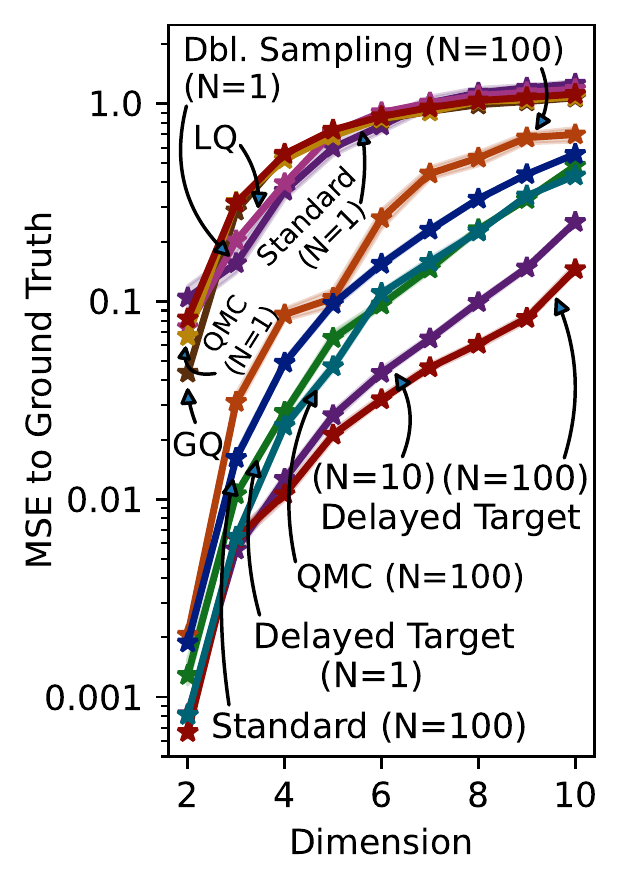}
	\vspace{-5mm}\caption{The solution and performance curves in higher-dimensional Poisson problems. \textit{The left panel} shows the solution curves for the delayed target $(N=1)$, the standard $(N=100)$, and the double-sampling $(N=100)$ methods. The top and the bottom rows show 2- and 10-dimensional problems, respectively. In these problems, a single charge is located at the origin, so that the analytical solution is a function of the evaluation point radii $\|x\|$. The horizontal axis shows the evaluation point radii and covers 98\% of points within the training volumes. \textit{The right chart} is a performance curve against the problem dimension (lower is better). The normalized MSE values were shown to be comparable. These results suggest that (1) higher dimensions make the problem challenging, and (2) delayed targeting with $N=1$ is comparable to standard trainings with $N=100$. GQ and LQ refer to Gaussian and Leja quadrature, respectively, under a Smolyak sparse grid. Sections~\ref{sec:quadqmcsamplng}, ~\ref{sec:dtsampsizeabls}, and~\ref{sec:hdpevalproto} of the supplementary material describe the effect of sampling dimension on numerical quadrature and QMC, the effective way of scaling up $N$ for delayed targeting, and the performance evaluation profile, respectively.} 
	\label{fig:hidimpoiss}\vspace{-4mm}
\end{figure*}

\vspace{-2.5mm}\subsection{The Deterministic Sampling Approaches} 
\vspace{-0.5mm}One approach to eliminate the excess variance term in Equation~\eqref{eq:excessvargen}, is to sample the $(\xpi)$ tuple in a way that $\NSPDFI$ would be a point mass distribution at a fixed $\ax$ tuple. This way, $\NSPDFI$ yields a zero excess variance: 
\vspace{-2.5mm}\begin{equation}
    \VV_{\NSPDFI}[\axgea] = 0.
\vspace{-1.5mm}\end{equation} 
This induces the following deterministic loss.
\vspace{-2.0mm}\begin{equation}\label{eq:dtsloss}
\hat{\LL}^{\dts}_{\theta}(x) := \bigg( f_\theta(x) - \axgea - y(x)\bigg)^2.
\vspace{-2.0mm}\end{equation}
Although this approach removes the excess variance term in Equation~\eqref{eq:excessvargen} thanks to its deterministic nature, it biases the optimization loss by re-defining it: $\LL_\theta(x) \neq \hat{\LL}^{\dts}_{\theta}(x)$. The choice of the $\ax$ samples can influence the extent of this discrepancy. One reasonable choice is to evenly space the $N$ samples to cover the entire sampling domain as uniformly as possible. For a demonstration, Figure~\ref{fig:detiidsampling} shows a number of example sets used for applying the divergence theorem to the Poisson problem. Of course, this sampling strategy can be impractical in high-dimensional spaces as the number of samples needed to cover the entire sampling domain grows exponentially with the sampling space dimension. This could be partially ameliorated by the use of QMC methods~\citep{morokoff1995quasi}.

\vspace{-1mm}Numerical quadrature offers another deterministic approach for accurate integral estimation. By choosing specific integration points and weights, they can provably yield accurate integral estimates under certain function classes; for instance, Gaussian quadrature~\citep{gauss1814methodus} with $N$ samples can produce exact $(2N-1)$-degree polynomial integrals. However, these methods still suffer from the curse of dimensionality and are more restrictive than the QMC alternatives in their choice of $N$. This exponential sample requirement can be partially ameliorated by the use of sparse grid methods such as Smolyak's quadrature~\citep{smolyak1963quadrature}.



\vspace{-2mm}\subsection{The Double-Sampling Trick}
If we have two independent $x'$ samples, namely $x'_1$ and $x'_2$, we can replace the objective in Equation~\eqref{eq:aprxmseloss} with
\begin{align}\label{eq:dblsamp}
\hspace{-1mm}\hat{\LL}^{\dbl}_{\theta}(x) = &\EE_{x'_1,x'_2\sim P(\cdot|x)}\bigg[\big( f_\theta(x) - g_\theta(x'_1) - y(x)\big)\nonumber\\
&\big( f_\theta(x) - g_\theta(x'_2) - y(x)\big)\bigg].
\end{align}
It is straightforward to show that $\hat{\LL}^{\dbl}_{\theta}(x) = \LL_{\theta}(x)$; the uncorrelation between $g_\theta(x'_1)$ and $g_\theta(x'_2)$ will remove the induced bias on average. However, this approach requires access to two i.i.d.\ samples, which may not be plausible in many sampling schemes. In particular, Monte-Carlo samplings used in reinforcement learning do not usually afford the learning method with the freedom to choose multiple next samples or the ability to reset to a previous state. Besides reinforcement learning, offline learning using a given collection of samples may make this approach impractical. It is possible to simulate $N=1$ (and similarly for larger $N$) in problems of the form $\mathbb{E}_{\NSPDF}[g_\theta(x')] = y(x)$, such as Examples~\ref{ex:poisson} and~\ref{ex:maxwell}, by redefining $\hat{\LL}^{\dbl}$ as
\begin{align}
	\hat{\LL}^{\dbl}_{\theta}(x) = &\EE_{x'_1,x'_2\sim P(\cdot|x)}\bigg[\big( g_\theta(x'_1) - y(x)\big)\nonumber\\ 
	&\big(g_\theta(x'_2) - y(x)\big)\bigg].
\end{align}
\begin{figure*}[t]
	\centering
	\vspace{-1mm}\includegraphics[width=0.67\linewidth]{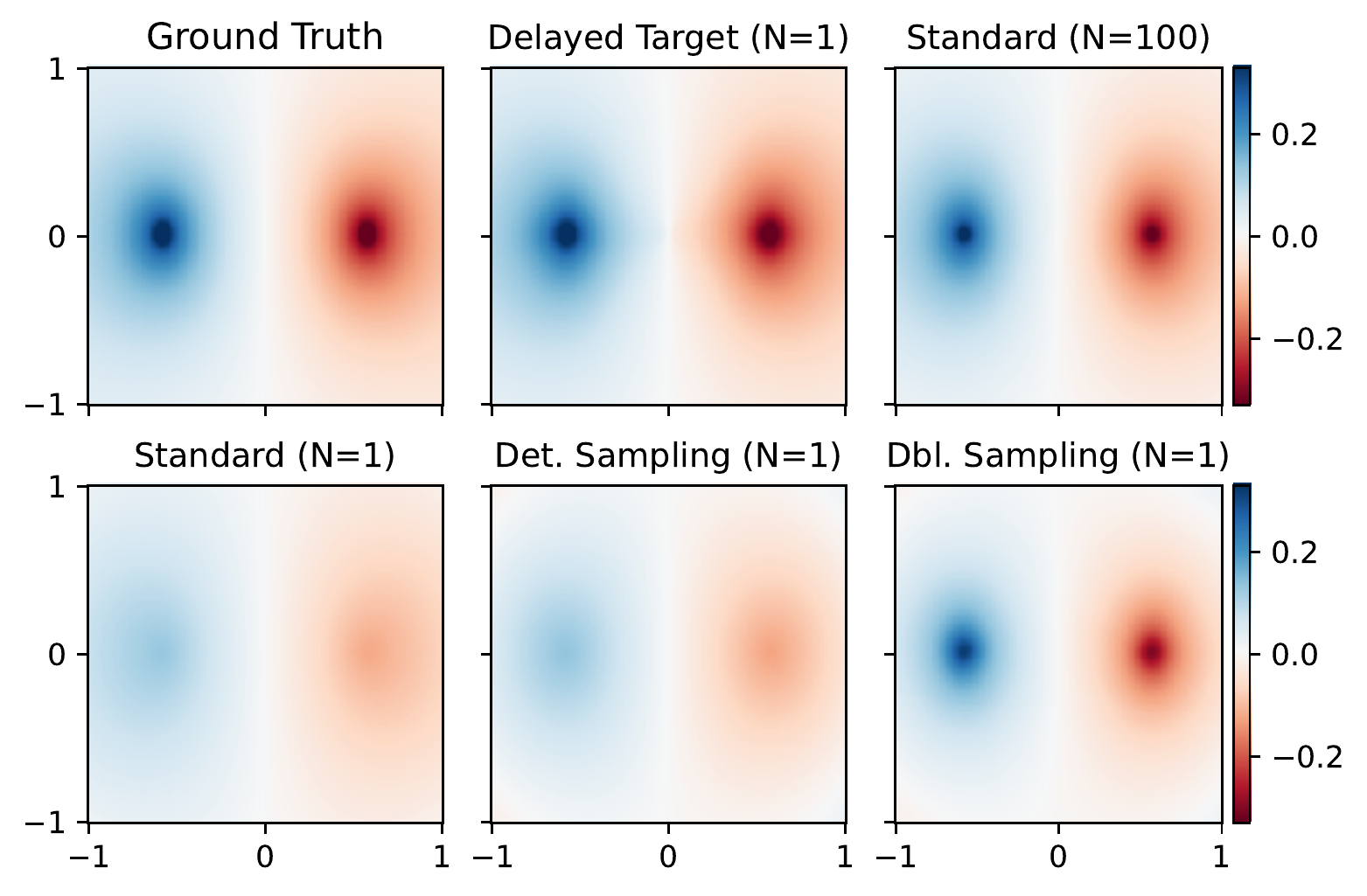}
	\includegraphics[width=0.32\linewidth]{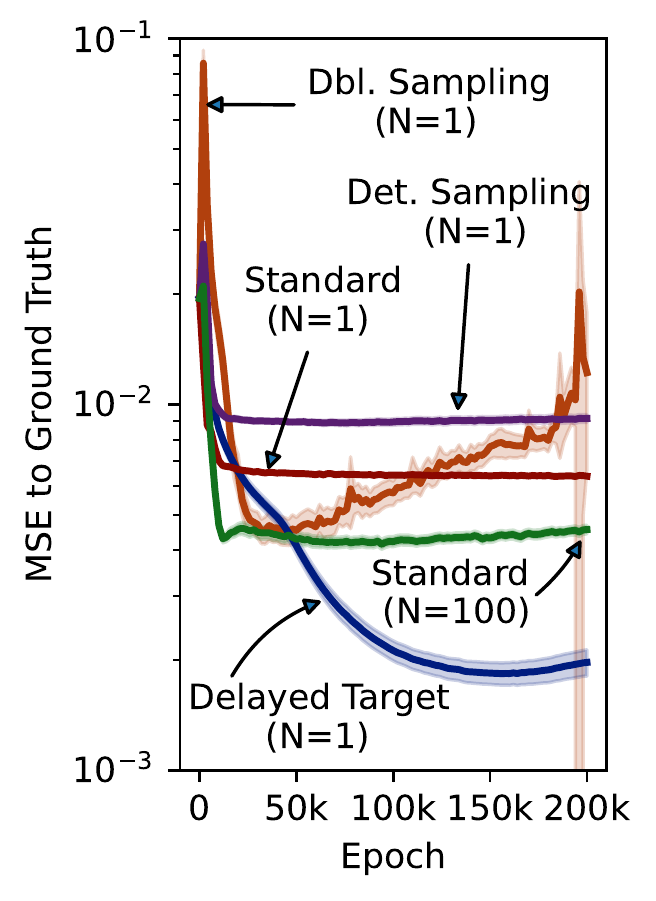}
	\vspace{-2mm}\caption{The solution heatmaps and the training curves for different methods to the Maxwell problem. \textit{In the left panel}, we show a single component of the magnetic potentials ($A_z$) in a 2D slice of the training space with $z=0$ for visual comparison. \textit{In the right plot}, we show the training curves. The results suggest that (1) the standard and deterministic trainings with $N=1$ produce overly smooth solutions, and (2) delayed targeting with $N=1$ is comparable to standard trainings with $N=100$. Section~\ref{sec:trgtaulammaxwell} of the supplementary material studies the target smoothing and regularization weights of the delayed target method in this problem.}
	\label{fig:maxwellmain}\vspace{-3mm}
\end{figure*}
\vspace{-6mm}\subsection{The Delayed Target Method} 
This approach replaces the objective in Equation~\eqref{eq:aprxmseloss} with
\begin{equation}\label{eq:bstraploss}
\LL^{\BS}_{\theta}(x) = \EE_{\NSPDF}\bigg[\big(f_\theta(x) - g_{\theta^*}(x') - y(x)\big)^2\bigg],
\end{equation}
where we have $\theta^* := \argmin_{\tilde{\theta}}  \LL_{\tilde{\theta}}(x)$. Assuming a complete function approximation set $\Theta$ (where $\theta \in \Theta$), we know that $\theta^*$ satisfies Equation~\eqref{eq:typicalipde} at all $x$. Therefore, we have
\begin{equation}
\nabla_{\theta} \LL_\theta(x)\big|_{\theta=\theta^*} = \nabla_\theta \LL^{\BS}_{\theta}(x)\big|_{\theta=\theta^*}=0.
\end{equation}
As a result, we can claim
\begin{equation}
\theta^* = \argmin_\theta \EE_x[\LL^{\BS}_{\theta}(x)] = \argmin_\theta \EE_x[\LL_{\theta}(x)].
\end{equation}
In other words, optimizing Equation~\eqref{eq:bstraploss} should yield the same solution as optimizing the true objective $\LL_\theta(x)$ in Equation~\eqref{eq:origmseloss}. Of course, finding $\theta^*$ is as difficult as solving the original problem. The simplest heuristic replaces $\theta^*$ with a supposedly independent, yet identically valued, version of the latest $\theta$ named $\theta^\trg$, hence the \textit{delayed}, \textit{detached}, and \textit{bootstrapped target} naming conventions:
\begin{equation}\label{eq:bstraplossapprx}
	\scalemath{0.845}{\hat{\LL}^{\BS}_{\theta}(x) = \EE_{\NSPDFI}\bigg[\big(f_\theta(x) - \geatrg - y(x)\big)^2\bigg].}
\end{equation}
Our hope would be for this approximation to improve as well as $\theta$ over training. The only practical difference between implementing this approach and minimizing the loss in Equation~\eqref{eq:origmseloss} is to use an incomplete gradient for updating $\theta$ by detaching the $g(x')$ node from the computational graph in the automatic differentiation software. This naive implementation of the delayed target method can lead to divergence in optimization, as we will show in Section~\ref{sec:results} with numerical examples (i.e., Figure~\ref{fig:bstrapconv}). Here, we introduce two mitigation factors contributing to the stabilization of such a technique.

\vspace{-2mm}\paragraph{Moving Target Stabilization} One disadvantage of the aforementioned technique is that it does not define a global optimization objective; even the average target for $f_{\theta}(x)$ (i.e., $\EE_{P(x'|x)}\big[g_{\theta^\trg}(x')\big]+y(x)$) changes throughout the training. Therefore, a naive implementation can risk training instability or even divergence thanks to the moving targets.

To alleviate the fast-moving targets issue, prior work suggested fixing the target network for many time-steps~\citep{mnih2015human}. This causes the training trajectory to be divided into a number of episodes, where the target is locally constant and the training is therefore locally stable in each episode. Alternatively, this stabilization can be implemented continuously using Polyak averaging~\citep{fujimoto2018addressing}; instead of fixing the target network for a window of $T$ steps, the target parameters $\theta^\trg$ can be updated slowly with the following rule:
\begin{equation}
\theta^\trg \leftarrow \tau \theta^\trg + (1-\tau) \theta.
\end{equation}
This exponential moving average defines a corresponding stability window of $T=O(1/(1-\tau))$.
\vspace{-2mm}\paragraph{Prior Imposition for Highly Stochastic Targets}
In certain instances, the $\geatrg+y(x)$ target in Equation~\eqref{eq:bstraploss} can be excessively stochastic, leading to divergence in the training of the delayed target model. For instance, based on Equations~\eqref{eq:poissdefs1},~\eqref{eq:poissdefs2},~\eqref{eq:poissdefs3},~\eqref{eq:poissdefs4}, and~\eqref{eq:poissdefs5} for the Poisson problem, we can write $g_\theta(x_i) = (M-1) f_\theta(x_i)$. Therefore, we can analyze the target variance as
\begin{align}\label{eq:targetvar}
\VV \big[&\geatrg+y(x) \mid x\big] = \nonumber\\
&\frac{(M-1)^2}{N} \VV[f_{\theta^\trg}(x')\mid x] + \VV[y(x)\mid x].
\end{align}

Ideally, $M\rightarrow \infty$ in order for Equation~\eqref{eq:reimansum} to hold. Setting arbitrarily large $M$ will lead to unbounded target variances in Equation~\eqref{eq:targetvar}, which can slow down the convergence of the training or result in divergence. In particular, such unbounded variances can cause the main and the target models to drift away from each other, leading to incorrect solutions as we will show in Figure~\ref{fig:divergedbstrap} for example.

To prevent this drift, one technique is to impose a Bayesian prior on the main and the target models. Therefore, to discourage this divergence phenomenon, we regularize the delayed target objective in Equation~\eqref{eq:divthmbsloss} and replace it with
\begin{align}\label{eq:divthmbsloss}
\hat{\LL}_\theta^{\BS\text{R}} := \hat{\LL}^{\BS}_{\theta}(x) + \lambda \cdot (f_\theta(x) - f_{\theta^\trg}(x))^2.
\end{align} 

A formal description of the regularized delayed targeting process is given in Algorithm~\ref{alg:brs}, which covers both the moving target stabilization and the Bayesian prior imposition.

\begin{figure*}[t]
	\centering
	\includegraphics[width=0.59\linewidth]{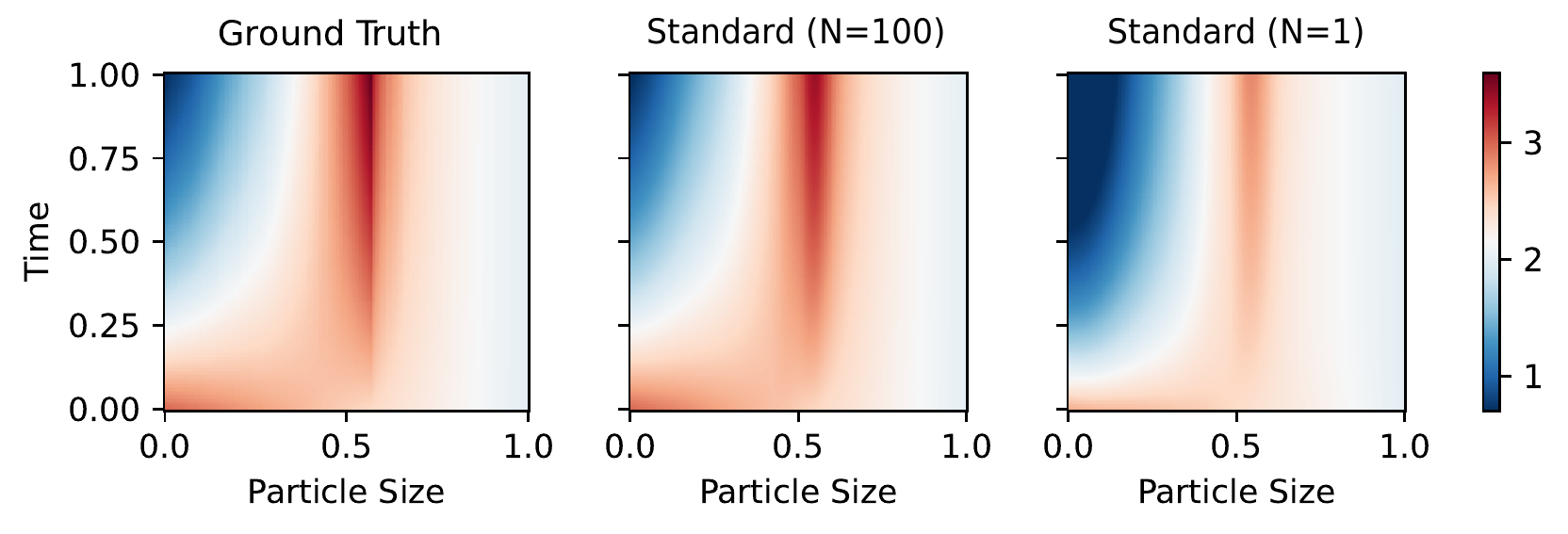}\hspace{-1mm}
	\raisebox{0.1\height}{\includegraphics[width=0.4\linewidth]{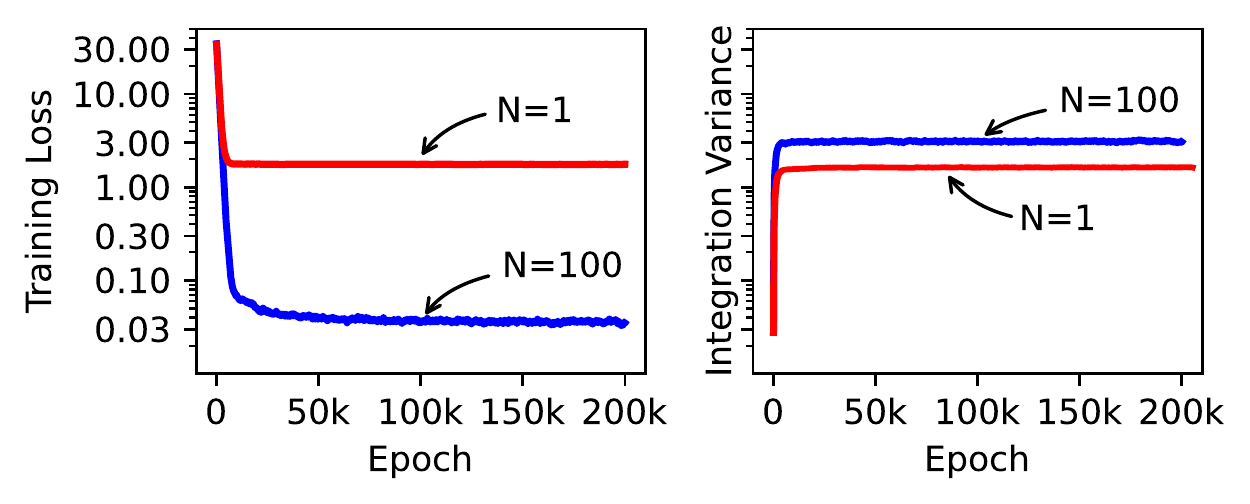}}\\
	\includegraphics[width=0.45\linewidth]{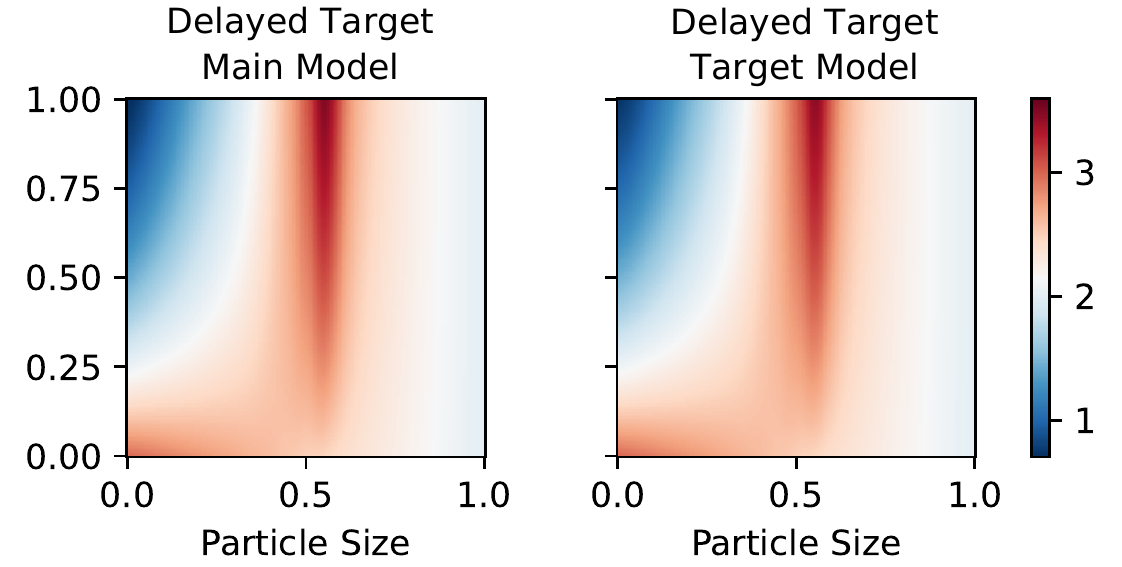}\hspace{25mm}
	\raisebox{0.05\height}{\includegraphics[width=0.23\linewidth]{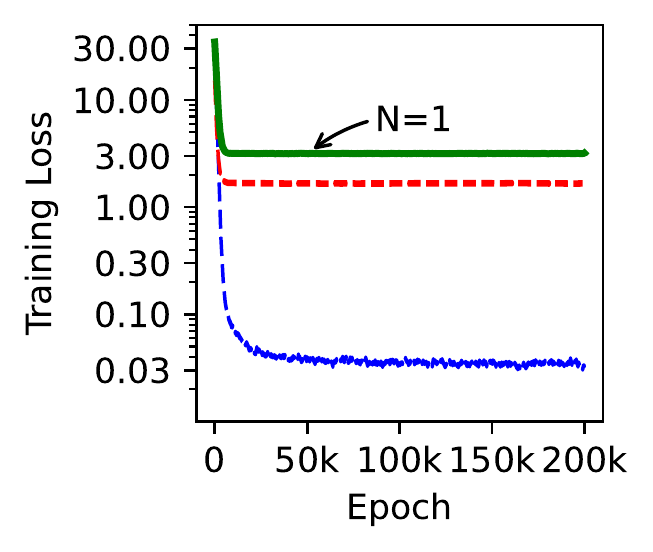}}
	\vspace{-3mm}\caption{Training results on the Smoluchowski coagulation problem. The top left panel shows the ground truth solution, along with the standard $N=100$ and $N=1$ solutions minimizing the $\hat{\LL}_{\theta}(x)$ in Equation~\eqref{eq:excessvar}. The training loss and the integration variance represent the $\hat{\LL}_{\theta}(x)$ and $\VV_{\NSPDF}[g_\theta(x')]$ quantities in Equation~\eqref{eq:excessvar}. The top right figure shows the training curve for both of the standard trainings. The bottom left panel shows the delayed target solution heatmaps using $N=1$ sample with its training curve next to it.}\label{fig:smpanel}\vspace{-1.5mm}
\end{figure*}

\section{Theoretical Results}\label{sec:pinntheorymainpaper}
The double-sampling method and all the deterministic sampling variants use complete gradients for optimization. Thus, they enjoy all the classical convergence guarantees and computational complexity analyses pertaining to the traditional stochastic gradient descent. Essentially, all these methods aim to solve for the fixed point of Equation~\ref{eq:typicalipde}. 

However, the delayed target method is different. In fact, the delayed target method can be presented as an instance of stochastic approximation to solve a slightly different fixed point problem:
\begin{theorem}\label{thm:delayedtargetlinsmry}
	Following the assumptions and notation defined in Section~\ref{sec:pinndefs} of the supplementary material, notably 
	
	(1) a linear function approximation $f_{\theta}(x)=\phi(x)\TRNSPS \theta$, 
	
	(2) appropriate $\eta_t$ learning rates such that $\sum_{t=0}^{\infty} \eta_t = \infty$ and $\sum_{t=0}^{\infty} \eta_t^2 \leq \infty$, 
	
	(3) a small $\tau \rightarrow 0$ with $\lambda=0$ and $N=1$, 
	
	(4) $\UU$ denoting the training update operator, and 
	
	(5) $\PI$ being a projection operator to the function approximation class, 
	
	the delayed target method is an instance of stochastic approximation~\citep{robbins1951stochastic,kiefer1952stochastic} and converges to the fixed point of the $\PI\,\UU$ composite operator in the following equation:
	\begin{equation}
		\Phi\theta^*_{\text{DT}} = \PI\,\UU\SPC \Phi\theta^*_{\text{DT}}.
	\end{equation}
	This is in contrast to the standard training method, which solves for the fixed point of the $\UU$ update operator under the same conditions:
	\begin{equation}
		\Phi\theta^* = \UU\SPC \Phi\theta^*.
	\end{equation}
	Also, assuming that $f^{*}$ is the fixed point to the $\UU$ operator, the approximation error for the delayed target method under these conditions can be upper-bounded as
	\begin{align}\label{eq:dtaproxerrorsmry}
		\mathbb{E}&_{x\sim P}[(f_{\thetapu}(x) - f^{*}(x) )^2] \leq \nonumber\\
		&\frac{1}{1-\sigma_{\mathbf{P}^{|}\Lambda}} \mathbb{E}_{x\sim P} \big[ (\PI\SPC f^{*}(x) - f^{*}(x))^2\big].
	\end{align}
\end{theorem}
For the detailed statement and proof of Theorem~\ref{thm:delayedtargetlinsmry} as well as the rest of the assumptions and notation (e.g., $\Phi$, $\sigma$, $\mathbf{P}^{|}$, $\Lambda$, and $f^*$), see Section~\ref{sec:pinntheory} of the supplementary material. Since the delayed target method is an instance of stochastic approximation, its total computational cost to reach an optimization error of $\epsilon$ in a $d$-dimensional parameter space can be $O(d/\epsilon)$, whereas the standard training method may cost $O(N d \log(1/\epsilon))$ to achieve the same goal. This is discussed further in Section~\ref{sec:dtcompcmplxty} of the supplementary material.

Of course, with a non-linear function approximation class, the delayed target method may not converge to reasonable solutions; Figure~\ref{fig:divergedbstrap} demonstrates such an incorrect solution example. Our setup is more general than reinforcement learning, where such effects have been a topic of research for many decades~\citep{baird1995residual,boyan1995generalization,gordon1995stable,tsitsiklis1997analysis,tsitsiklis1996feature,bertsekas1995counterexample,dayan1992convergence,bertsekas1996neuro}.

\begin{figure*}[t]
	\centering
	\includegraphics[width=0.99\linewidth]{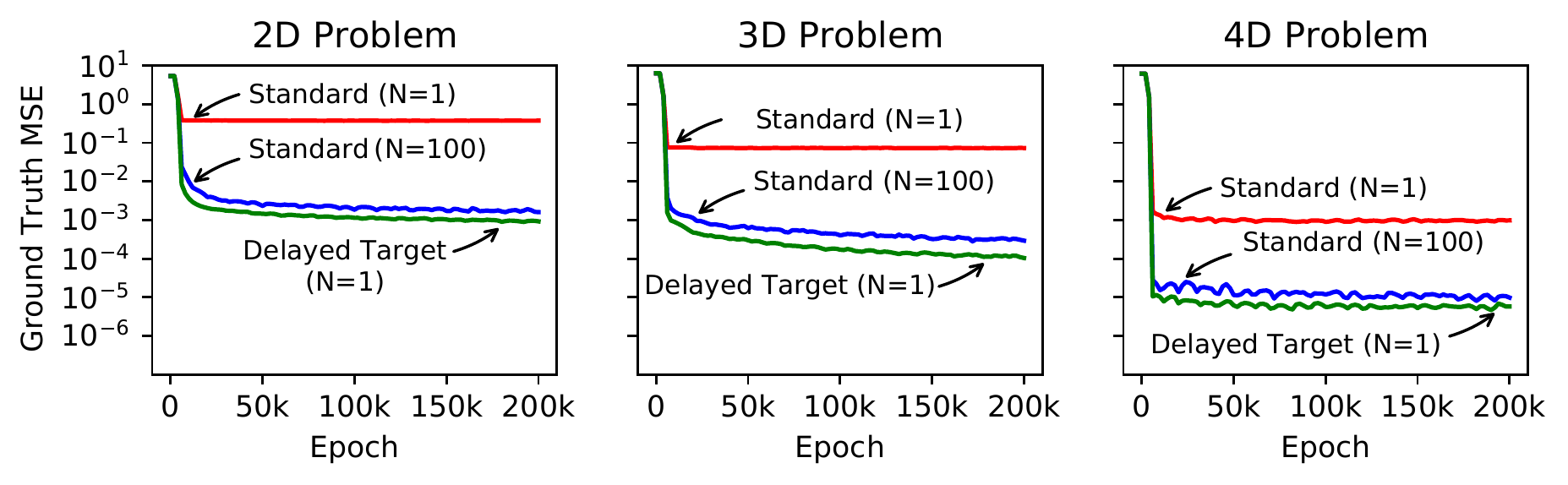}
	\vspace{-3.5mm}\caption{The solution mean squared error to the ground truth in the 2, 3, and 4-dimensional Smoluchowski coagulation problem. The vertical axis shows the solution error, and the horizontal axis shows the training epochs. The standard solutions were trained by the ordinary MSE loss $\LL_{\theta}(x)$ in Equation~\eqref{eq:origmseloss} with $N=1$ and $N=100$ samples. The delayed target solution used $N=1$ sample, yet produced slightly better results than the standard method with $N=100$.}\vspace{-1mm}\label{fig:smlosshidim}
\end{figure*}

\section{Experiments}\label{sec:results}

We examine solving three problems. First, we solve a Poisson problem with singular charges using the divergence theorem as a proxy for learning. In Section~\ref{sec:poissexpsubsec}, we define a 2D Poisson problem with three unit Dirac-delta charges at $[0,0]$, $[-0.5, -0.5]$, and $[0.5, 0.5]$. Figures~\ref{fig:msegt},~\ref{fig:dtsdblpoisson}, and~\ref{fig:bstrapconv} demonstrate the potential solutions to this problem. We also study higher-dimensional Poisson problems with a unit charge at the origin in Figure~\ref{fig:hidimpoiss}. Our second example in Section~\ref{sec:maxwellexpsubsec} looks at finding the magnetic potentials and fields around a current circuit. The current circuit consists of four wire segments and defines a singular $\J$ current density profile. Finally, in Section~\ref{sec:smolexpsubsec} we consider a Smoluchowski coagulation problem to simulate particle evolution dynamics. We designed the coagulation kernel $\SMK$ to induce non-trivial solutions in our solution intervals. 

We employed multi-layer perceptrons as our deep neural networks, using 64 hidden neural units in each layer, and either the SiLU or $\tanh$ activation functions. We trained our networks using the Adam~\citep{kingma2014adam} variant of the stochastic gradient descent algorithm under a learning rate of $0.001$. We afforded each method 1000 point evaluations for each epoch. A wealth of ablation studies with more datasets and other experimental details were left to Section~\ref{sec:pinnablations} of the supplementary material.

\subsection{The Poisson Problem with Singular Charges}\label{sec:poissexpsubsec}
To show the solution bias, we first train two models: one with $N=100$ samples per sphere, and another one with only $N=1$ sample per sphere. These models represent a baseline for later comparisons. Based on Equation~\eqref{eq:excessvar}, the induced solution bias should be lower in the former scenario. Figure~\ref{fig:msegt} shows the solution defined by these models along with the analytical solution and their respective training curves. The model trained with high estimation variance derives an overly smooth solution. We hypothesize that this is due to the excess variance in the loss. This hypothesis is confirmed by matching the training loss and the excess variance curves; the training loss of the model with $N=1$ is lower bounded by its excess variance, although it successfully finds a solution with a smaller excess variance than the $N=100$ model. An alternative capable of producing similar quality solutions with $N=1$ sample would be ideal.

To investigate the effect of highly stochastic targets on delayed target models, Figure~\ref{fig:divergedbstrap} shows the training results with both $M=100$ and $M=10$. The former is unstable, while the latter is stable; this confirms the influence of $M$ in the convergence of the delayed target trainings. Furthermore, when this divergence happens, a clear drift between the main and the target models can be observed. Figure~\eqref{fig:bstrapconv} shows that imposing the Bayesian prior of Equation~\eqref{eq:divthmbsloss} can lead to training convergence even with a larger $M=1000$, which demonstrates the utility of our proposed solution.

We also investigated the performance of the deterministic and double-sampling techniques in this problem. Figure~\ref{fig:dtsdblpoisson} shows these results when $N=1$ and $N=100$ samples are used for integral estimation. With $N=1$, the training with the deterministic sampling approach is stable and yields similar results to those seen in Figure~\eqref{fig:msegt}. The double-sampling trick, on the other hand, exhibits unstable trainings and sub-optimal solutions. We suspect that (a) the singular nature of the analytical solution, and (b) the stochasticity profile of the training loss function $\hat{\LL}^{\dbl}_{\theta}(x)$ in Equation~\eqref{eq:dblsamp} are two of the major factors contributing to this outcome. With $N=100$, both the deterministic and double-sampling trainings yield stable training curves and better solutions. This suggests that both methods can still be considered viable options for training integro-differential PINNs, conditioned on that the specified $N$ is large enough for these methods to train stably and well.

The regularized delayed target training with $N=1$ sample is also shown in the training curves of Figure~\ref{fig:dtsdblpoisson} for easier comparison. The delayed target method yields better performance than the deterministic or double-sampling in this problem. This may seemingly contradict the fact that the double-sampling method enjoys better theoretical guarantees than the delayed target method since it optimizes a complete gradient. However, our results are consistent with recent findings in off-policy reinforcement learning; even in deterministic environments where the application of the double-sampling method can be facilitated with a single sample, incomplete gradient methods (e.g., TD-learning) may still be preferable over the full gradient methods (e.g., double-sampling)~\citep{saleh2019deterministic,fujimoto2022should,yin2022experimental,chen2021instrumental}. Intuitively, incomplete gradient methods detach parts of the gradient, depriving the optimizer from exercising full control over the decent direction and make it avoid over-fitting. In other words, incomplete gradient methods can be viewed as a middle ground between zero-order and first-order optimization and may be preferable over both of them.

Figure~\ref{fig:hidimpoiss} also studies the effect of problem dimensionality on our methods. The results confirm that the problem becomes significantly more difficult with higher dimensions. However, the delayed target solutions maintain comparable quality to standard trainings with large $N$. Gaussian and Leja numerical quadrature seem to be less effective in this problem. QMC methods, on the other hand, certainly improve upon the standard i.i.d.\ estimators. The delayed target with $N=1$ performs similarly to the standard and QMC methods with $N=100$, and can be improved further by increasing $N$ (see Section~\ref{sec:dtsampsizeabls} of the supplementary material on scaling up $N$ effectively in the delayed target method).

\subsection{The Maxwell Problem with a Wired Circuit}\label{sec:maxwellexpsubsec}
Figure~\ref{fig:maxwellmain} shows the training results for the Maxwell problem. The results suggest that the standard and the deterministic trainings with small $N$ produce overly smooth solutions. The double-sampling method with small $N$ improves the solution quality at first but has difficulty maintaining a stable improvement. However, delayed targeting with small $N$ seems to produce comparable solutions to the standard training with large $N$. 

\subsection{The Smoluchowski Coagulation Problem}\label{sec:smolexpsubsec}
Figure~\ref{fig:smpanel} shows the training results for the Smoluchowski coagulation problem. Similar to the results in Figure~\ref{fig:msegt}, the standard training using $N=1$ sample for computing the residual summations leads to a biased and sub-optimal solution. However, the standard training with $N=100$ samples suffers less from the effect of bias. The delayed target solution using only $N=1$ sample produces comparable solution quality to the standard evaluation with $N=100$ and is not bottlenecked by the integration variance. Figure~\ref{fig:smlosshidim} compares the solution quality for each of the standard and delayed target methods under different problem dimensions. The results suggest that the delayed target solution maintains its quality even in higher dimensional problems, where the excess variance issue leading to biased solutions may be more pronounced.

\vspace{-3mm}\section{Discussion}
In this work, we investigated the problem of learning PINNs in partial integro-differential equations. We presented a general framework for the problem of learning from integral losses and theoretically showed that naive approximations of the parametrized integrals lead to biased loss functions due to the induced excess variance term in the optimization objective. We confirmed the existence of this issue in numerical simulations. Then, we studied three potential solutions to account for this issue, and we found the delayed target method to perform best in a wide class of problems. Our numerical results support the utility of this method on three classes of problems, (1) Poisson problems with singular charges and up to 10 dimensions, (2) an electromagnetic problem under a Maxwell equation, and (3) a Smoluchowski coagulation problem. The limitations of our work include its narrow scope in learning PINNs; this work could have broader applications in other areas of machine learning. Also, future work should consider the applications of the delayed target method to more problem classes in both scientific and traditional machine learning. Developing adaptive processes for setting each method's hyper-parameters, such as the training batch-sizes and regularization weights in the delayed target method, and combining importance sampling, numerical quadrature, or QMC techniques with our methods are two other worthwhile future endeavors.

\section*{Acknowledgements}

This work used GPU resources at the Delta supercomputer of the National Center for Supercomputing Applications through Allocation CIS220111 from the Advanced Cyberinfrastructure Coordination Ecosystem: Services and Support (ACCESS) program~\citep{boerner2023access}, which is supported by National Science Foundation grants \#2138259, \#2138286, \#2138307, \#2137603, and \#2138296.

\section*{Impact Statement}

This work provides foundational theoretical results and builds upon methods for training neural PDE solvers within the area of scientific learning. Scientific learning methods and neural PDE solvers can provide valuable models for a solving range of challenging applications in additive manufacturing~\citep{zhu2021machine,niaki2021physics,henkes2022physics}, robotics~\citep{sun2022physics}, high-speed flows~\citep{mao2020physics}, weather-forecasting~\citep{mammedov2021weather}, finance systems~\citep{bai2022application} chemistry~\citep{ji2021stiff}, computational biology~\citep{lagergren2020biologically}, and heat transfer and thermodynamics~\citep{cai2021physics}. 

Although many implications could result from the application of scientific learning, in this work we focused especially on settings where precision, singular inputs, and compatibility with partial observations are required for solving the PDEs. Our work particularly investigated methods for learning PDEs with integral forms and provided effective solutions for solving them. Such improvements could help democratize the usage of physics-informed networks in applications where independent observations are difficult or expensive to obtain, and the inter-sample relationships and constraints may contain the majority of the training information. Such problems may be challenging and the trained models are usually less precise than the traditional solvers. These errors can propagate to any downstream analysis and decision-making processes and result in significant issues. Other negative consequences of this work could include weak interpretability of the trained models, increased costs for re-training the models given varying inputs, difficulty in estimating the performance of such trained models, and the existence of unforeseen artifacts in the trained models~\citep{wang2021understanding}.

To mitigate the risks, we encourage further research to develop methods to provide guarantees and definitive answers about model behaviors. In other words, a general framework for making guaranteed statements about the behavior of the trained models is missing. Furthermore, more efficient methods for training such models on a large variety of inputs should be prioritized for research. Also, a better understanding of the pathology of neural solvers is of paramount concern to use these models safely and effectively.

\bibliography{refs_bspinn}
\bibliographystyle{icml2024}

\newpage
\appendix
\onecolumn


\section{Probabilistic and Mathematical Notation}\label{sec:pinnnotation} 
\newcommand{\zz}{z}
We denote expectations with the $\EE_{P(\zz)}[h(\zz)]:=\int_{\zz} h(\zz)P(\zz)\diff \zz$, and variances with the $\VV_{P(\zz)}[h(\zz)]:=\EE_{P(\zz)}[h(\zz)^2] - \EE_{P(\zz)}[h(\zz)]^2$ notation. Note that only the random variable in the subscript (i.e., $\zz$) is eliminated after the expectation. The set of samples $\{x'_1, \cdots x'_n\}$ is denoted with $\xpi$, and we abuse the notation by replacing $\xpi$ with $x'_{1:n}$ for brevity. Throughout the manuscript, $f_{\theta}(x)$ denotes the output of a neural network, parameterized by $\theta$, on the input $x$. The loss functions used for minimization are denoted with the $\LL$ notation (e.g., $\LL_\theta(x)$). $\nabla U:= [\frac{\partial}{\partial x_1} U, \cdots, \frac{\partial}{\partial x_d} U]$ denotes the gradient of a scalar function $U$, $\nabla \cdot E:=\frac{\partial E_1}{\partial x_1} + \cdots + \frac{\partial E_d}{\partial x_d}$ denotes the divergence of the vector field $E$, and $\nabla^2 U:=\nabla\cdot \nabla U$ denotes the Laplacian of the function $U$. The $d$-dimensional Dirac-delta function is denoted with $\delta^{d}$, volumes are denoted with $\vol$, and surfaces are denoted with $\sur$. The Gamma function is denoted with $\Gamma$, where $\Gamma(n):=(n-1)!$ for integer $n$. The uniform probability distribution over an area $A$ is denoted with $\text{Unif}(A)$. These operators and notation are summarized in Tables~\ref{tab:mathnotation} and~\ref{tab:mathoperators}.

\renewcommand{\arraystretch}{1.6}
\begin{table}[h]
	\centering
	\aboverulesep=0ex
	\belowrulesep=0ex
	\begin{tabular}{|p{0.15\textwidth}|p{0.80\textwidth}|}
		\toprule
		Notation & Description \\ \midrule\midrule
		$f_{\theta}(x)$ & The main neural output parameterized by $\theta$ \\\hline
		$g_{\theta}(x)$ & Secondary neural output parameterized by $\theta$ \\\hline
		$\LL$ & Generic loss functions representation \\\hline
		$\LL_\theta(x) $ & Loss $\LL$ parametrized by $\theta$ evaluated at $x$ \\\hline
		$\hat{\LL}$ & Generic approximated loss representation \\\hline
		$N$ & Number of samples used for integral estimation \\\hline
		$\tau$ & The delayed target Polyak averaging factor in Algorithm~\ref{alg:brs} of the main paper\\\hline
		$\lambda$ & The delayed target regularization weight defined in Equation~\eqref{eq:divthmbsloss} \\\hline\hline
		$\delta^{d}$ & The $d$-dimensional Dirac-delta function \\\hline
		$\vol$ & Volume representation \\\hline
		$\sur$ & Surface representation \\\hline
		$\Gamma$ & The Gamma function, where $\Gamma(n):=(n-1)!$ for integer values\\\hline
		$\text{Unif}(Z)$ & The uniform probability distribution over the $Z$ set or interval\\\hline
		$\UPOIS$ & The potential function in the Poisson problem \\\hline
		$\EPOIS$ & The gradient field in the Poisson problem \\\hline
		$\rhopois$ & The input charge density in the Poisson problem \\\hline\hline
		$\A$ & The magnetic potentials in the Maxwell-Ampere problem \\\hline
		$\B$ & The magnetic field in the Maxwell-Ampere problem \\\hline
		$\J$ & The current density field in the Maxwell-Ampere problem \\\hline
		$\I$ & The current flowing through a plane in the Maxwell-Ampere problem \\\hline\hline
		$\SMK$ & The Smoluchowski coagulation kernel used in Equation~\eqref{eq:smoll} \\\hline
		$\smrho$ & Particle densities in the Smoluchowski equation \\\hline
	\end{tabular}	
	\vspace{0mm}\caption{The mathematical notation used throughout the paper.}
	\label{tab:mathnotation}
\end{table}

\renewcommand{\arraystretch}{1.6}
\begin{table}[h]
	\centering
	\aboverulesep=0ex
	\belowrulesep=0ex
	\begin{tabular}{|p{0.11\textwidth}|p{0.44\textwidth}|p{0.35\textwidth}|}
		\toprule
		Notation & Definition & Description \\ \midrule\midrule
		$\nabla \cdot \EPOIS$ & $\frac{\partial \EPOIS_1}{\partial x_1} + \cdots + \frac{\partial \EPOIS_d}{\partial x_d}$ & Divergence of the $\EPOIS$ field \vspace{2mm}\\\hline
		$\nabla^2 \UPOIS$ & $\nabla\cdot \nabla \UPOIS$ & Laplacian of the $\UPOIS$ potential  \vspace{1mm}\\\hline
		$\nabla \times \A$ & $\big[ \frac{\partial \A_3}{\partial x_2} - \frac{\partial \A_2}{\partial x_3},\quad \frac{\partial \A_1}{\partial x_3} - \frac{\partial \A_3}{\partial x_1},\quad \frac{\partial \A_2}{\partial x_1} - \frac{\partial \A_1}{\partial x_2}\big]^{\TRNSPS}$ & Curl of the 3D $\A$ field \vspace{2mm}\\\hline
	\end{tabular}	
	\vspace{0mm}\caption{The differential operators used throughout the paper.}
	\label{tab:mathoperators}
\end{table}

\section{Theoretical Results}\label{sec:pinntheory}

Given the empirical performance of the delayed target method, some supporting theoretical results may provide more insight into this method. Here, under a linear function approximation class and certain assumptions described in Section~\ref{sec:pinndefs}, we present the delayed target method as an instance of stochastic approximation to solve a fixed point problem and upper-bound its approximation error in Section~\ref{sec:pinntheorem}. We also compare the computational complexity of the delayed target and standard training methods in Section~\ref{sec:dtcompcmplxty}.

\vspace{0mm}\subsection{Definitions and Assumptions}\label{sec:pinndefs}

As stated earlier, we will assume a linear function approximation class:
\begin{align} 
f_{\theta}(x) &:= \phi(x)\TRNSPS \theta,\label{eq:linearf}\\
g_{\theta}(x) &:= \psi(x)\TRNSPS \theta.\label{eq:linearg}
\end{align}

For convenience in the theoretical derivations, we will assume that the $x$ domain is discretized into $\SS$ bins. This is commonly known as an abstraction, where a mapping is applied to compress the original continuous input domain $\mathcal{X}$ into some finite abstract space~\citep{li2006towards}:
\begin{equation}\label{eq:abstractX}
x\in \mathcal{X} \cong \{x_1, x_2, \cdots, x_\SS\}.
\end{equation}
Under this regime, Certainty Equivalence (CE) models were studied to potentially improve the generalization of functions learned within the abstract space~\citep{givan2003equivalence,ravindran2004algebraic,li2009unifying,jong2005state,jiang2015abstraction}. Here, we only use an abstraction to better understand the role of the input domain size and make the theoretical derivations easier to follow. We assume $\SS \rightarrow \infty$, that is, $\SS$ is an infinitely large integer. To be clear, this has no practical impact on our algorithms, as they run in the original continuous domain. We are mainly abstracting the input domain to express the terms in matrix and vector product forms rather than integrals. 

Next, we define the following notation:
\begin{itemize}
\item $d$ denotes the parameter dimensions:
\begin{equation}\label{eq:thetadimdef}
    d := \text{dim}(\theta).
\end{equation}
\item $\Phi$ denotes the feature matrix of the $f$ function: 
\begin{equation}\label{eq:Phidef}
    \Phi := [\phi(x_1), \phi(x_2), \cdots, \phi(x_{\SS})]\TRNSPS \in \mathbb{R}^{\SS \times d}. 
\end{equation}
\item $\Psi$ denotes the feature matrix of the $g$ function: 
\begin{equation}\label{eq:Psidef}
    \Psi := [\psi(x_1), \psi(x_2), \cdots, \psi(x_{\SS})]\TRNSPS \in \mathbb{R}^{\SS \times d}.
\end{equation}
\item $\mathbf{D}_{P}$ denotes the diagonal matrix consisting of the input sampling probabilities: 
\begin{equation}\label{eq:Dpdef}
    \mathbf{D}_{P} := \text{diag}([P(x_1), P(x_2), \cdots, P(x_{\SS})]) \in \mathbb{R}^{\SS \times \SS}.
\end{equation}
\item $\mathbf{Y}$ denotes the compiled labels for each input:
\begin{equation}\label{eq:Ydef}
    \mathbf{Y} := [y(x_1), y(x_2), \cdots, y(x_{\SS})] \in \mathbb{R}^{\SS}.
\end{equation}
\item $\mathbf{P}^{|}$ denotes the conditional sampling distribution in a matrix form:
\begin{equation}\label{eq:Pconddef}
    \mathbf{P}^{|} := [P(x'=x_i|x=x_j)]_{i,j} \in \mathbb{R}^{\SS \times \SS}.
\end{equation}
\item Assuming $M$ is a square matrix, $\sigma_{M}$ denotes the spectral radius of $M$:
\begin{equation}\label{eq:specraddef}
    \sigma_{M} := \max_{\|z\|_2=1} |z\TRNSPS M z|.
\end{equation}
\item We denote $f^*$ and $g^*$ to be the perfect solution to Equation~\ref{eq:typicalipde}. In the vector form over the abstract space, they can be represented as $\mathbf{f}^{*}$ and $\mathbf{g}^{*}$, respectively:
\begin{align}
\mathbf{f}^{*} := [f^{*}(x_1), \cdots, f^*(x_{\SS})] \in \mathbb{R}^{\SS},\label{eq:fstardef}\\
\mathbf{g}^{*} := [g^{*}(x_1), \cdots, g^*(x_{\SS})] \in \mathbb{R}^{\SS}\label{eq:gstardef}.
\end{align}
\item Continuous functions can be expressed as vectors in the abstracted space. We denote $h$ for an arbitrary function, and represent it as $\mathbf{H}$ in the vector form over the abstract space:
\begin{equation}\label{eq:Hdef}
\mathbf{H} := [h(x_1), h(x_2), \cdots, h(x_N)]\TRNSPS \in \mathbb{R}^{\SS}.
\end{equation}
\item The weighted L2-norm for $\mathbf{H}$ under the $P$ distribution can be defined as
\begin{equation}\label{eq:Pnormdef}
\|\mathbf{H}\|_{P} := \mathbb{E}_{x\sim P}[h(x)^2].
\end{equation}
\end{itemize}

\renewcommand{\arraystretch}{1.65}
\begin{table}[t]
	\centering
	\aboverulesep=0ex
	\belowrulesep=0ex
	\begin{tabular}{|p{0.105\linewidth}|p{0.13\linewidth}|p{0.67\linewidth}|}
		\toprule
		Notation & Domain & Description \\ \midrule\midrule
        $\SS$ & $\mathbb{R}$ & The abstracted input domain size\\\hline
        $d$ & $\mathbb{R}$ & The parameter dimension \\\hline
        $\theta$ & $\mathbb{R}^{d}$ & The learned parameters \\\hline
		$\phi(x)$ & $\mathbb{R}^{d}$ & The input feature representations within $f_{\theta}$ \\\hline
        $\psi(x)$ & $\mathbb{R}^{d}$ & The input feature representations within $g_{\theta}$\\\hline
        $\Phi$ & $\mathbb{R}^{\SS \times d}$ & The compiled matrix of $\phi$ features \\\hline
        $\Psi$ & $\mathbb{R}^{\SS \times d}$ & The compiled matrix of $\psi$ features \\\hline
        $\mathbf{D}_{P}$ & $\mathbb{R}^{\SS \times \SS}$ & The diagonal matrix of the $P$ sampling probabilities \\\hline
        $\mathbf{Y}$ & $\mathbb{R}^{\SS}$ & The non-parametric labels \\\hline
        $\mathbf{P}^{|}$ & $\mathbb{R}^{\SS \times \SS}$ & The $P(x'|x)$ conditional distribution in matrix form \\\hline
        $\sigma_{M}$& $\mathbb{R}_{0}^{+}$ & The spectral radius of Matrix $M$ \\\hline
		$\mathbf{f}^{*}$ & $\mathbb{R}^{\SS}$ & The perfect $f$ solution to Equation~\eqref{eq:typicalipde} \\\hline
        $\mathbf{g}^{*}$ & $\mathbb{R}^{\SS}$ & The perfect $g$ solution to Equation~\eqref{eq:typicalipde}\\\hline
        $\Lambda$ & $\mathbb{R}^{\SS \times \SS}$ & The matrix relating the $\Phi$ and $\Psi$ features \\\hline
        $\PI\SPC \mathbf{H}$ & $\mathbb{R}^{\SS}$ & The projection operator to $\text{span}(\Phi)$ \\\hline
        $\UU\SPC \mathbf{H}$ & $\mathbb{R}^{\SS}$ & The update operator to solve Equation~\eqref{eq:typicalipde} \\\hline
		$\|\mathbf{H}\|_{P}$ & $\mathbb{R}_{0}^{+}$ & The L2-norm weighted by the $P$ distribution \\\hline
	\end{tabular}	
	\vspace{1mm}\caption{The notation used in the theoretical analyses of the delayed target method.}
	\label{tab:pinntheorynotation}
\end{table}
\renewcommand{\arraystretch}{1.5}

We will assume distinctive and bounded features and non-zero probability for sampling all input values: 
\begin{equation}
\forall x\in \mathcal{X}: \|\phi(x)\|\leq 1, \quad \|\psi(x)\| \leq 1, \quad P(x) > 0,
\end{equation}
\begin{equation}
\text{rank}(\Phi) = \text{rank}(\Psi) = d.
\end{equation}

Since $\SS >d$, for some $\Lambda \in \mathbb{R}^{\SS \times \SS}$ we have
\vspace{-1mm}\begin{equation}\label{eq:Lambdadef}
    \Psi = \Lambda \Phi.
\end{equation}
Note that $\Lambda$ can be constructed by a simple linear regression of the rows. Since $\SS$ is much larger than $d$, this is an over-parameterized setting and such $\Lambda$ will always exist.

We will assume that the $\Lambda$ matrix has a sub-unit spectral radius. In other words, we assume $\sigma_{\mathbf{P}^{|}\Lambda} < 1$ for all $\SS \geq \SS_{\min}$, where $\SS_{\min}$ is a constant. This practically means that \textit{$g$ is not ``over-powering'' $f$} and that \textit{$f$ has the most control over the delayed target updates}. For instance, the Bellman equation with a $\gamma<1$ discount factor
\begin{equation}
	V^{\pi}_{\theta}(x) = \gamma \mathbb{E}_{P(x'|x)}[V^{\pi}_{\theta}(x')] + R(x,\pi)
\end{equation} 
satisfies this condition, since it has $\Lambda = \gamma I$ and $\sigma_{\mathbf{P}^{|}\Lambda} = \gamma \sigma_{\mathbf{P}^{|}} < 1$.

Next, we define the projection and update operators:
\begin{align}
\PI\SPC \mathbf{H} &:= \argmin_{z=\Phi\theta} (\mathbf{H}-\Phi\theta)\TRNSPS \mathbf{D}_{P} (\mathbf{H}-\Phi\theta),\label{eq:PIdef}\\
\UU\SPC \mathbf{H} &:= \mathbf{Y} + \mathbf{P}^{|} \Lambda \mathbf{H}.\label{eq:UUdef}
\end{align}
We can abuse the notation, and express these operators in the original domain as well:
\begin{equation}
    \PI\SPC h := \argmin_{\substack{z_{\theta'}\\ \text{s.t.}\;\;\forall x:\;z_{\theta'}(x)=\phi(x)\TRNSPS \theta'}} \mathbb{E}_{x\sim P}[(h(x)-z_{\theta'}(x))^2],
\end{equation}
\begin{equation}\label{eq:uucontdef}
	\UU\SPC h := \mathbb{E}_{P(x'|\cdot)}\bigg[\int \Lambda_{x',x''} h(x'') \diff x''\bigg] + y(\cdot). 
\end{equation}
Notice that
\begin{equation}
    \UU\SPC \Phi\theta = \mathbf{Y} + \mathbf{P}^{|} \Psi \theta.
\end{equation}
Therefore, under these assumptions and notation, solving the original system defined by Equation~\eqref{eq:typicalipde} can be re-stated as finding the fixed point solution to the update operator:
\begin{equation}
    \Phi\theta = \UU\SPC \Phi\theta \quad\Longleftrightarrow\quad \forall x\in \mathcal{X}: f_\theta(x) = \mathbb{E}_{\NSPDF}[g_\theta(x')] + y(x).
\end{equation}
Table~\ref{tab:pinntheorynotation} summarizes this notation.

\subsection{Theoretical Analysis}\label{sec:pinntheorem}
Here, we restate Theorem~\ref{thm:delayedtargetlinsmry} as our main theoretical result. This theorem is a generalization of the ideas originally described in~\citet{tsitsiklis1997analysis}.

{\renewcommand{\thetheorem}{\ref{thm:delayedtargetlinsmry}}
\begin{theorem}\label{thm:delayedtargetlin}
Following the assumptions and notation in Section~\ref{sec:pinndefs}, given learning rates that vanish neither too fast nor too slow, i.e.
\begin{equation}
    \sum_{t=0}^{\infty} \eta_t = \infty, \qquad \sum_{t=0}^{\infty} \eta_t^2 \leq \infty,
\end{equation}
and using a small $\tau \rightarrow 0$ with $\lambda=0$ and $N=1$, the delayed target method is an instance of stochastic approximation~\citep{robbins1951stochastic,kiefer1952stochastic} and converges to the fixed point of the $\PI\,\UU$ composite operator in the following equation:
\begin{equation}
	\Phi\theta^*_{\text{DT}} = \PI\,\UU\SPC \Phi\theta^*_{\text{DT}}.
\end{equation}
Furthermore, assuming that $f^{*}$ is the fixed point to the $\UU$ operator in Equation~\eqref{eq:uucontdef}, the approximation error for the delayed target method under these conditions is upper-bounded as
\begin{equation}\label{eq:dtaproxerror}
    \mathbb{E}_{x\sim P} [(f_{\thetapu}(x) - f^{*}(x) )^2] \leq \frac{1}{1-\sigma_{\mathbf{P}^{|}\Lambda}} \mathbb{E}_{x\sim P} \big[(\PI\SPC f^{*}(x) - f^{*}(x))^2\big].
\end{equation}
\end{theorem}
\addtocounter{theorem}{-1}}

\begin{corollary}
Under the conditions stated earlier and a realizability assumption over the function approximation class, that is, by having the perfect solution be a member of the function class: 
\begin{equation}
\exists \theta \;\forall x: f^{*}(x) = \phi(x)\TRNSPS \theta,
\end{equation}
delayed targeting has no approximation error. That is, both sides of Inequality~\eqref{eq:dtaproxerror} are zero under a realizability assumption.
\end{corollary}

Moreover, due to the stochastic approximation properties, to reach an optimization error of $\epsilon$, the total computational cost for the delayed target method can be $O(d/\epsilon)$, whereas the standard training with $N$ samples can take $O(N d \log(1/\epsilon))$ to achieve the same goal. Section~\ref{sec:dtcompcmplxty} will discuss this computational complexity analysis in more detail.

\begin{proof}
Given $d<\SS$ and the Moore-Penrose pseudo-inverse, we have
\begin{equation}
    \PI\SPC \mathbf{H} = \Phi (\Phi\TRNSPS \mathbf{D}_{P} \Phi)^{-1} \Phi\TRNSPS \mathbf{D}_{P} \mathbf{H}.
\end{equation}
Consider the following Project-Update (PU) equation:
\begin{equation}
    \Phi \theta = \PI\,\UU\SPC \Phi\theta.
\end{equation}
The least-square solution to the PU equation, namely $\thetapu$, satisfies the following:
\begin{equation}
    \thetapu = (\Phi\TRNSPS \mathbf{D}_{P} \Phi)^{-1} \Phi\TRNSPS \mathbf{D}_{P} (\mathbf{Y} + \mathbf{P}^{|} \Lambda \Phi \thetapu).
\end{equation}
Therefore, we have
\begin{equation}
    (\Phi\TRNSPS \mathbf{D}_{P} \Phi) \thetapu = \Phi\TRNSPS (\mathbf{Y} + \mathbf{P}^{|} \Lambda \Phi \thetapu).
\end{equation}
Rearranging this will give us
\begin{equation}
    \Phi\TRNSPS \mathbf{D}_{P} (I - \mathbf{P}^{|} \Lambda) \Phi \thetapu = \Phi\TRNSPS \mathbf{D}_{P} \mathbf{Y}.
\end{equation}
This is essentially to say $\thetapu$ satisfies the $A\theta=b$ system, where
\begin{align}
    \mathbf{A} &:= \Phi\TRNSPS \mathbf{D}_{P} (I - \mathbf{P}^{|} \Lambda) \Phi, \\
    \mathbf{b} &:= \Phi\TRNSPS \mathbf{D}_{P} \mathbf{Y}.
\end{align}
For a small $\tau$, the Polyak averaging will produce almost identical $\theta$ and $\theta_{\text{Target}}$:
\begin{equation}
    \lim_{\tau\rightarrow 0} \theta_{\text{Target}}^{(t)} = \theta^{(t)}.
\end{equation}
Under linear function approximation, the delayed target update at Iteration $t$ simplifies to
\begin{equation}
    \theta_{t+1} \leftarrow \theta_t - \eta_t \cdot (A_t \theta_t - b_t),
\end{equation}
where we have
\begin{align}
    A_t &:= \phi(x) (\phi(x) - \psi(x')) \in \mathbb{R}^{d\times d}, \\
    b_t &:= \phi(x) y(x) \in \mathbb{R}^{d}.
\end{align}
It is fairly straightforward to derive the expectation of $A_t$ and $b_t$ as
\begin{align}
    \mathbb{E}_{x\sim P,x'\sim P(\cdot|x)}[A_t] &= \Phi\TRNSPS \mathbf{D}_{P} (I - \mathbf{P}^{|} \Lambda) \Phi = \mathbf{A}, \\
    \mathbb{E}_{x\sim P,x'\sim P(\cdot|x)}[b_t] &= \Phi\TRNSPS \mathbf{D}_{P} \mathbf{Y} = \mathbf{b}.
\end{align}
In other words, the delayed target method is an instance of stochastic approximation; the delayed target method is applying stochastic gradient descent to minimize $(\mathbf{A}\theta - \mathbf{b})\TRNSPS (\mathbf{A}\theta - \mathbf{b})$.

To upper-bound of the delayed target's approximation error, we can write
\begin{align}
    \| \Phi \thetapu - \mathbf{f}^{*} \|_{P} &\leq \| \Phi \thetapu - \PI\SPC \mathbf{f}^{*} \|_{P} + \| \PI\SPC \mathbf{f}^{*} - \mathbf{f}^{*} \|_{P} \label{eq:dtaproxineq1}\\
    & \leq \| \PI\,\UU\SPC \Phi \thetapu - \PI\SPC \mathbf{f}^{*} \|_{P} + \| \PI\SPC \mathbf{f}^{*} - \mathbf{f}^{*} \|_{P} \label{eq:dtaproxineq2}\\
    &\leq \| \UU\SPC \Phi \thetapu - \mathbf{f}^{*} \|_{P} + \| \PI\SPC \mathbf{f}^{*} - \mathbf{f}^{*} \|_{P} \label{eq:dtaproxineq3}\\
    & \leq \| \UU\SPC \Phi \thetapu - \UU\SPC \mathbf{f}^{*} \|_{P} + \| \UU\SPC \mathbf{f}^{*} - \mathbf{f}^{*} \|_{P} + \| \PI\SPC \mathbf{f}^{*} - \mathbf{f}^{*} \|_{P} \label{eq:dtaproxineq4}\\
    & \leq \sigma_{\mathbf{P}^{|}\Lambda} \| \Phi\thetapu - \mathbf{f}^{*} \|_{P} + \| \UU\SPC \mathbf{f}^{*} - \mathbf{f}^{*} \|_{P} + \| \PI\SPC \mathbf{f}^{*} - \mathbf{f}^{*} \|_{P}. \label{eq:dtaproxineq5}
\end{align}
The first inequality in the chain is a triangle inequality. Inequality~\eqref{eq:dtaproxineq2} holds since $\Phi \thetapu$ is a fixed point for $\PI\,\UU$. Inequality~\eqref{eq:dtaproxineq3} holds because the $\Phi (\Phi\TRNSPS \mathbf{D}_{P} \Phi)^{-1} \Phi\TRNSPS \mathbf{D}_{P}$ matrix is idempotent and square. As a result, its eigenvalues can either be zero or one causing it to be a non-expansion. Inequality~\eqref{eq:dtaproxineq4} is a triangle inequality applied after subtraction and addition of a $\UU\SPC \mathbf{f}^{*}$ term. Finally, Inequality~\eqref{eq:dtaproxineq5} holds because all eigenvalues of $\mathbf{P}^{|}\Lambda$ are upper-bounded by $\sigma_{\mathbf{P}^{|}\Lambda}$ in absolute value.

Assuming $\sigma_{\mathbf{P}^{|}\Lambda} < 1$, we can re-arrange Inequality~\eqref{eq:dtaproxineq5} and write
\vspace{-1mm}\begin{equation}
    \| \Phi \thetapu - \mathbf{f}^{*} \|_{P} \leq \frac{1}{1-\sigma_{\mathbf{P}^{|}\Lambda}} \big( \| \UU\SPC \mathbf{f}^{*} - \mathbf{f}^{*} \|_{P} + \| \PI\SPC \mathbf{f}^{*} - \mathbf{f}^{*} \|_{P} \big).
\end{equation}\vspace{-1mm}
In other words, we have
\vspace{-1mm}\begin{equation}\label{eq:aprxineqfinal}
    \mathbb{E}_{x\sim P} [(f_{\thetapu}(x) - f^{*}(x) )^2] \leq \frac{1}{1-\sigma_{\mathbf{P}^{|}\Lambda}} \mathbb{E}_{x\sim P} \big[(\PI\SPC f^{*}(x) - f^{*}(x))^2 + (\UU\SPC f^{*}(x) - f^{*}(x))^2\big].\vspace{-1mm}
\end{equation}
This essentially upper-bounds the MSE error to the ground truth by a $(1-\sigma_{\mathbf{P}^{|}\Lambda})^{-1}$ coefficient times the projection and update errors of the true $f^*$ to the function approximation class (i.e., the function approximation error). Given the assumption that $f^*$ is the fixed point to the $\UU$ operator, Inequality~\eqref{eq:aprxineqfinal} reduces to the upper-bound stated in the theorem. 
\end{proof}

\begin{figure*}[!t]
	\centering
	\includegraphics[width=0.34\linewidth]{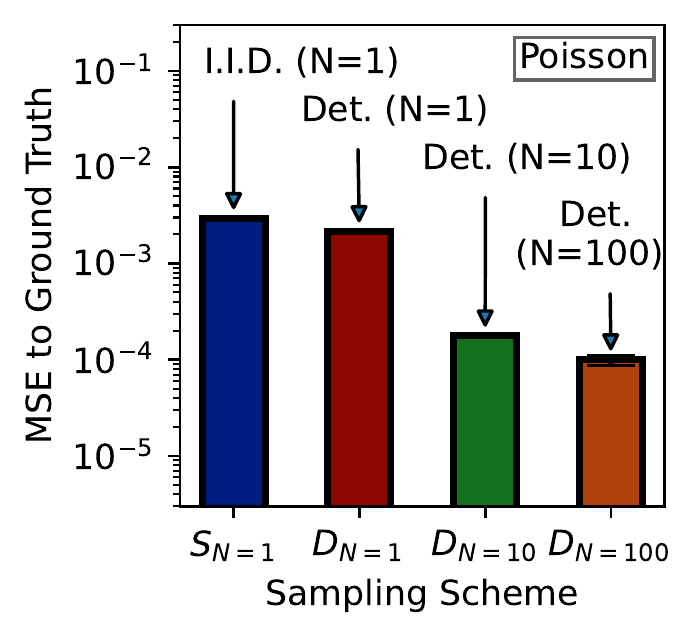}
	\includegraphics[width=0.31\linewidth]{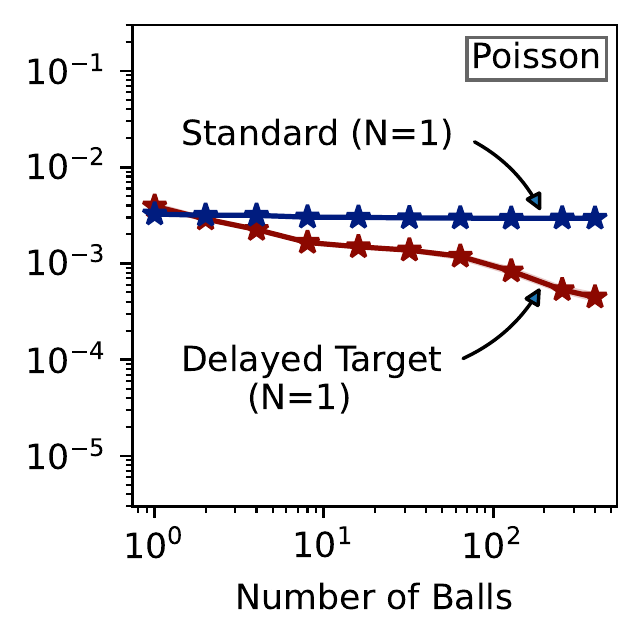}
	\includegraphics[width=0.33\linewidth]{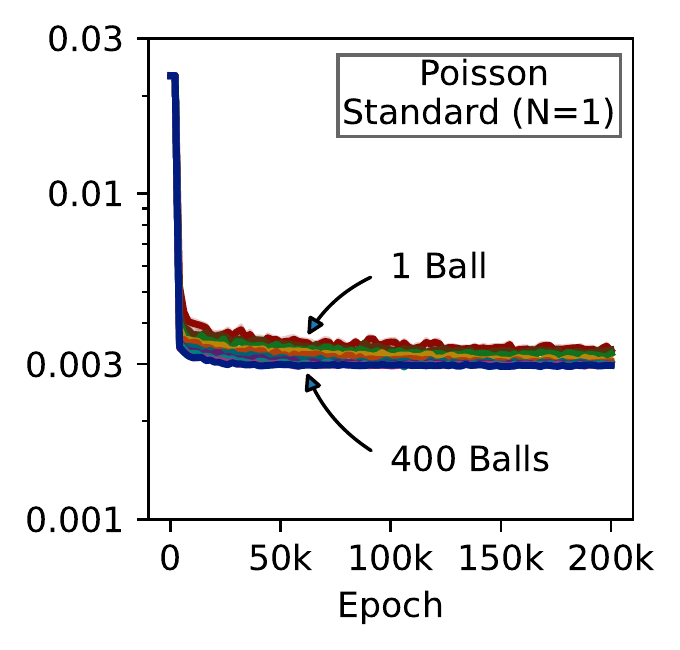}
	\caption{Ablation studies of the sampling hyper-parameters and settings in the 2D Poisson problem of Figure~\ref{fig:msegt} in the main paper. \textit{In the left plot}, we compare the deterministic and i.i.d.\ sampling on the standard trainings with various $N$. \textit{In the middle plot}, the horizontal axis shows the number of balls sampled in each epoch. Both the standard and the delayed target methods are shown in this plot with $N=1$. \textit{The right plot} shows the training curves for the standard method with $N=1$ target samples. Similar ablations for the Maxwell-Ampere and Smoluchowski problems are presented in Figure~\ref{fig:samplingablmaxsmol}.}
	\label{fig:samplingabl}
\end{figure*}

\begin{figure*}[!t]
	\centering
	\includegraphics[width=0.34\linewidth]{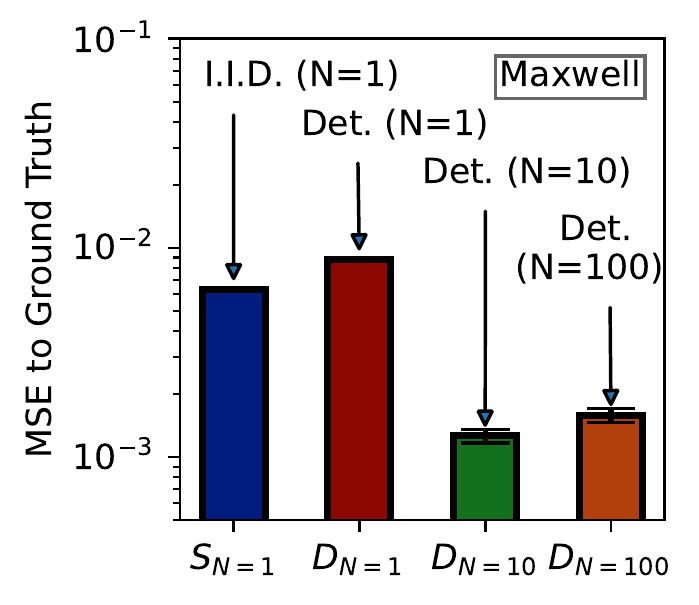}
	\includegraphics[width=0.31\linewidth]{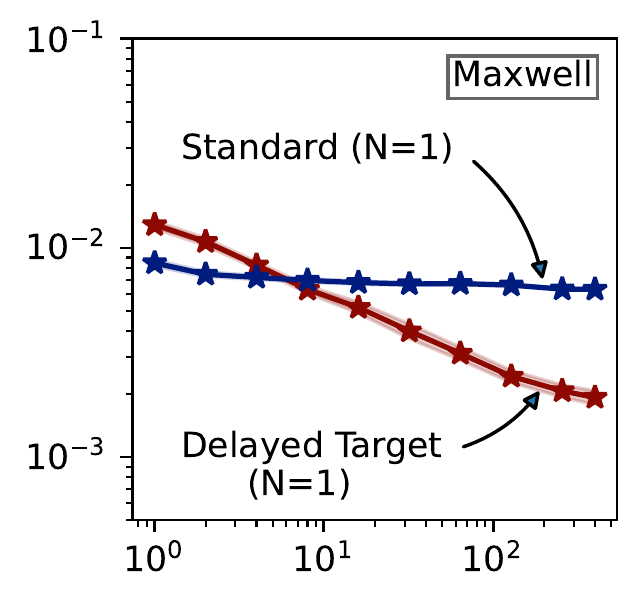}
	\includegraphics[width=0.33\linewidth]{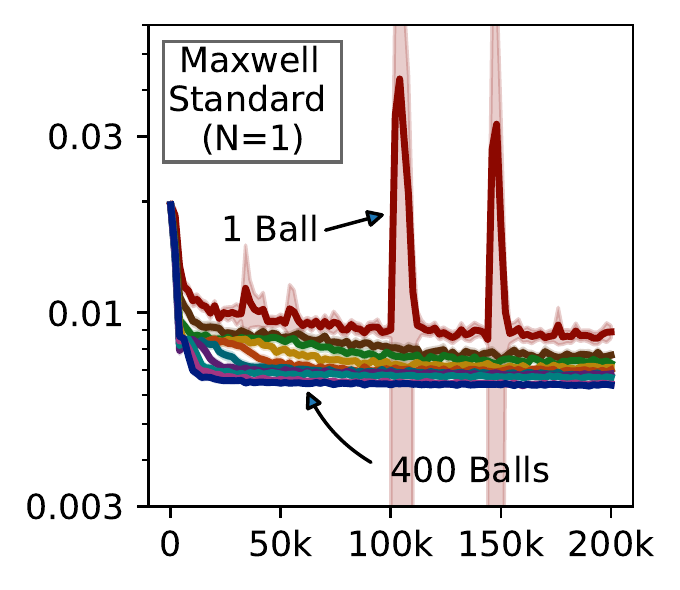}\\
	\includegraphics[width=0.34\linewidth]{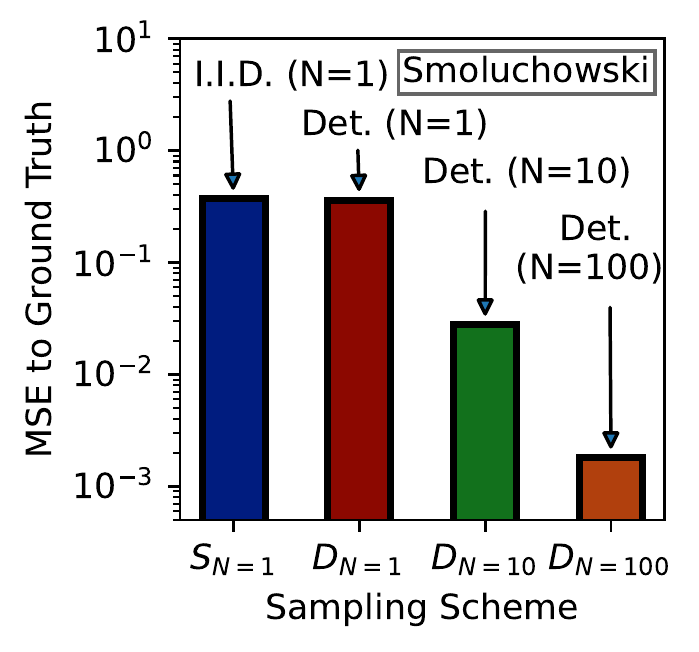}
	\includegraphics[width=0.31\linewidth]{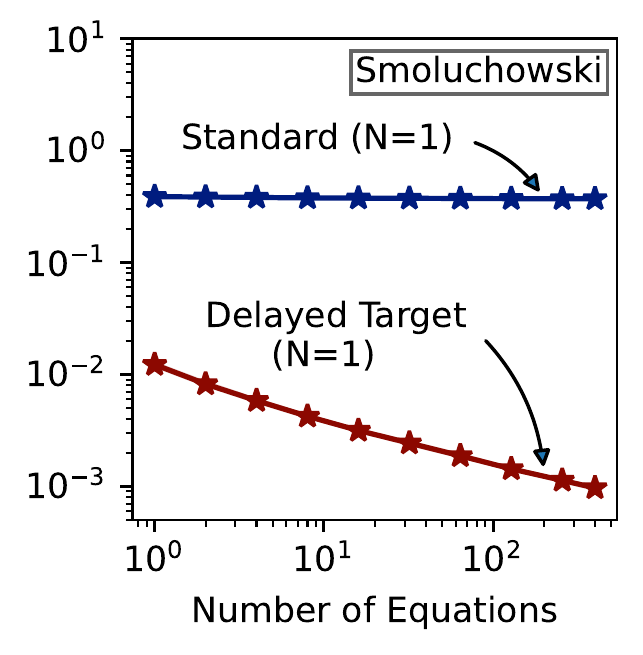}
	\includegraphics[width=0.31\linewidth]{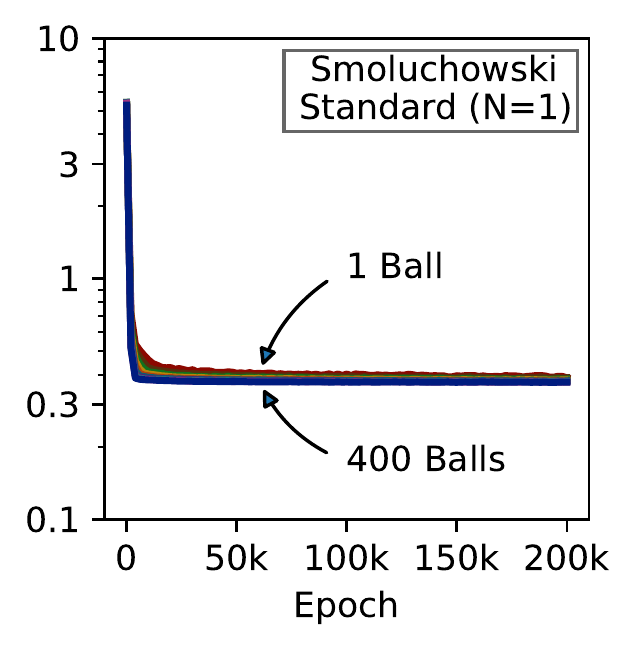}
	\caption{Ablation studies of the sampling hyper-parameters and settings in the Maxwell-Ampere and Smoluchowski problems. \textit{The top panel} corresponds to the Maxwell-Ampere problem of Figure~\ref{fig:maxwellmain} in the main paper, and {the bottom panel} corresponds to the 2D Smoluchowski problem of Figure~\ref{fig:smpanel} in the main paper. \textit{In the left column}, we compare the deterministic and i.i.d.\ sampling on the standard trainings with various $N$. \textit{In the middle column}, the horizontal axis shows the number of balls or equations sampled in each epoch. Both the standard and the delayed target methods are shown in this plot with $N=1$. \textit{The right column} shows the training curves for the standard method with $N=1$ target samples. Similar ablations for the Poisson problem are presented in Figure~\ref{fig:samplingabl}.\vspace{-2mm}}
	\label{fig:samplingablmaxsmol}
\end{figure*}

\subsection{The Computational Complexity of the Delayed Target Method}\label{sec:dtcompcmplxty}

For the sake of simplicity, assume that the delayed target and standard methods are executed over the abstracted space with maximal $x$ batch-sizes. In other words, consider the case where at each iteration, the gradients for all $x\in \mathcal{X}$ are estimated using either the delayed target or standard methods, and then the parameters are updated with the average gradient estimates of all $x\in \mathcal{X}$. We mainly make this assumption to strip the irrelevant $x\sim P$ sampling stochasticity from the standard method and make it deterministic. As a result, the standard method can be expressed as a gradient descent optimization instance. However, we still assume the $x'\sim P(\cdot|x)$ sampling stochasticity to remain within the delayed target method.

As we discussed, the delayed target method is an instance of stochastic approximation:
\begin{enumerate}
\item The cost of each iteration for the standard training method is $O(N d)$. However, the cost for each iteration of the delayed target method is $O(d)$ under the conditions of Theorem~\ref{thm:delayedtargetlin}.
\item To achieve an optimization error of $\epsilon$ in its respective problem, the standard training method needs $O(\log(1/\epsilon))$ iterations, since it is an instant of the gradient descent algorithm. However, the delayed target method needs $O(1/\epsilon)$ iterations to reach the same optimization error, since it is an instance of the stochastic gradient descent algorithm.
\end{enumerate}

Therefore, the total computational cost for the delayed target method to reach an optimization error of $\epsilon$ is $O(d/\epsilon)$. However, the standard training method needs $O(N d \log(1/\epsilon))$ to achieve the same goal.

Notice that the gradient averaging assumption over all $x\sim P$ is not overly restrictive. Both methods can identically use a smaller $x$ batch-size in practice, and the computational complexity insights regarding the $x'$ stochasticity remain applicable. We only introduced this assumption to make the computational complexity analysis easier to express.

\begin{figure*}[!t]
	\centering
	\includegraphics[width=0.345\linewidth]{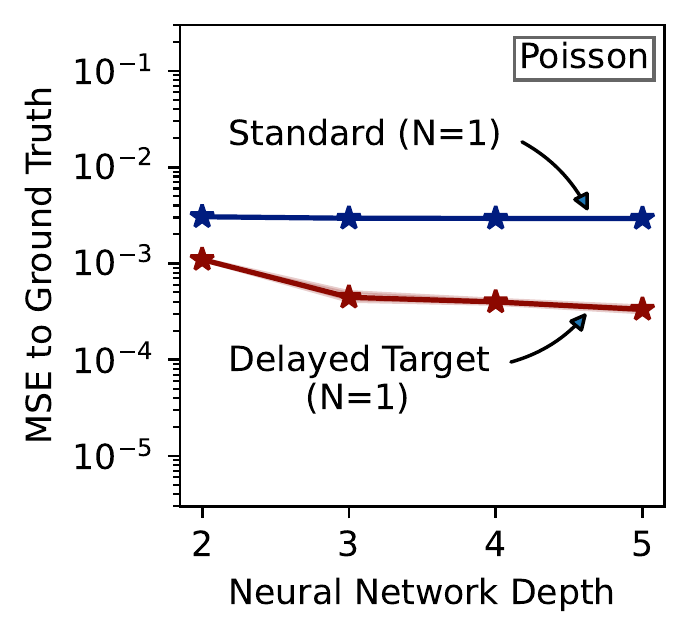}
	\includegraphics[width=0.215\linewidth]{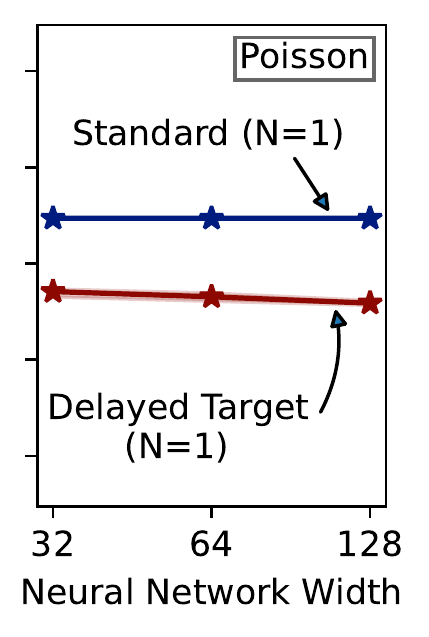}
	\includegraphics[width=0.2075\linewidth]{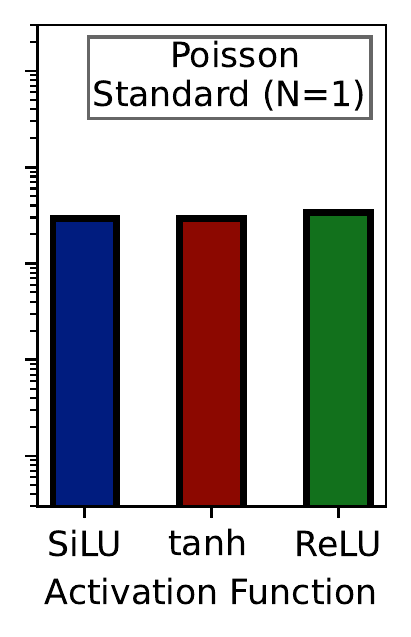}
	\includegraphics[width=0.2075\linewidth]{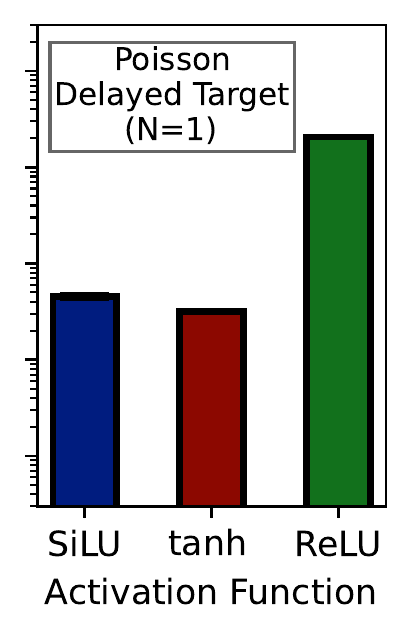}
	\vspace{-3mm}\caption{Ablating the function approximation class attributes on the 2D Poisson problem of Figure~\ref{fig:msegt} in the main paper. A multi-layer perceptron was used in all of our experiments. \textit{The left and right line plots} show the effect of the neural network's depth and width, respectively, on each of the standard and delayed target methods. \textit{The left and right bar plots} demonstrate the effect of the neural activation function on the standard and the delayed target methods, respectively. These results indicate that the function approximation class can have a more substantial impact on the delayed target method than the standard trainings. Similar ablations for the Maxwell-Ampere and Smoluchowski problems are presented in Figure~\ref{fig:archablmaxsmol}.}\vspace{-3mm}
	\label{fig:archabl}
\end{figure*}

\section{Ablation Studies}\label{sec:pinnablations}

Here, we examine the effect of different design choices and hyper-parameters with various ablation studies.

\subsection{Surface Point Sampling Scheme Ablations}
Figure~\ref{fig:samplingabl} compares the deterministic vs. i.i.d.\ sampling schemes and the effect of various mini-batch sizes (i.e., the number of volumes sampled in each epoch) on the Poisson problem of Figure~\ref{fig:msegt} in the main paper. Figure~\ref{fig:samplingablmaxsmol} shows similar ablations for the Maxwell-Ampere and Smoluchowski problems. The results suggest that the deterministic sampling scheme can train successfully with large $N$, however, it may not remedy the biased solution problem with the standard training at small $N$ values. Furthermore, the results indicate that the number of volumes in each epoch has minimal to no effect on the standard training method, which indicates that such a parallelization is not the bottleneck for the standard training method. To make this clear, we showed the training curves for the standard method, and they indicate similar trends and performance across a large range ($1$ to $400$) of batch-size values for the standard method.

On the other hand, the performance of the delayed target method improves upon using a larger batch size, which possibly indicates that this problem has a high objective estimation variance. The ability to trade small-quantity high-quality data (i.e., large $N$ with small mini-batch sizes) with large-quantity low-quality data (i.e., small $N$ with larger mini-batch sizes) is an advantage of the delayed target method relative to standard trainings.

\subsection{Function Approximation Ablations} 
\begin{figure*}[!t]
	\centering
	\includegraphics[width=0.345\linewidth]{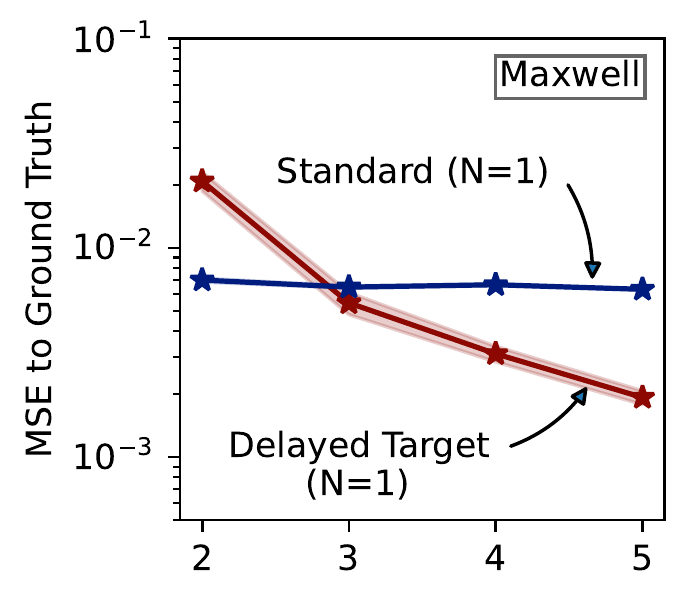}
	\includegraphics[width=0.215\linewidth]{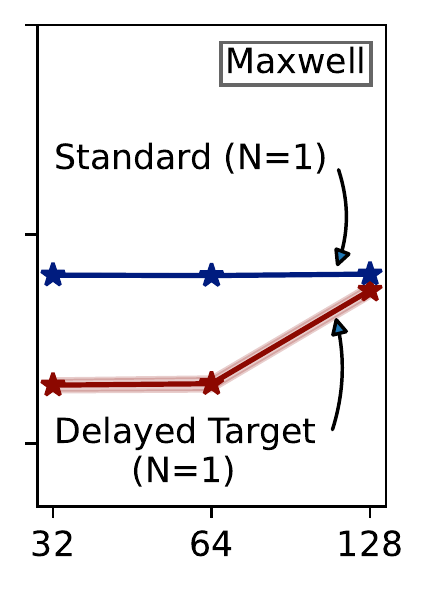}
	\includegraphics[width=0.2075\linewidth]{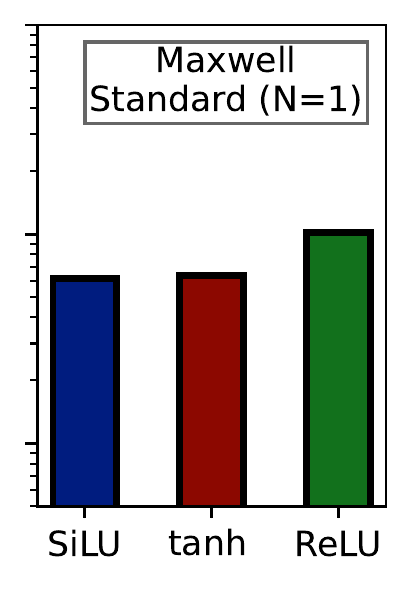}
	\includegraphics[width=0.2075\linewidth]{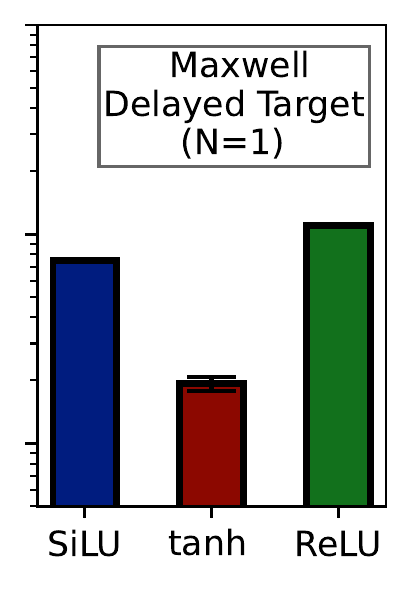}\\
	\vspace{-3mm}\includegraphics[width=0.345\linewidth]{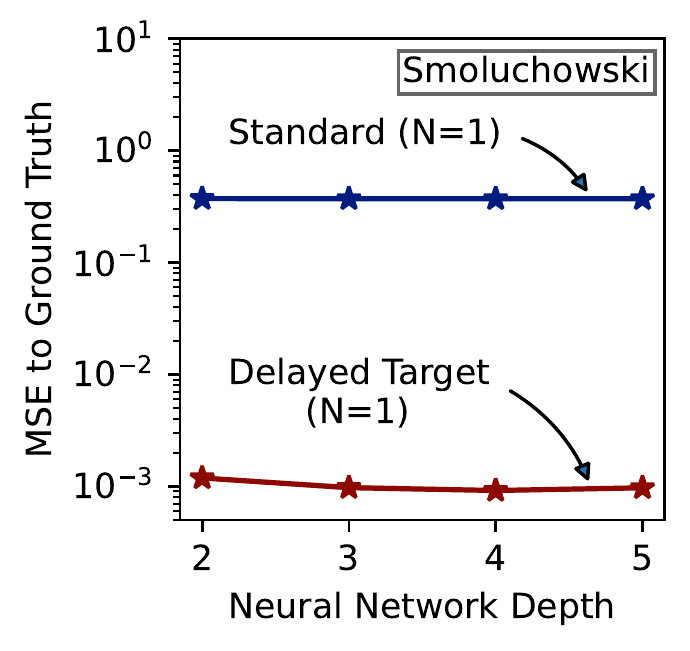}
	\includegraphics[width=0.215\linewidth]{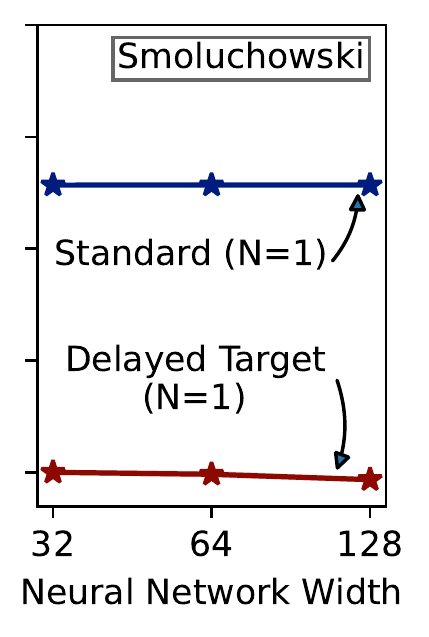}
	\includegraphics[width=0.2075\linewidth]{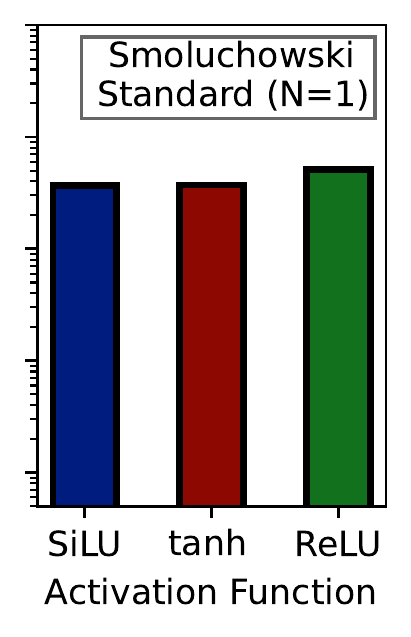}
	\includegraphics[width=0.2075\linewidth]{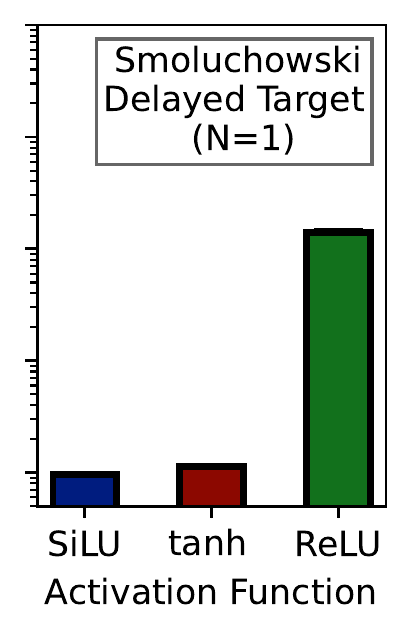}
	\vspace{-5mm}\caption{Ablating the function approximation class attributes on the Maxwell-Ampere and Smoluchowski problems. A multi-layer perceptron was used in all of our experiments. \textit{The top panel} corresponds to the Maxwell-Ampere problem of Figure~\ref{fig:maxwellmain} in the main paper, and {the bottom panel} corresponds to the 2D Smoluchowski problem of Figure~\ref{fig:smpanel} in the main paper. \textit{The left and right line plots} show the effect of the neural network's depth and width, respectively, on each of the standard and delayed target methods. \textit{The left and right bar plots} demonstrate the effect of the neural activation function on the standard and the delayed target methods, respectively. These results indicate that the function approximation class can have a more substantial impact on the delayed target method than the standard trainings. Similar ablations for the Poisson problem are presented in Figure~\ref{fig:archabl}.}\vspace{-5mm}
	\label{fig:archablmaxsmol}
\end{figure*}

Figure~\ref{fig:archabl} compares the effect of the neural architecture parameters on the performance of the standard training vs. the delayed target method on the 2D Poisson problem of Figure~\ref{fig:msegt} in the main paper. Figure~\ref{fig:archablmaxsmol} shows similar ablations for the Maxwell-Ampere and Smoluchowski problems. The standard training exhibits a steady performance across all neural network depths, widths, and activation functions. However, the performance of the delayed target method seems to be enhancing with deeper and wider networks. The effect of the neural activation functions is more pronounced than the depth and width of the network. In particular, the ReLU activation performs substantially worse than the $\tanh$ or SiLU activations, and the SiLU or $\tanh$ activations seem to work similarly.

\begin{figure*}[!t]
	\centering
	\includegraphics[width=0.38\linewidth]{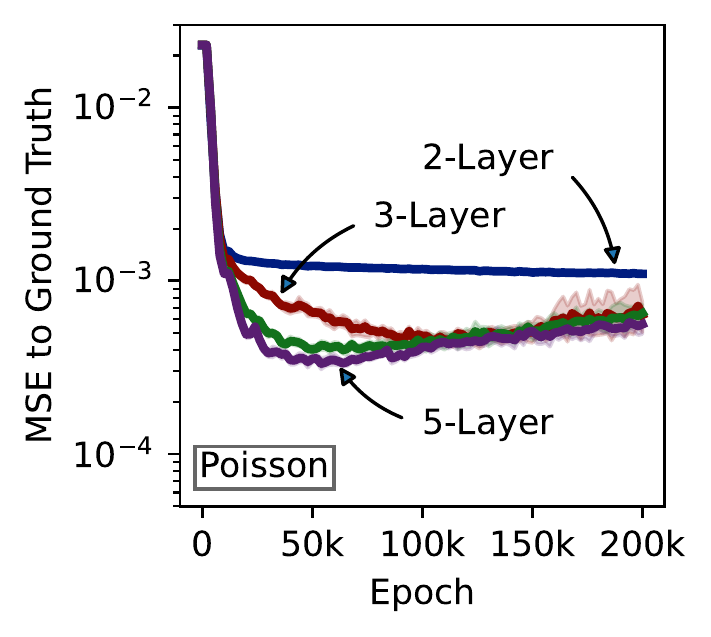}
	\includegraphics[width=0.301\linewidth]{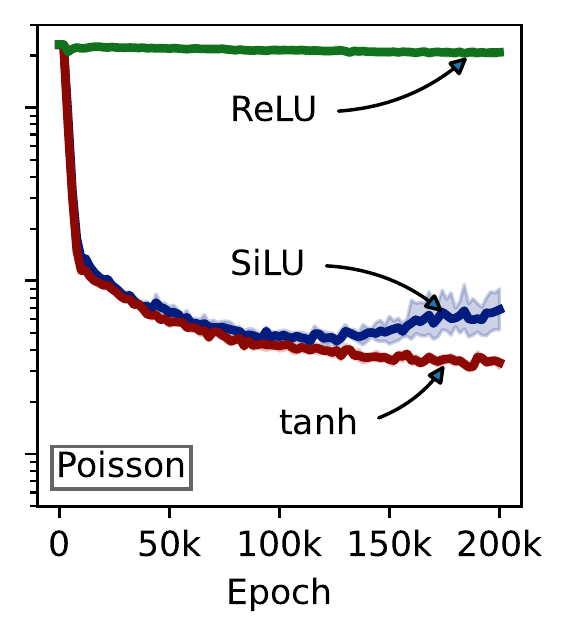}
	\includegraphics[width=0.301\linewidth]{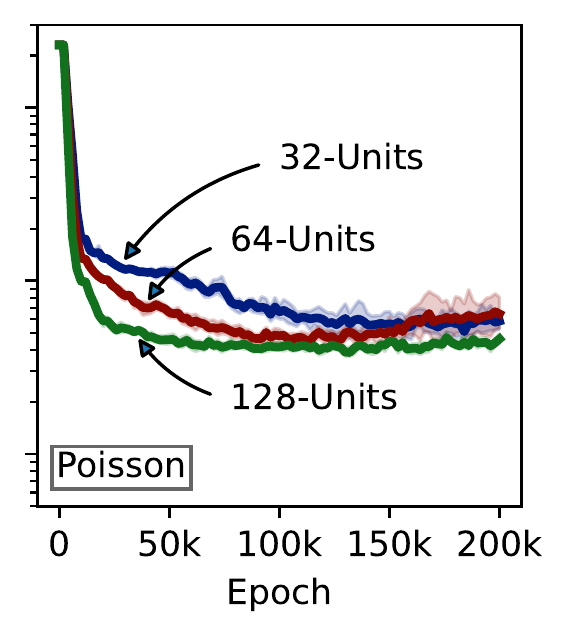}
	\vspace{-4mm}\caption{A closer look at the training curves for the delayed target method with different neural network hyper-parameters on the 2D Poisson problem of Figure~\ref{fig:msegt} in the main paper. \textit{The left, middle, and right plots} show the training curves for various neural depths, activations, and widths, respectively. Similar ablations for the Maxwell-Ampere and Smoluchowski problems are presented in Figure~\ref{fig:btsarchtrncurvemaxsmol}.}
	\label{fig:btsarchtrncurve}
\end{figure*}

\begin{figure*}[!t]
	\centering
	\includegraphics[width=0.38\linewidth]{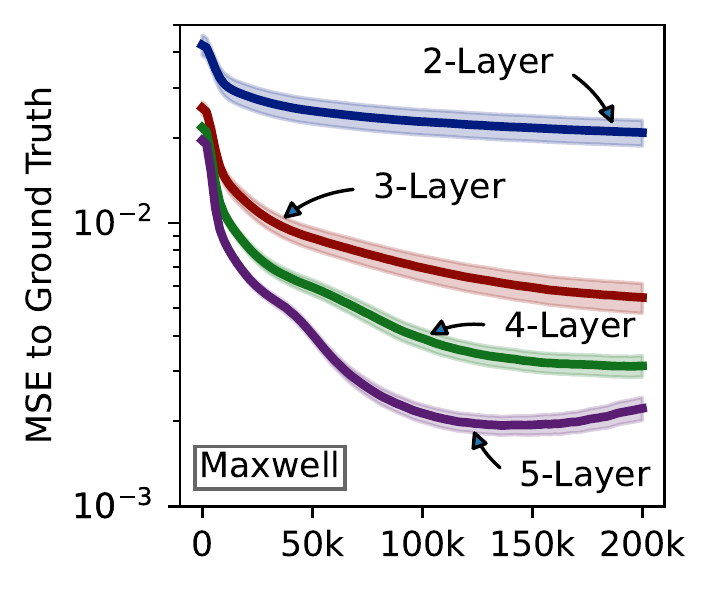}
	\includegraphics[width=0.301\linewidth]{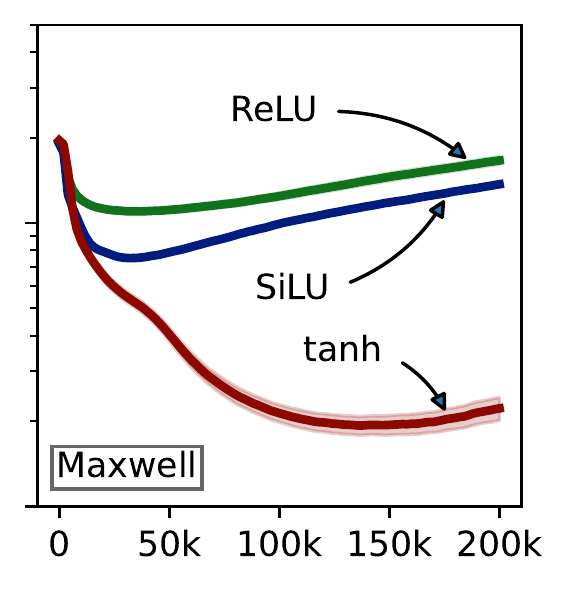}
	\includegraphics[width=0.301\linewidth]{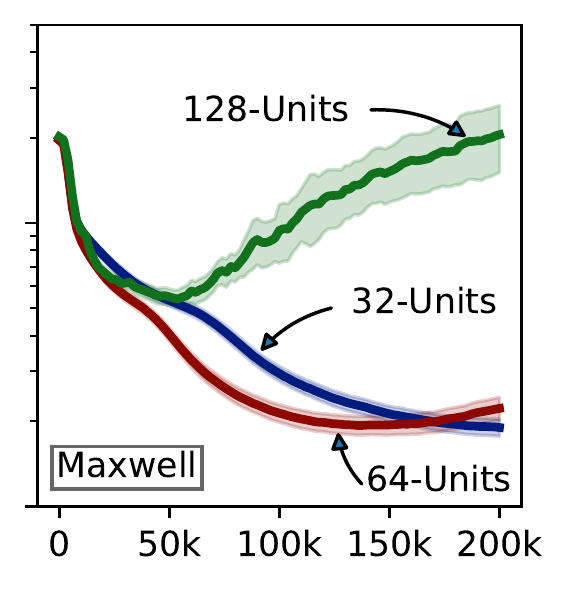}\\
	\vspace{-3mm}\includegraphics[width=0.38\linewidth]{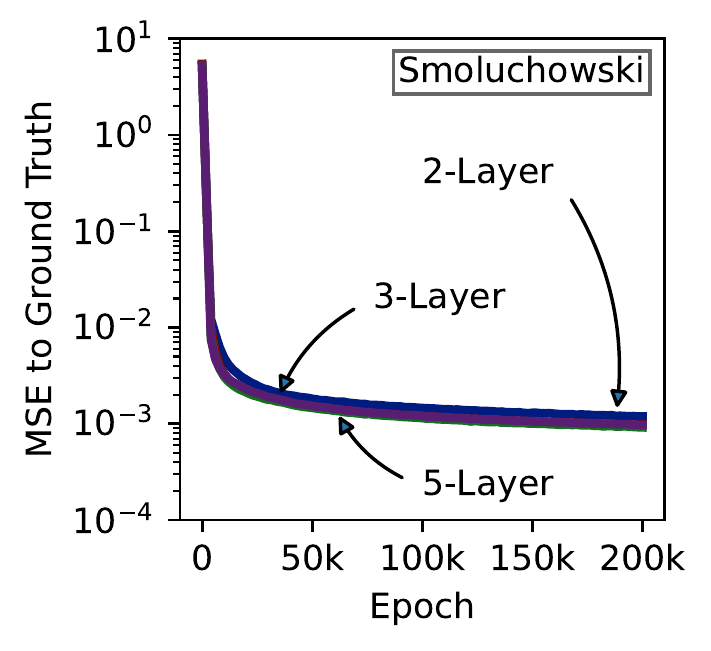}
	\includegraphics[width=0.301\linewidth]{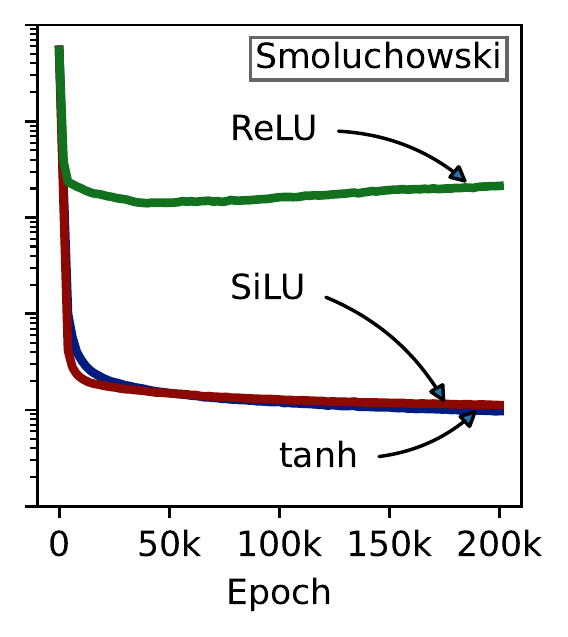}
	\includegraphics[width=0.301\linewidth]{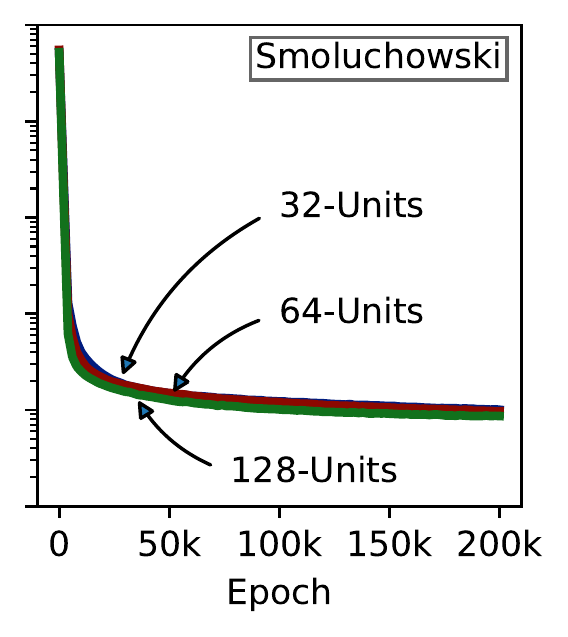}
	\vspace{-4mm}\caption{A closer look at the training curves for the delayed target method with different neural network hyper-parameters on the Maxwell-Ampere and Smoluchowski problems. \textit{The top panel} corresponds to the Maxwell-Ampere problem of Figure~\ref{fig:maxwellmain} in the main paper, and {the bottom panel} corresponds to the 2D Smoluchowski problem of Figure~\ref{fig:smpanel} in the main paper. \textit{The left, middle, and right columns} show the training curves for various neural depths, activations, and widths, respectively. Similar ablations for the Poisson problem are presented in Figure~\ref{fig:btsarchtrncurve}.}\vspace{-3mm}
	\label{fig:btsarchtrncurvemaxsmol}
\end{figure*}

To shed some further light on the training behavior of the delayed target method, we show the training curves for different neural hyper-parameters in Figures~\ref{fig:btsarchtrncurve} and~\ref{fig:btsarchtrncurvemaxsmol}. The results indicate that the ReLU activation prevents the delayed target method from improving during the entire training. On the other hand, the SiLU activation yields better initial improvements but struggles to maintain this trend consistently in the Poisson and Maxwell-Ampere problems. Based on this, we speculate that some activation functions (e.g., ReLU) may induce poor local optima in the optimization landscape of the delayed target method, which may be difficult to run away from. 

Our neural depth analysis in Figure~\ref{fig:btsarchtrncurve} indicates that deeper networks can yield quicker improvements in performance. However, such improvements are difficult to maintain stably over the entire course of training. In particular, the 2-layer training yields worse performance than the deeper networks, but maintains a monotonic improvement throughout the training, unlike the other methods. Of course, such behavior may be closely tied together with the activation function used for function approximation, as we discussed earlier. On the other hand, deeper networks in the Maxwell-Ampere problem produce better results consistently in Figure~\ref{fig:btsarchtrncurvemaxsmol}. In the Smoluchowski problem, the depth of the network makes little to no difference in the performance. These examples demonstrate a variety of different behaviors for the role of neural depth in the delayed target method. A better understanding of this pathology with respect to the problem conditions and the rest of the hyper-parameters is an important topic for future research.

\vspace{-0.5mm}We also show the effect of network width on the performance of the delayed target method in Figures~\ref{fig:btsarchtrncurve} and~\ref{fig:btsarchtrncurvemaxsmol}. In the Poisson problem, wider networks are more likely to provide better initial improvements. In the Maxwell-Ampere problem, the widest network produces poor performance. Finally, in the Smoluchowski problem, there is no substantial difference in performance between different network widths. That being said, as the networks are trained for longer, the differences in performance between different network widths shrink, and narrower networks tend to show similar final performances as the wider networks (assuming a stable training behavior).

\vspace{-0.5mm}All in all, our results indicate that the choice of the neural function approximation class, particularly with varying activation and depths, can have a notable impact on the performance of the delayed target method. We speculate that this is due to the incomplete gradients used during the optimization process of the delayed target method. The effect of incomplete function approximation on bootstrapping methods has been studied frequently, both in theory and practice, in other contexts such as the Fitted Q-Iteration (FQI) and Q-Learning methods within reinforcement learning. This is in contrast to the standard training methods, which seem quite robust to function approximation artifacts at the expense of solving an excess-variance diluted optimization problem.

\subsection{Integration Volume Sampling Ablations}

\begin{figure}[!t]
	\centering
	\includegraphics[width=0.33\linewidth]{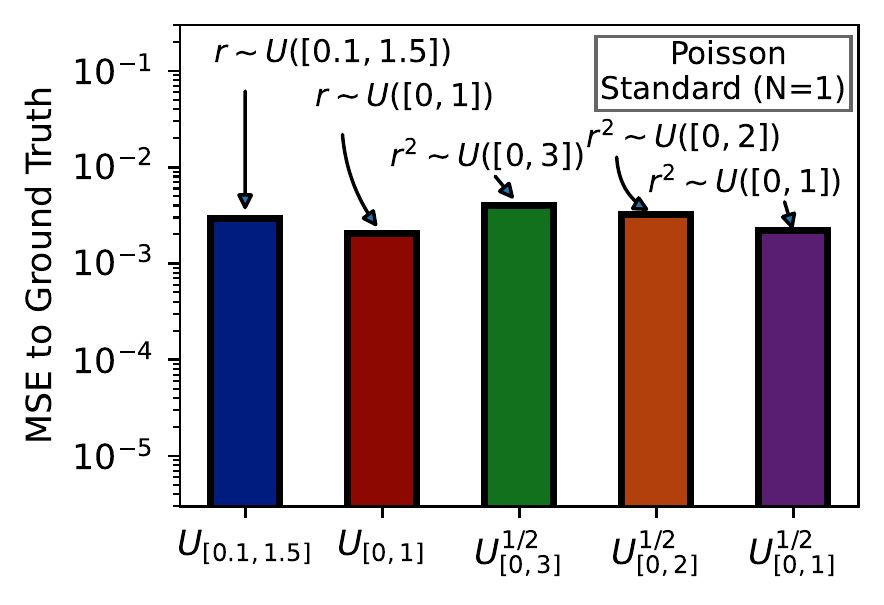}
	\includegraphics[width=0.185\linewidth]{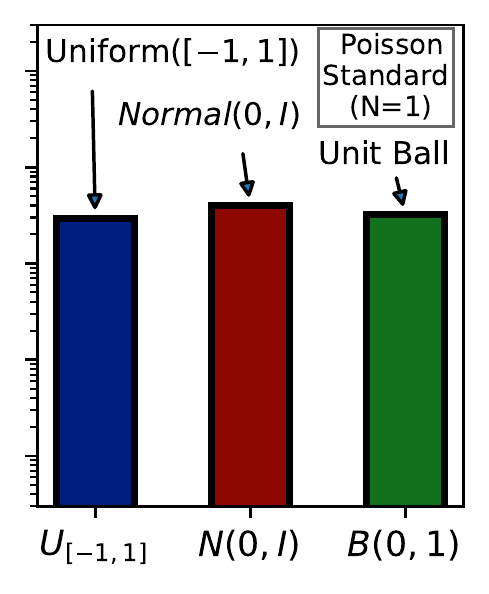}
	\includegraphics[width=0.275\linewidth]{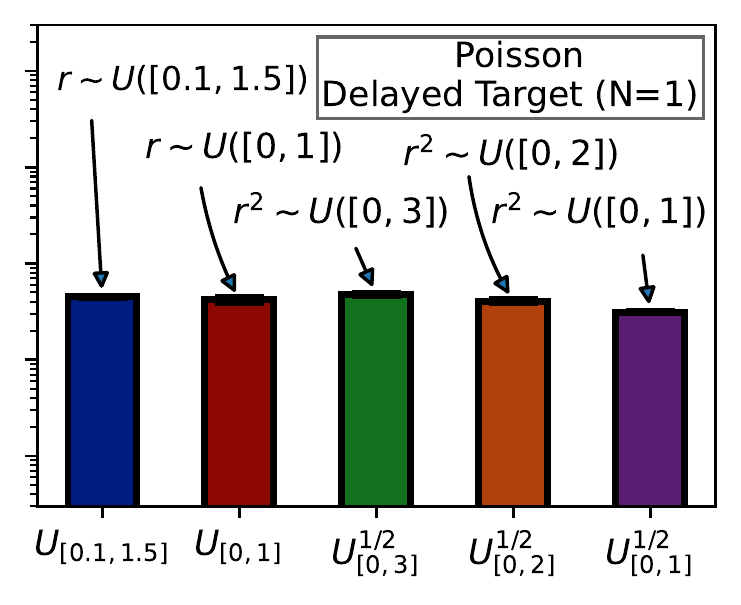}
	\includegraphics[width=0.185\linewidth]{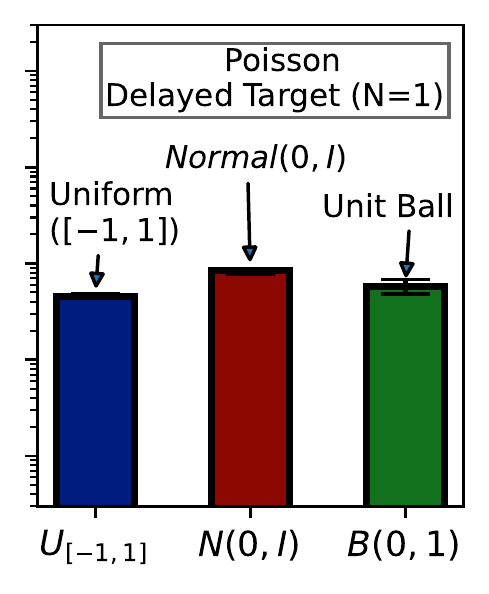}\\
	\includegraphics[width=0.33\linewidth]{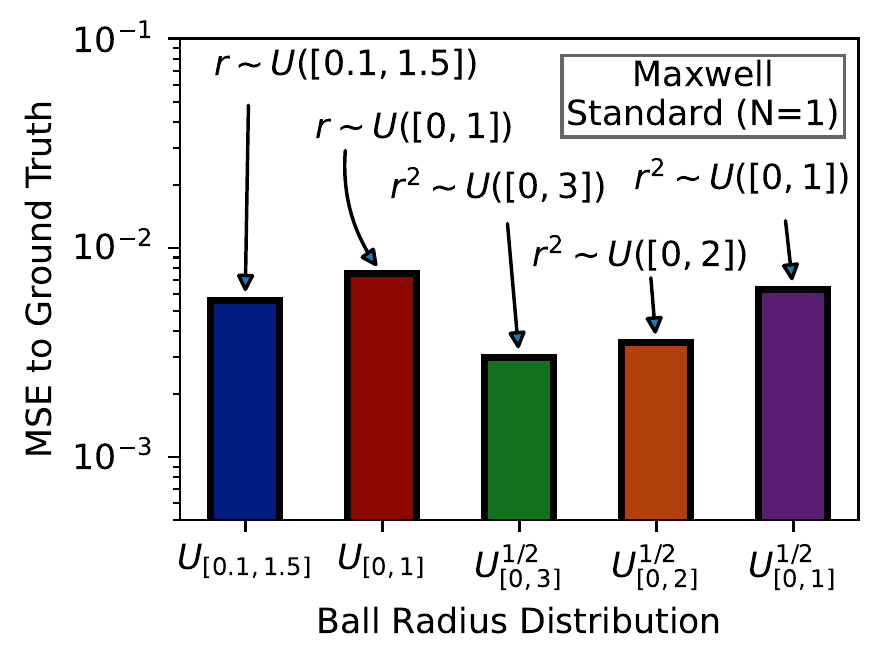}
	\includegraphics[width=0.185\linewidth]{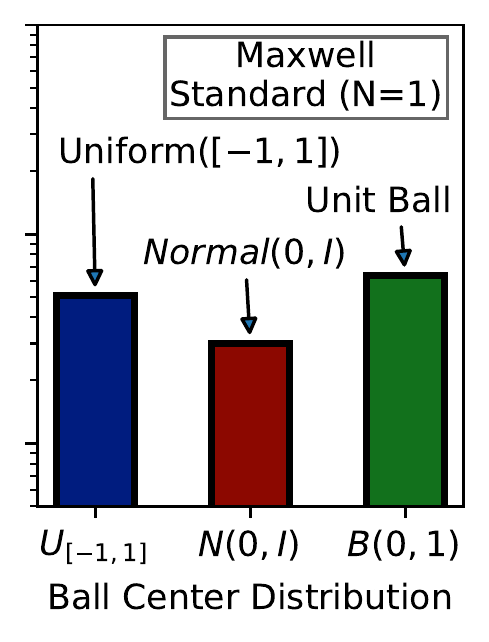}
	\includegraphics[width=0.275\linewidth]{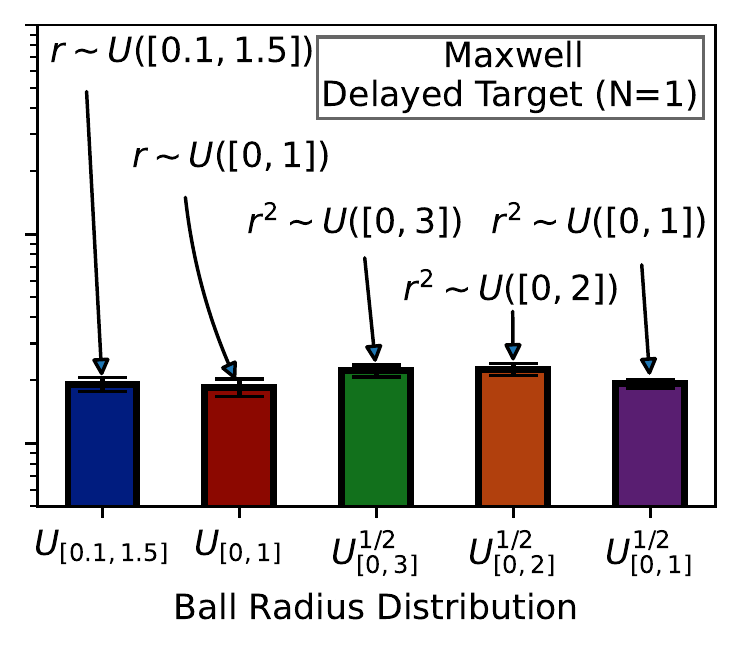}
	\includegraphics[width=0.185\linewidth]{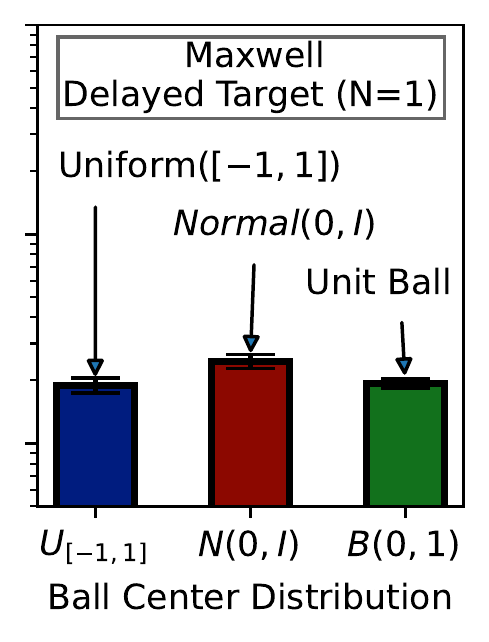}
	\vspace{-4mm}\caption{Ablating the distribution of ball centers and radii for both of the standard and the delayed target methods, both with $N=1$.  \textit{The top panel} corresponds to the 2D Poisson problem of Figure~\ref{fig:msegt} in the main paper, and {the bottom panel} corresponds to the Maxwell-Ampere problem of Figure~\ref{fig:maxwellmain} in the main paper. \textit{The two left columns} correspond to the standard method, and the \textit{the two right columns} correspond to the delayed target method. In each quarter, \textit{the left plot} shows the effect of the ball radius distribution, and \textit{the right plot} shows the effect of the ball center distribution. In the radius ablations, $r\sim U([a,b])$ means the ball radius was sampled from a uniform distribution over $[a,b]$. Also, $r^2\sim U([a,b])$ means that the radius was the square root of a uniform random variable between $a$ and $b$. The ball centers were randomly picked either (1) uniformly over a square with the $[-1, -1], [-1, 1], [1, 1], [1, -1]$ vertices, (2) normally, or (3) uniformly within the unit ball.}\vspace{-3mm}
	\label{fig:ballabl}
\end{figure}

For our integration volumes, we randomly sampled balls of varying radii and centers. The distribution of the sampled radii and centers could impact the performance of different methods. Figure~\ref{fig:ballabl} studies such effects on both the standard and the delayed target methods in the 2D Poisson problem of Figure~\ref{fig:msegt} and the Maxwell-Ampere problem of Figure~\ref{fig:maxwellmain} in the main paper. In short, we find that the delayed target method is robust to such sampling variations; an ideal method should find the same optimal solution with little regard to the integration volume distribution. On the other hand, our results indicate that the standard training performance tends to be sensitive to the integration volume distributions. This may be because the standard trainings need to minimize two loss terms; the optimal balance between the desired loss function $\LL_{\theta}(x)$ and the excess variance $\VV_{\NSPDF}[g_\theta(x')]$ in Equation~\eqref{eq:excessvar} may be sensitive to the distribution of $x$ itself.

\subsection{Poisson charge placement ablations} 
\begin{figure*}[t]
	\centering
	\includegraphics[width=0.533\linewidth]{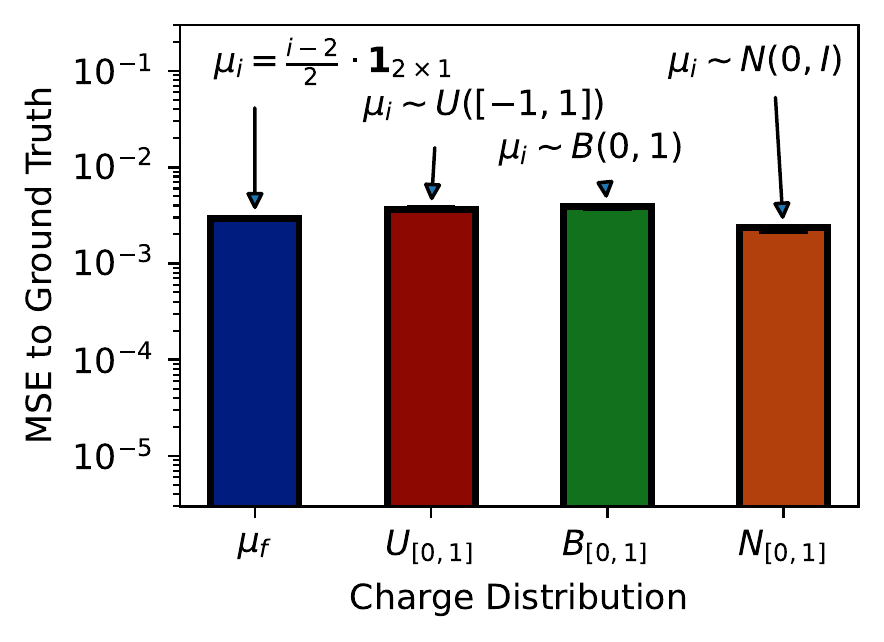}
	\includegraphics[width=0.447\linewidth]{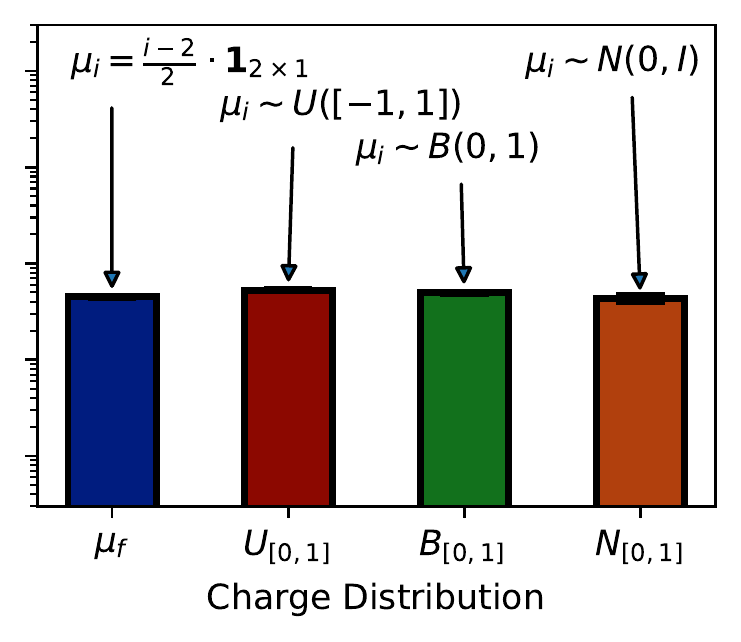}
	\vspace{-5mm}\caption{Ablating the location distribution of the three charges on the 2D Poisson problem of Figure~\ref{fig:msegt} in the main paper. \textit{The left plot} shows the results for the standard training method, while the \textit{the right plot} corresponds to the delayed target method, all with $N=1$. The blue bar represents fixing the charge locations at the $[0,0]$, $[-0.5, -0.5]$, and $[0.5, 0.5]$ coordinates. We also show the results for picking the charge locations in an i.i.d.\ manner (1) uniformly between $[-1, -1]$ and $[1,1]$ (denoted as $U([-1,1]$), (2) uniformly over the unit ball (denoted as $B(0,1)$), and (3) normally (denoted as $N(0,I)$).}\vspace{-5mm}
	\label{fig:chrgabl}
\end{figure*}

The charge locations in the 2D Poisson problem may impact the results of our methods. For this, we compare the standard and the delayed target method over a wide variety of charge distributions. Figure~\ref{fig:chrgabl} summarizes these results. Here, the three fixed charge locations shown in Figure~\ref{fig:msegt} of the main paper are shown as a baseline. We also show various problems where the charge locations were picked uniformly or normally in an i.i.d.\ manner. The performance trends seem to be quite consistent for each method, and the fixed charge locations seem to represent a wide range of such problems and datasets.

\subsection{Robustness to the Initial Conditions} 
Our method is extremely robust to the neural network initializations as shown by the small confidence intervals in our results. In addition, physics-informed networks can readily handle different PDE initial conditions. The delayed target method is less sensitive to the weight placed on the initial condition loss term since it can effectively eliminate the excess variance term. This is in contrast to the standard training method where the initial condition enforcement may be negatively influenced by the excess variance term.

\subsection{Sampling Quadrature and QMC Integration Points}\label{sec:quadqmcsamplng}
Obtaining uniform samples on the surface of the integration volumes is a necessary step to estimate the divergence theorem integrals. Deterministic methods, such as numerical quadrature and QMC, can be sensitive to the specific choice of these points as they define a constant arrangement of surface points for the entirety of the training. On the other hand, stochastic methods, such as the standard, the double-sampling, and the delayed target methods, can be less sensitive to this problem as they sample the integration points in an i.i.d.\ manner and any point on the integration surfaces have the same non-zero probability of appearing in the integral estimates at each iteration.

For instance, consider a 2D Poisson problem. Uniform samples on the surface of a 2D unit ball (i.e., the unit circle) can be obtained in one of two ways. One approach is to sample 2-dimensional Gaussian random variables and normalize them so that they fall on the unit circle. This process defines a 2-dimensional integral since the normal random variables were sampled in the 2-dimensional space. Alternatively, we could sample uniform variables in the 1-dimensional space, and apply an appropriate transformation so that they cover the unit circle. This approach defines a 1-dimensional integral, instead. 

This distinction can be generalized to higher dimensional problems as well; to solve a $d$-dimensional Poisson problem, the Gaussian sampling and normalization approach defines a $d$-dimensional integral for the divergence theorem, while sampling from the $(d-1)$-dimensional cube and transforming the points to the surface of the $d$-dimensional unit ball defines a $(d-1)$-dimensional integral for the divergence theorem. Although both estimators are statistically consistent (i.e., yield the same integrals with infinite samples), under finite sample sizes, they define different integration point distributions on the integration surfaces for the deterministic methods.

\begin{figure*}[t]
	\centering
	\includegraphics[width=0.98\linewidth]{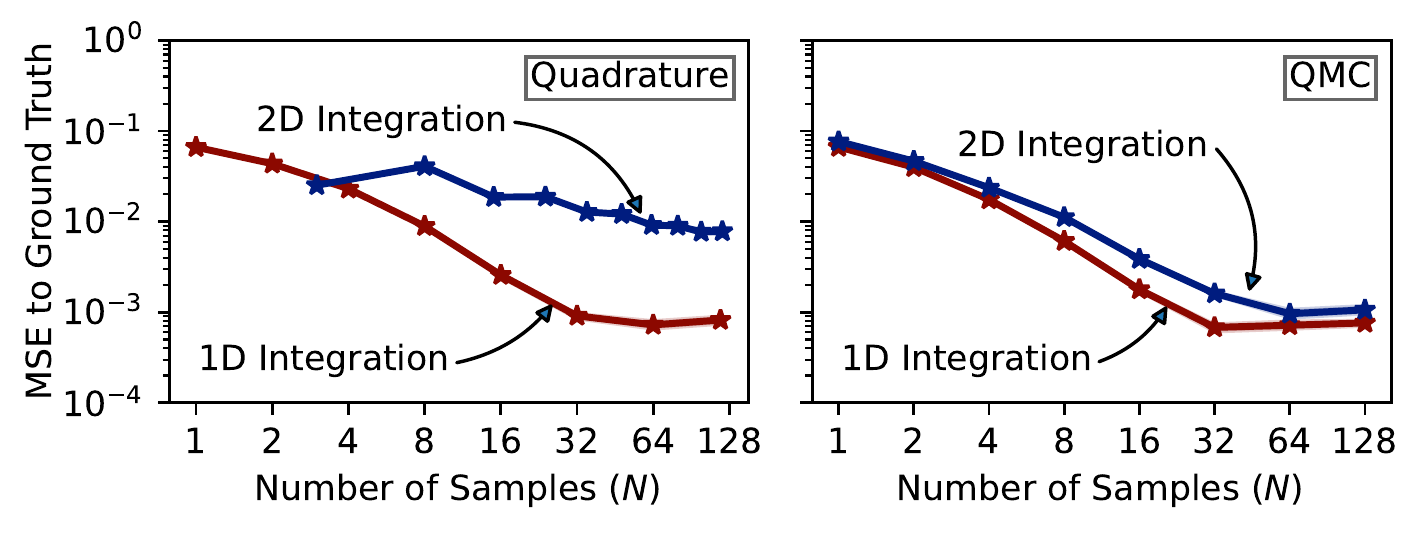}
	\vspace{-3mm}\caption{Demonstrating the effect of the specific sampling procedures on the quadrature and QMC methods. The \textit{left and the right plots} correspond to Gaussian quadrature and QMC methods, respectively. All runs were performed on a 2D Poisson problem from Figure~\ref{fig:hidimpoiss} of the main paper. No sparse grids were used here. Each curve denotes a different sampling procedure for obtaining the surface points in the divergence theorem. 2D integrations were defined by sampling normal random variables and normalizing them to fall on the unit circle. 1D integrations directly sampled uniform points on the unit circle. The latter approach yields the best results, and numerical quadrature seems to be particularly sensitive to this choice.}
	\label{fig:quadqmcintegdim}
\end{figure*}

Figure~\ref{fig:quadqmcintegdim} studies the effect of these sampling procedures on the deterministic methods in a 2D Poisson problem from Figure~\ref{fig:hidimpoiss} of the main paper. Sampling points from the $(d-1)$-dimensional area and transforming them to the surface of a $d$-dimensional ball seems to work best for both the QMC and numerical quadrature methods. However, numerical quadrature seems to be particularly sensitive to this choice; our results suggest that the specific choice of the $d$- or $(d-1)$-dimensional integral estimators can significantly impact the performance of numerical quadrature. 

Based on these results, we sampled points from the $(d-1)$-dimensional cube and transformed the points to the surface of the $d$-dimensional ball for the QMC and quadrature methods. This was done to present these methods in the best light.

\subsection{Joint Target Smoothing and Regularization}\label{sec:trgtaulammaxwell}
Both the target smoothing ($\tau$) and regularization ($\lambda$) weights have a role in regulating the delayed target updates. In particular, target smoothings regulate the target model updates. On the other hand, the target regularization controls the main model updates and prevents the main parameters from diverging from the vicinity of the target parameters. One may wonder how these two factors are practically different when regulating the parameter updates.

\begin{figure*}[t]
	\centering
	\includegraphics[width=0.98\linewidth]{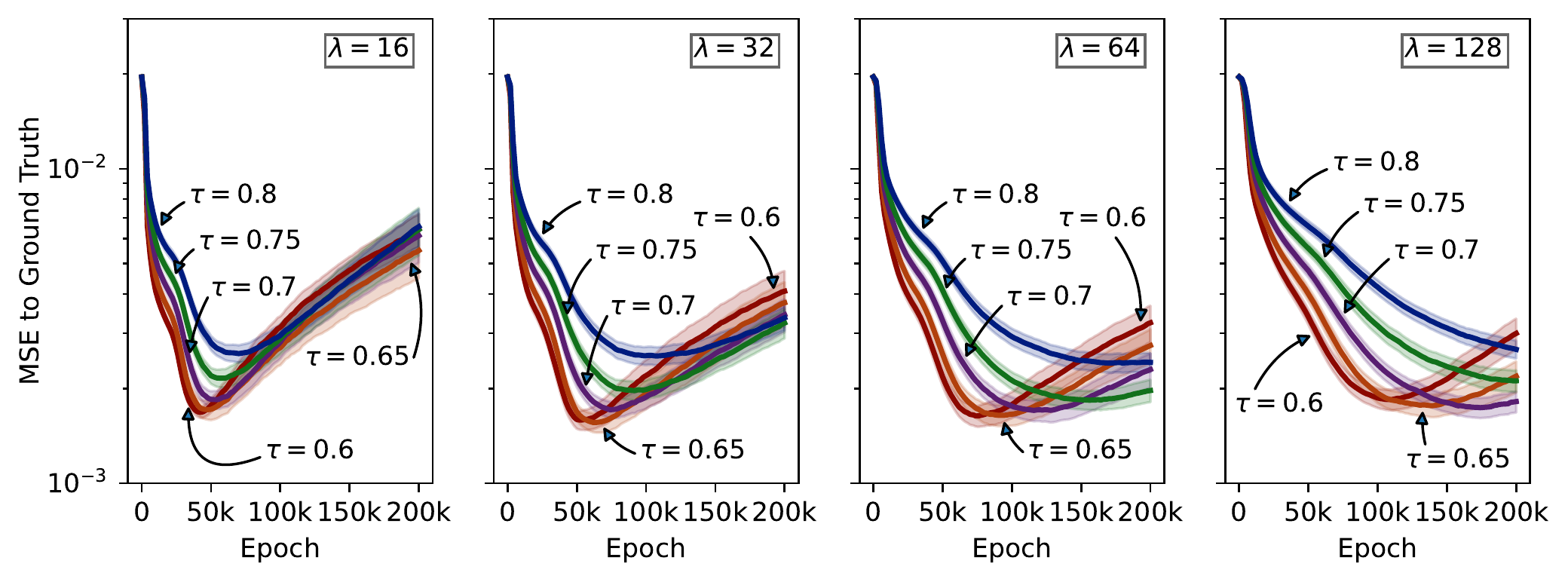}
	\vspace{-3mm}\caption{Studying the joint effect of target smoothing ($\tau$) and regularization ($\lambda$) on the performance of the delayed target method in the Maxwell problem of Figure~\ref{fig:maxwellmain} in the main paper. Each plot corresponds to a fixed target regularization; we set $\lambda$ to 16, 32, 64, and 128 in increasing order from the left to the right plots. Both hyper-parameters control the training speed. However, increasing the $\lambda$ target regularization has a stabilizing effect on the training curves, whereas tuning the target smoothing $\tau$ mainly improves the peak performance.}
	\label{fig:lamtauablmaxwell}
\end{figure*}

To understand the joint impact of these regulation factors, Figure~\ref{fig:lamtauablmaxwell} shows the delayed target training curves on a grid of these hyper-parameters. Since this is a singular problem and we used $M=10^3$ with $N=1$, this is a relatively challenging problem for the delayed target method to solve. The results indicate that both the target smoothing and regularization factors play a role in controlling the convergence of the delayed target method. The $\lambda$  target regularization can stabilize the training curves in such challenging setups. On the other hand, tuning the target smoothing $\tau$ can improve the peak performance of the method. Together, these two factors can regulate the delayed target updates effectively.

\subsection{Further Delayed Target Ablations}
\begin{figure*}[!t]
	\centering
	\includegraphics[width=0.44\linewidth]{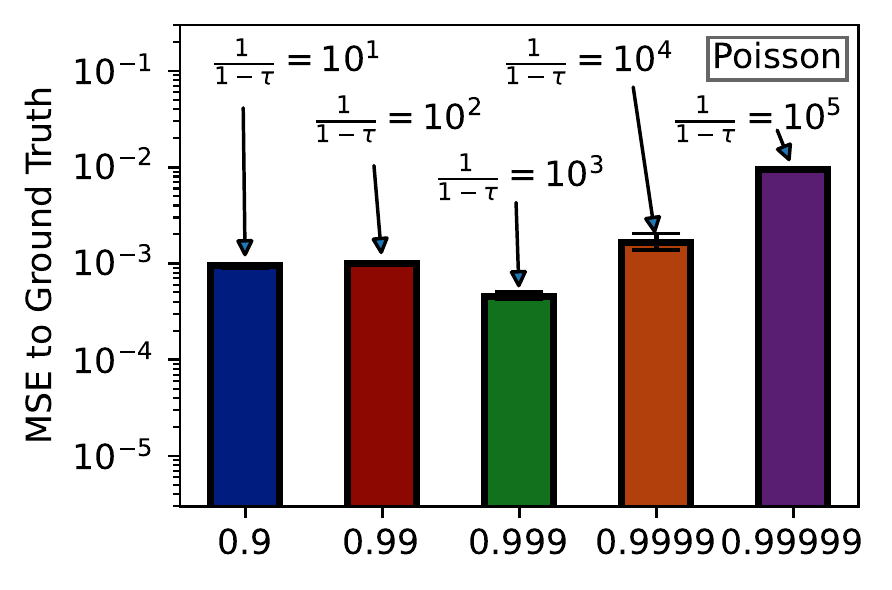}
	\includegraphics[width=0.273\linewidth]{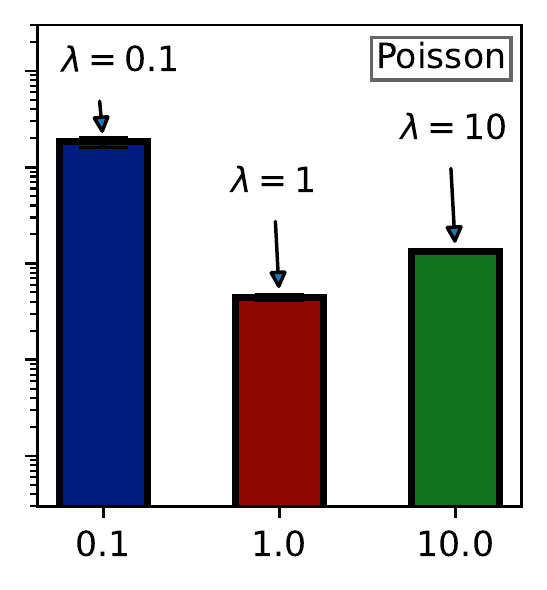}
	\includegraphics[width=0.268\linewidth]{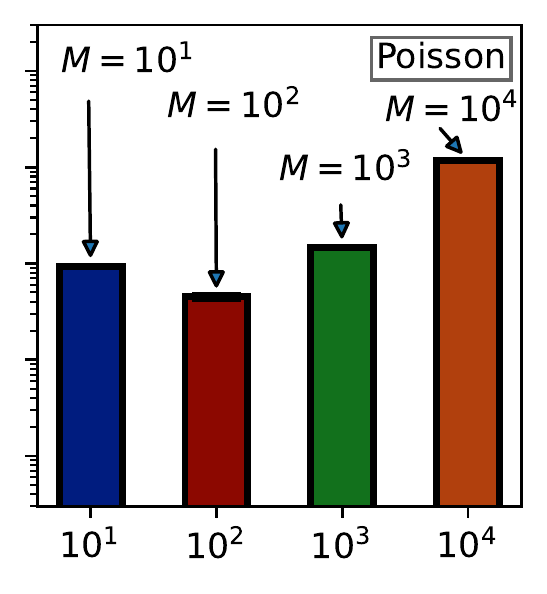}\\
	\includegraphics[width=0.44\linewidth]{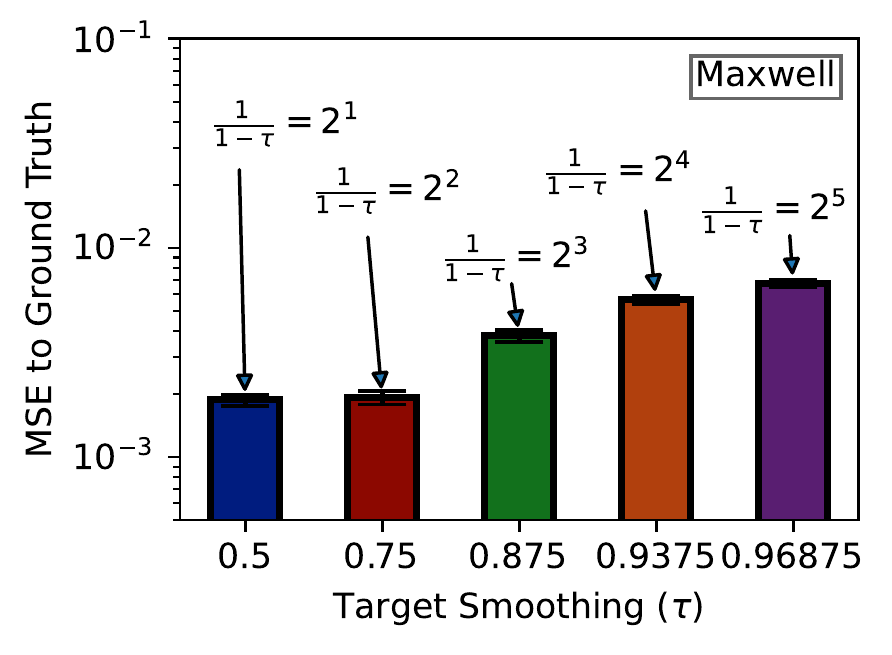}
	\includegraphics[width=0.273\linewidth]{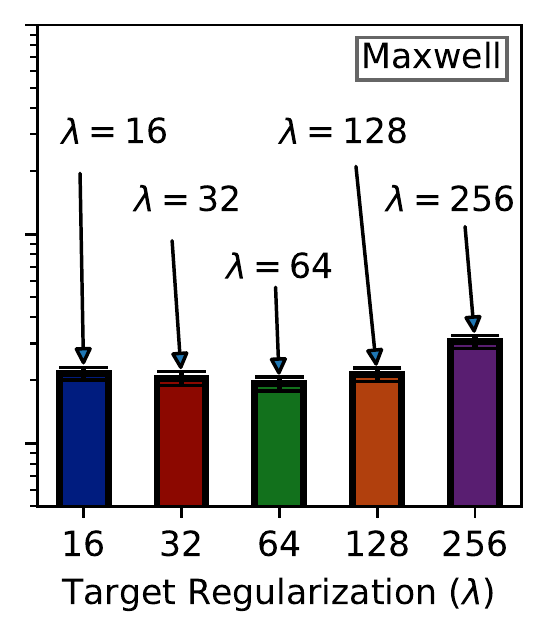}
	\includegraphics[width=0.268\linewidth]{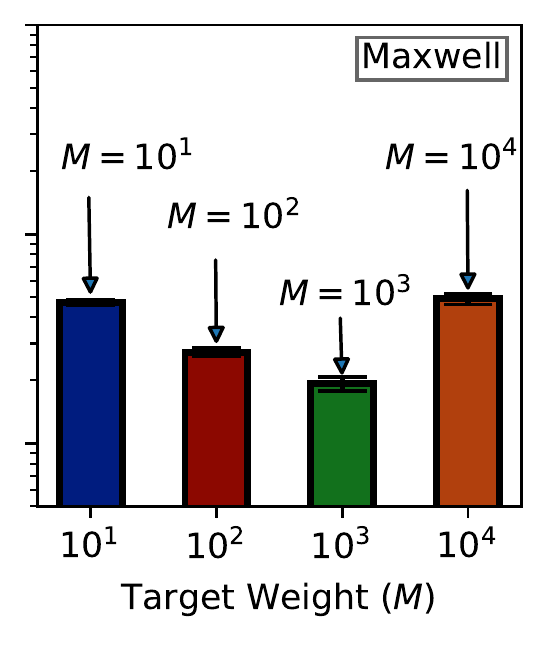}
	\vspace{-5mm}\caption{Ablating the effect of the delayed target method parameters on the Poisson and Maxwell-Ampere problems. \textit{The top panel} corresponds to the 2D Poisson problem of Figure~\ref{fig:msegt} in the main paper, and {the bottom panel} corresponds to the Maxwell-Ampere problem of Figure~\ref{fig:maxwellmain} in the main paper. \textit{The left column} shows the effect of the target smoothing parameter $\tau$ in Algorithm~\ref{alg:brs} of the main paper. \textit{The middle column} shows the effect of the target regularization parameter $\lambda$ in Algorithm~\ref{alg:brs} of the main paper. \textit{The right column} shows the effect of the target weight factor $M$ in Equation~\eqref{eq:targetvar}. Similar ablations for the Smoluchowski problem are presented in Figure~\ref{fig:btsablsmol}.}\vspace{-4mm}
	\label{fig:btsabl}
\end{figure*}

\begin{figure*}[!t]
	\centering
	\includegraphics[width=0.535\linewidth]{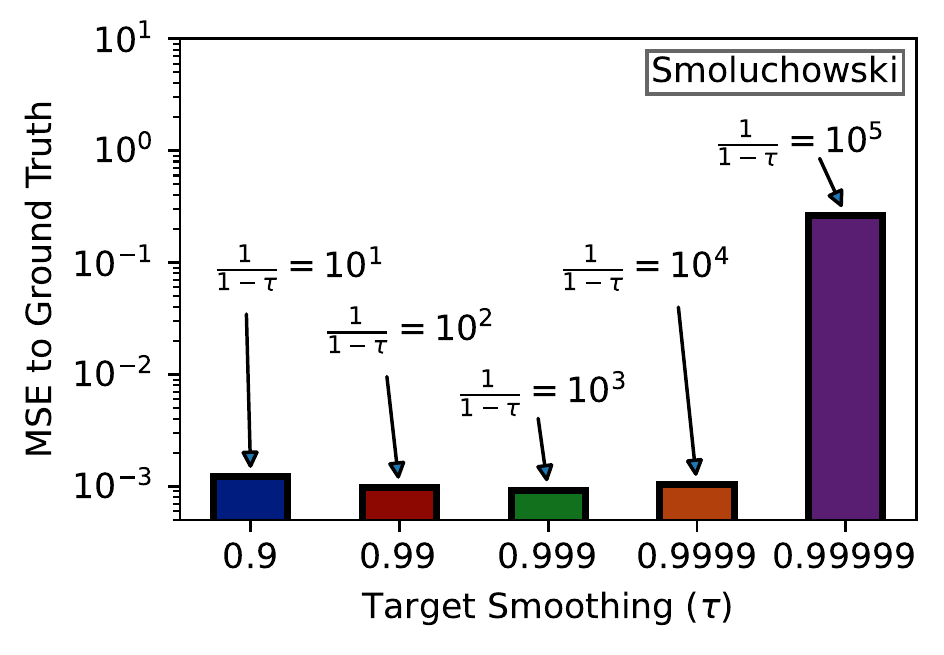}
	\includegraphics[width=0.455\linewidth]{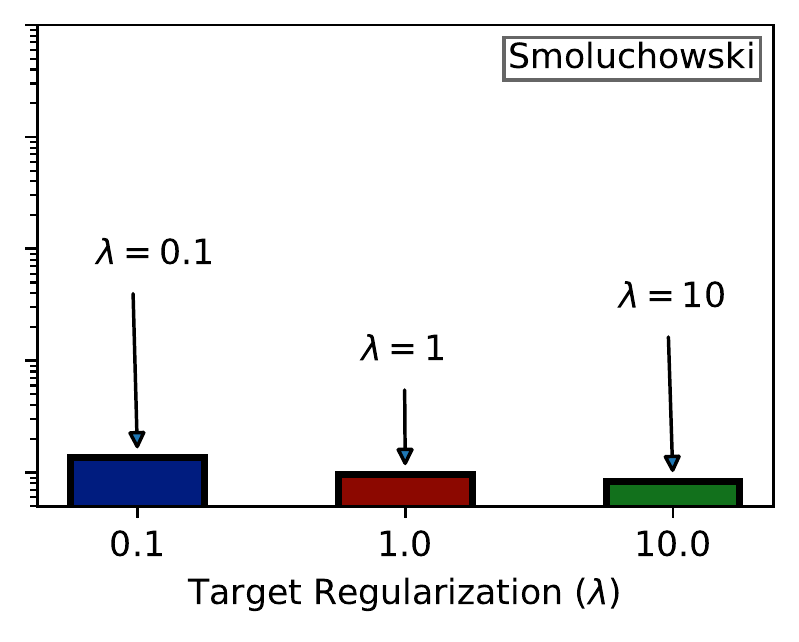}
	\vspace{-5mm}\caption{Ablating the effect of the delayed target method parameters on the 2D Smoluchowski problem of Figure~\ref{fig:smpanel} in the main paper. \textit{The left column} shows the effect of the target smoothing parameter $\tau$ in Algorithm~\ref{alg:brs} of the main paper. \textit{The right column} shows the effect of the target regularization parameter $\lambda$ in Algorithm~\ref{alg:brs} of the main paper. Similar ablations for the Poisson and Maxwell-Ampere problems are presented in Figure~\ref{fig:btsabl}.}
	\label{fig:btsablsmol}
\end{figure*}

Three main hyper-parameters are involved in the definition of the delayed target method: (1) the target smoothing $\tau$, (2) the target regularization weight $\lambda$, and (3) the target weight $M$ described in Equation~\eqref{eq:targetvar}. Figures~\ref{fig:btsabl}  and~\ref{fig:btsablsmol} study the effect of each of these hyper-parameters on the performance of the delayed target method in the Poisson, Maxwell-Ampere, and Smoluchowski problems.

Our results indicate that choosing a proper target smoothing can improve the performance of the delayed target method. In particular, neither a significantly small nor a substantially large $\tau$ can yield optimal training. Small $\tau$ values cause the training target to evolve rapidly. This may accelerate the training initially, but it can negatively impact the final performance of the method as we show in Figure~\ref{fig:btsabl}. On the other hand, too large values of $\tau$ can cause the target network to lag behind the main solution, thus bottlenecking the training. The optimal $\tau$ in this problem defines a smoothing window of size $1/(1-\tau)=1000$ in the Poisson problem, which seems small enough for a training duration of $200 k$ epochs. On the other hand, the optimal smoothing window is much smaller in the Maxwell-Ampere problem, and the delayed target method benefits from more frequent changes to the target parameters in this problem.

Next, we studied the effect of target regularization weight $\lambda$ in Algorithm~\ref{alg:brs} of the main paper. A small target weight causes this method to diverge in this particular problem, as we've shown in Figure~\ref{fig:divergedbstrap} of the main paper. On the other hand, a regularization weight too large can slow down the training, as the main model remains too constrained to the target model during training.

We also show the effect of various target weight $M$ values in this problem. Ideally, $M\rightarrow \infty$ to make our approximations more accurate. A small target weight can effectively cause the method to seek biased solutions. On the other hand, setting $M$ too large may be impractical and instead cause the loss estimator's variance to explode as discussed in Equation~\eqref{eq:targetvar}. For this, $M$ must be set in conjunction with the $\lambda$ hyper-parameter in such challenging problems. 

\paragraph{Illustration of failure modes:} The delayed target method is more temperamental than the standard training; the set of delayed-target hyper-parameters, such as $\lambda$ and $\tau$, can have a significant impact on the solution quality. With poor hyper-parameters, the delayed target may poorly track the main solution, and the method can certainly diverge under an inappropriate set of hyper-parameters as we show in Figure~\ref{fig:divergedbstrap} of the main paper. Figure~\ref{fig:btsabl} also details the impact of the hyper-parameters related to the delayed target method. Furthermore, Equation~\eqref{eq:excessvargen} indicates that the excess-variance problem can be less severe when the underlying true solution is smooth in $g$ (i.e., when the optimal solution $\theta^*$ has a small $\mathbb{V}_{P(x'|x)}[g_\theta(x')]$ variance). Therefore, when the $g_{\theta^*}$ landscape is nearly flat, we expect the standard training to perform as well as the proposed method.

\subsection{Delayed Target Sample Size Scaling}\label{sec:dtsampsizeabls}
Two sets of parameters $\theta$ and $\theta^{\trg}$ participate in forming the training loss for the delayed target method. With $N=1$, half of the evaluated terms back-propagate the gradients; only the terms parameterized by $\theta$ can contribute to the gradient. With larger $N$, the target values become less noisy, but only a single term can still back-propagate the gradient. This is in contrast to all the other methods (e.g., the standard, the double-sampling, the deterministic approaches), where all of the evaluated points backpropagate gradients to the main set of parameters. One may wonder if this is the most efficient use of the evaluated terms for gradient estimation.

To account for this, instead of solving for
\vspace{-2mm}\begin{equation}
	f_{\theta}(x) = \frac{1}{N}\sum_{i=1}^{N} g_{\theta}(x_i') + y(x),
\end{equation}
in Equation~\eqref{eq:aprxmseloss}, we can reformulate the main objective to evaluate $N'$ points using the main model as 
\begin{equation}\label{eq:multinmdlrel}
\frac{1}{N'}\sum_{j=1}^{N'}f_\theta(x_j) = \gea + y(x_1,\cdots,x_{N'}).
\end{equation}
This is particularly straightforward in Examples~\ref{ex:poisson} and~\ref{ex:maxwell}, where an arbitrary number of points can be assigned to either side of the equation. In general, a portion of the $\gea$ terms in Equation~\eqref{eq:typicalipde} can always join the $f_\theta$ terms on the other side of the equation to form a training loss. This generalization allows the delayed target method to back-propagate the main parameter gradients using more points.

Of course, the introduction of the $N'$ hyper-parameter begs more questions about properly setting up the delayed target method. For instance, one may wonder whether the specific choice of $N$ and $N'$ could impact this method when the total number of samples $N+N'$ is controlled. For instance, one scaling strategy could be to maintain $N'=1$ while increasing $N$. Similarly, one could keep $N=1$ while increasing $N'$. Another option is to scale both $N$ and $N'$ equally.

\begin{figure*}[t]
	\centering
	\includegraphics[width=0.999\linewidth]{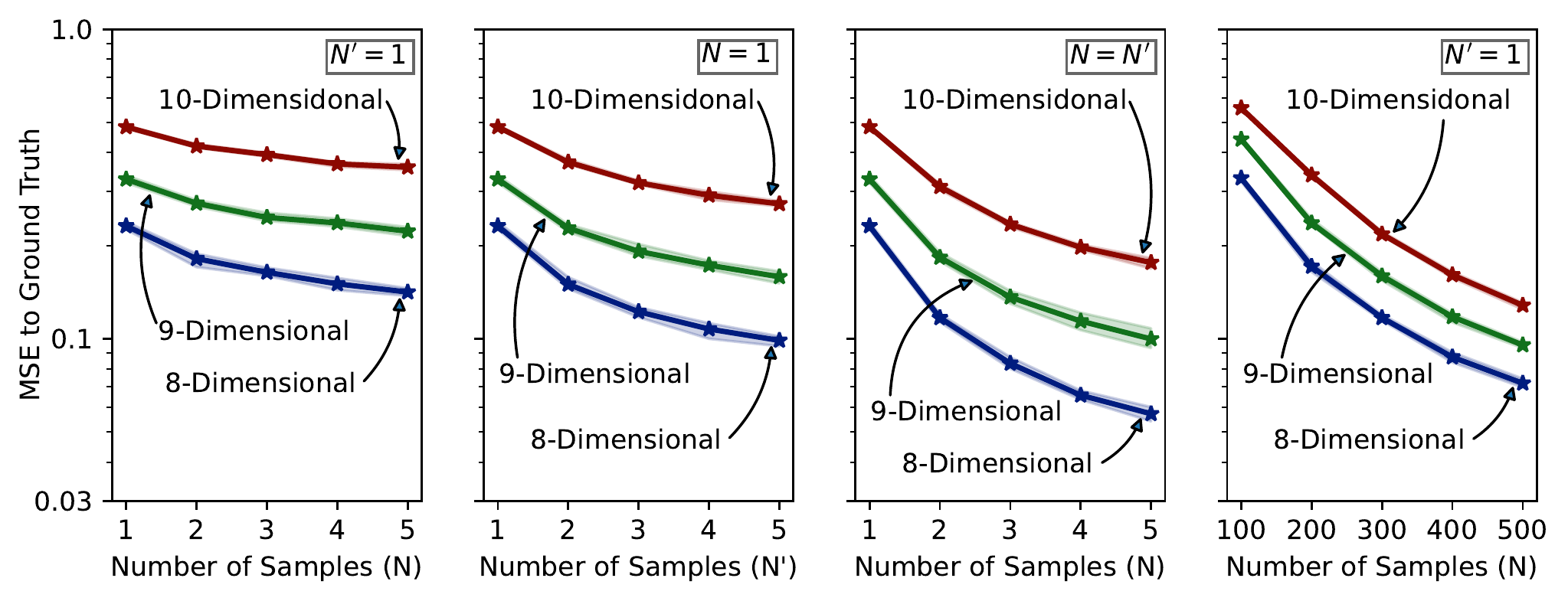}
	\caption{The effect of increasing the number of target samples on the solution quality in 8-, 9-, and 10-dimensional Poisson problems of Figure~\ref{fig:hidimpoiss} of the main paper. The three left subplots show various running configurations of the delayed target method with increased sample sizes. The leftmost plot shows the effect of increasing $N$ while keeping $N'=1$, whereas the second plot from the left shows the effect of increasing $N'$ while keeping $N=1$. The third plot from the left shows the effect of increasing both $N$ and $N'$ equally. Increasing $N'$ (i.e., the number of main model samples parameterized by $\theta$) seems to be more effective than increasing $N$ (i.e., the number of target model samples parameterized by $\theta^{\trg}$). Also, notice that all of the (1) $N=5$ and $N'=1$, (2) $N=1$ and $N'=5$, and (3) $N=N'=3$ configurations have the same $N+N'=6$ total value, yet, the last configuration yields the best performance. The rightmost plot shows the standard training method with 100 to 500 target samples.}
    \label{fig:msevsnstdbts}
\end{figure*}

Figure~\ref{fig:msevsnstdbts} shows the effect of increasing the $N$ and $N'$ sample sizes within the delayed target method. In particular, we see that having $N'=5$ and $N=1$ yields a better performance than having $N'=1$ with $N=5$. This is consistent with the theory that higher $N'$ values may lead to better gradient estimates to update the main model. However, having $N=N'=3$ seems to yield better results than both of the aforementioned approaches. This means that balancing the main and target terms may be the best decision when generalizing the delayed target method to larger sample sizes. 

Based on these results, we used $N=5$ and $N'=6$ to obtain the $N=10$ curve in Figure~\ref{fig:hidimpoiss} of the main paper. We abused the notation when labeling this curve to simplify the comparison between different methods (since the $N=10$ and $N'=1$ configuration has the same $N+N'$ total value). Similarly, we used $N=50$ and $N'=51$ for the delayed target curve labeled $N=100$ in the same figure.

\subsection{Learning Rate Ablation}\label{sec:lrablations}
\begin{figure*}[!t]
	\centering
	\includegraphics[width=0.35\linewidth]{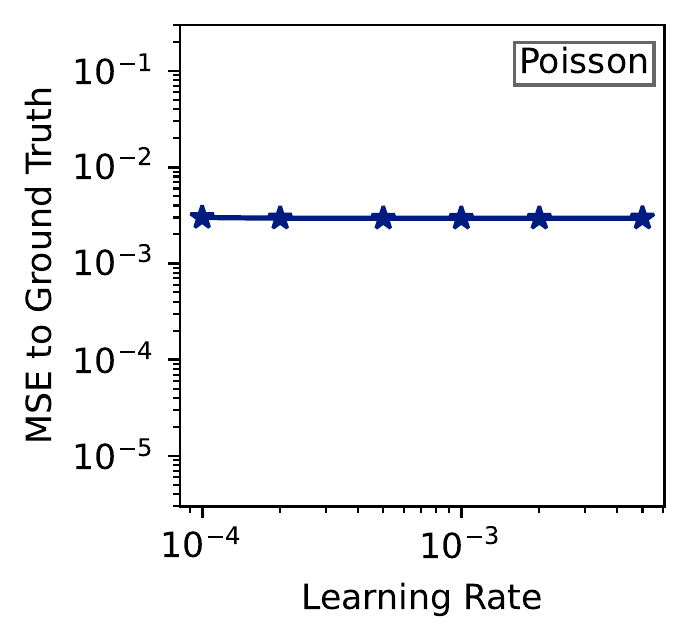}
	\includegraphics[width=0.32\linewidth]{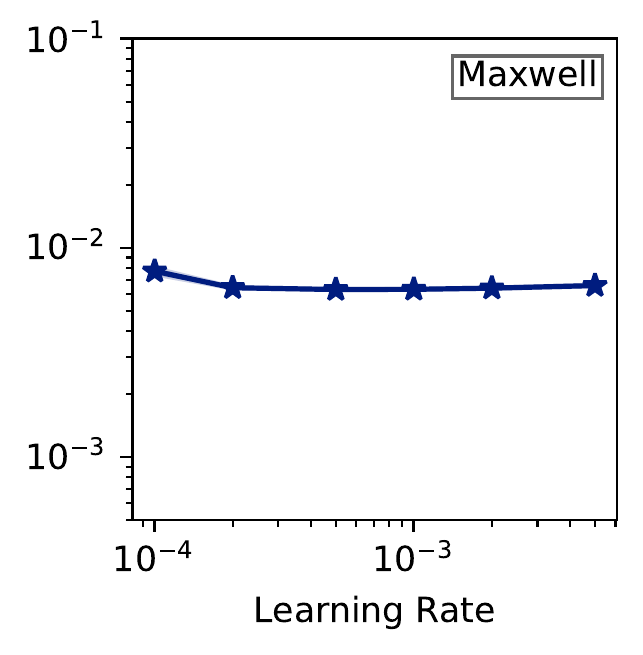}
	\includegraphics[width=0.32\linewidth]{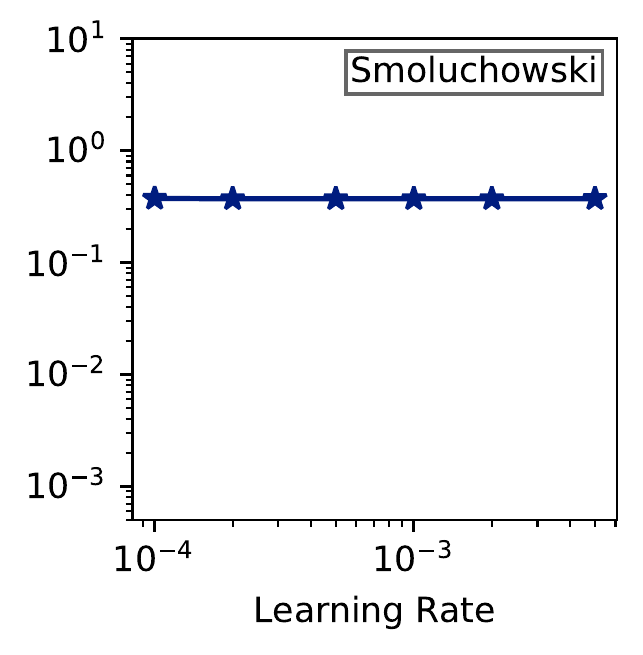}\\
	\vspace{-2mm}\includegraphics[width=0.35\linewidth]{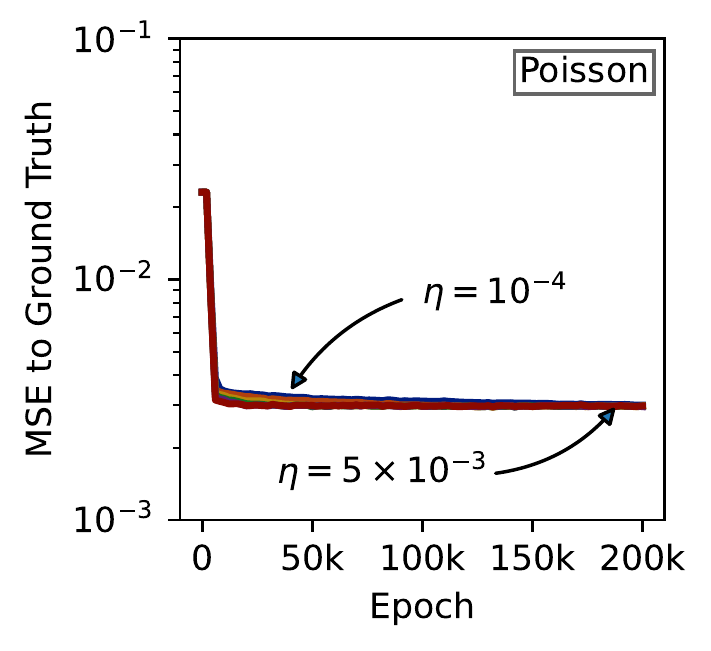}
	\includegraphics[width=0.32\linewidth]{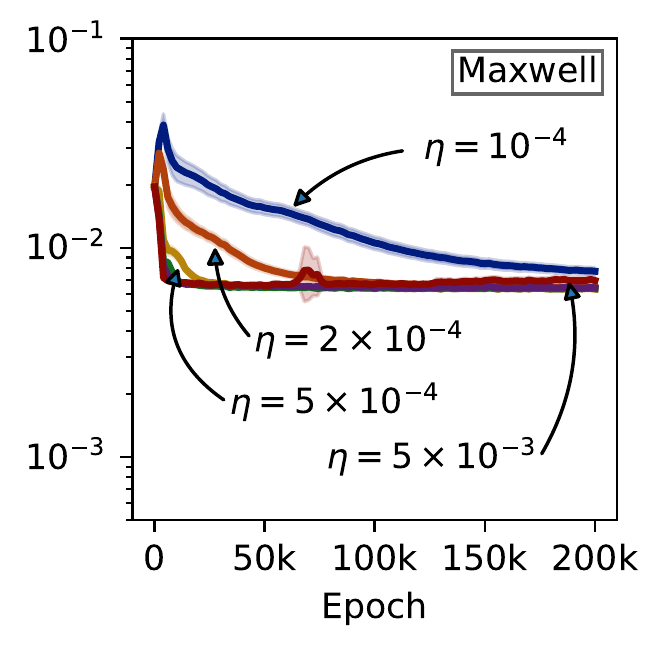}
	\includegraphics[width=0.32\linewidth]{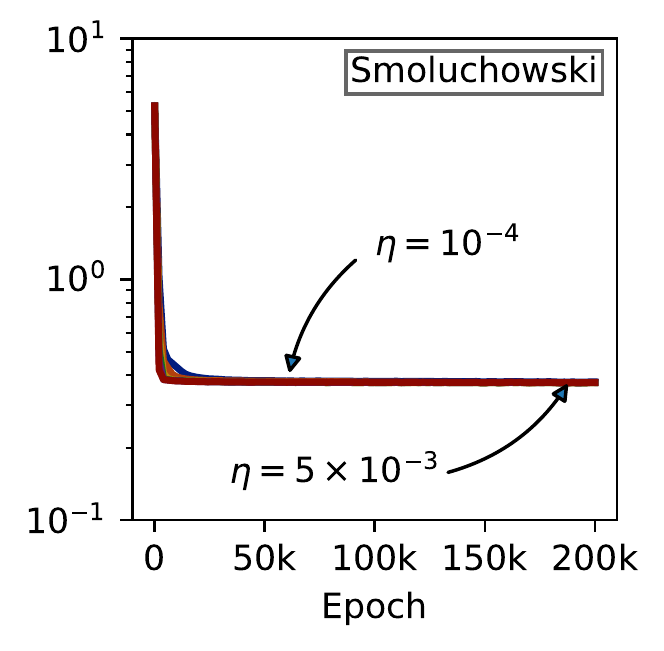}
	\vspace{-8mm}\caption{Ablating the effect of optimization learning rate on the standard training method with $N=1$. \textit{The top panel} shows the MSE to ground truth with respect to the learning rate. \textit{The bottom panel} shows the corresponding training curves to the top panel. \textit{The left, middle, and right columns} correspond to the 2D Poisson, 3D Maxwell-Ampere, and 2D Smoluchowski problems of Figures~\ref{fig:msegt},~\ref{fig:maxwellmain}, and~\ref{fig:smpanel} of the main paper, respectively.}
	\label{fig:lrabl}
\end{figure*}

Figure~\ref{fig:lrabl} shows the effect of the optimization learning rate on the standard method with $N=1$ in the Poisson, Maxwell-Ampere, and Smoluchowski problems. The poor performance of the standard method is consistent for a wide range of optimization learning rates. These results suggest that decaying the learning rate is not an effective solution to address the biased nature of the standard method's training objective.

\section{Implementation Details}\label{sec:pinnimplmntdtls}

In this section, we note the implementation details for each of the discussed problems. Sections~\ref{sec:poissdtls},~\ref{sec:maxwelldtls}, and~\ref{sec:smoldtls} describe the implementation details for the Poisson, Maxwell-Ampere, and Smoluchowski problems, respectively. Sections~\ref{sec:anlpois} and~\ref{sec:anlmaxwell} derive the analytical solutions for the Poisson and Maxwell-Ampere equations. Section~\ref{sec:rndmtchng} discusses the random effect matching process used throughout the experiments. Finally, Section~\ref{sec:pinnhpdtls} details the training hyper-parameters used in our numerical experiments.

\subsection{The Main Poisson Problems}\label{sec:poissdtls}
To solve this system in the integrated form, the standard method consists of fitting a neural model to the following loss:
\begin{equation}\label{eq:divthmloss}
\hat{\LL} = \mathbb{E}\bigg[\big(A_{d}^{r}\cdot \frac{1}{N}\sum_{i=1}^{N} E_\theta(x^{(i)})\cdot \hat{n}(x_i) - y_{\vol}\big)^2\bigg],
\end{equation}
where $A_{d}^{r}:=\int_{\sphr} 1\text{ d}S$ is the surface area of a $d$-dimensional ball with the $r$ radius, and the label is $
y_{\vol} := \iint_{\vol} \nabla\cdot \EPOIS\diff V$. The $x_i$ samples follow the $\text{Unif}\big(\sphr\big)$ distribution. The sampling intensity for volume $\vol$ defines the test constraint weights.

In Figures~\ref{fig:msegt},~\ref{fig:dtsdblpoisson}, and~\ref{fig:divergedbstrap} of the main paper, we consider a Poisson problem with $d=2$ dimensions and Dirac-delta charges. We place three unit charges at $[0,0]$, $[-0.5, -0.5]$, and $[0.5, 0.5]$ coordinates. For this setup, computing $y_{\vol}$ is as simple as summing the charges residing within the volume $\vol$. The integration volumes are defined as random spheres. The center coordinates and the radius of the spheres are sampled uniformly in the $[-1, 1]$ and $[0.1, 1.5]$ intervals, respectively. We train all models for 200,000 epochs, where each epoch samples 1000 points in total. We also study higher-dimensional problems with $d\in[2, 10]$ with a single charge at the origin in Figure~\ref{fig:hidimpoiss} of the main paper.


\newcommand{\aaa}{\frac{1}{\sqrt{3}}}
\newcommand{\bbb}{\frac{-1}{\sqrt{3}}}
\subsection{The Maxwell Problem with a Rectangular Current Circuit}\label{sec:maxwelldtls}
Our second example looks at finding the magnetic potentials and fields in a closed circuit with a constant current. This defines a singular $\J$ current density profile. We consider a rectangular closed circuit in the 3D space with the $[\aaa, \bbb, \bbb], [\aaa, \aaa, \aaa], [\bbb, \aaa, \aaa]$, and $[\bbb, \bbb, \bbb]$ vertices. The training volumes were defined as random circles, where the center coordinates and the surface normals were sampled from the unit ball, and the squared radii were sampled uniformly in the $[0.0, 1.0]$ interval.

\subsection{Smoluchowski Coagulation Problem}\label{sec:smoldtls}
To simulate particle evolution dynamics, we consider a Smoluchowski coagulation problem where particles evolve from an initial density. We considered the $x$ and $x'$ particle sizes to be in the $[0,1]$ unit interval, and the simulation time to be in the $[0,1]$ unit interval as well. We designed the $\SMK(x,x')$ coagulation kernel to induce non-trivial solutions in our unit solution intervals. Specifically, we defined $\SMK(x,x')=1.23 \times (\text{min}(1.14, \sqrt{x} + \sqrt{x'}))^3$. To find a reference solution, we performed Euler integration using exact time derivatives on a large grid size. The grid time derivatives were computed by evaluating the full summations in the Smoluchowski coagulation equation. 

\subsection{The Analytical Solution to the Poisson Problem}\label{sec:anlpois}
Consider the $d$-dimensional space $\mathbb{R}^{d}$ and the following charge:
\begin{equation}
\rhopois_{x} = \delta^d(x).
\end{equation}
For $d \neq 2$, the analytical solution to the $\EPOIS = \nabla \UPOIS$ and 
$\rhopois = \nabla \cdot \EPOIS$ system of Equations~\eqref{eq:poissys1} and~\eqref{eq:poissys2} can be derived as 
\begin{align}
\UPOIS(x) &= \frac{\Gamma(d/2)}{2\cdot\pi^{d/2}\cdot (2-d)} \|x\|^{2-d}, \\
\EPOIS(x) &= \frac{\Gamma(d/2)}{2\cdot \pi^{d/2}\cdot \|x\|^{d}} x.
\end{align}
For $d=2$, $E_{x}$ stays the same but we have
$\UPOIS({x}) = \frac{1}{2\pi} \ln(\|x\|)$.
To solve this system using the divergence theorem in Equation~\eqref{eq:divthm}, we can turn the integrals into scaled expectations. To find the appropriate scale, the $d-1$-dimensional surface area of a $d$-dimensional sphere with a radius of $r$ can be described as 
\begin{equation}
A_{d}^{r}:=\int_{\sphr} 1\text{ d}S = \frac{2\cdot \pi^{d/2}}{\Gamma(d/2)}\cdot r^{d-1}.
\end{equation}

\subsection{The Analytical Solution to The Maxwell-Ampere Equation}\label{sec:anlmaxwell}
Consider the $3$-dimensional space $\mathbb{R}^{3}$, and the following current along the z-axis:
\begin{equation}
\J(x) = \I \cdot \delta^2(x_1=0, x_2=0, x_3\in [z_1, z_2]).
\end{equation}
The analytical solution to the $\nabla\times \A = \B$ and $\nabla \times \B=\J$ system described in Example~\ref{ex:maxwell} can be expressed as
\begin{equation}
\A = \frac{-\I}{4\pi} \cdot \log\bigg(\frac{(z_2-x_3) + 
\sqrt{x_1^2 + x_2^2 + (z_2-x_3)^2}}{(z_1-x_3) + 
\sqrt{x_1^2 + x_2^2 + (z_1 - x_3)^2}}\bigg) \begin{bmatrix}0, 0, 1\end{bmatrix}\TRNSPS,
\end{equation}
and
\begin{equation}
\B= \frac{-\I}{4\pi\cdot \sqrt{x_1^2+x_2^2}} \cdot 
\bigg(\frac{z_2-x_3}{\sqrt{x_1^2 + x_2^2 + (z_2-x_3)^2}} - 
\frac{z_1-x_3}{\sqrt{x_1^2 + x_2^2 + (z_1-x_3)^2}}\bigg) \cdot 
\begin{bmatrix}\frac{-x_2}{\sqrt{x_1^2 + x_2^2}}\\
\frac{x_1}{\sqrt{x_1^2 + x_2^2}}\\0\end{bmatrix}.
\end{equation}

\subsection{Random Effect Matching}\label{sec:rndmtchng}
Random effects (random number generators seed; batch ordering; parameter initialization; and so on) complicate the study by creating variance in the measured statistics.  We use a matching procedure (so that the baseline and the proposed models share the same values of all random effects) to control this variance. As long as one does not search for random effects that yield a desired outcome (we did not), this yields an unbiased estimate of the improvement. Each experiment is repeated 100 times to obtain confidence intervals. Note that (1) confidence intervals are small, and (2) experiments over many settings yield consistent results.

\subsection{Training Hyper-parameters}\label{sec:pinnhpdtls}
We employed 3-5 layer perceptrons as our deep neural network, using 64 hidden neural units in each layer, and either the SiLU or $\tanh$ activation functions. We trained our networks using the Adam~\citep{kingma2014adam} variant of the stochastic gradient descent algorithm under a learning rate of $0.001$. We afforded each method and configuration 1000 function evaluations for each epoch. Table~\ref{tab:hpsmry} provides a summary of these hyper-parameters along with the volume and surface point. In all figures where the ground truth MSE was plotted against a hyper-parameter, the epoch with the minimal MSE value for each method was picked to summarize the training curve (rather than the last epoch). All heatmaps show the average prediction of each method across different trainings and randomization seeds, except in Figure~\ref{fig:dtsdblpoisson} of the main paper, where we hand-picked a single representative training for the heatmap visualizations to better illustrate the behaviors of the deterministic and double-sampling models.

In the high-dimensional Poisson problems of Figure~\ref{fig:hidimpoiss} in the main paper, we used a 5-layer MLP with the SiLU activation for all methods. For the delayed target method, we set $\tau=0.996$ and $\lambda=4$, with $M=500$, $250$, and $100$ for the $N=1$, $10$, and $100$ curves, respectively. We used second-order quadrature with Smolyak sparse grids which determined $N$ for each problem dimension; Gaussian quadrature defined $N$ between $2$ and $189$, and Leja quadrature defined $N$ between $2$ and $54$ for various dimensions. For the QMC method, we used the additive recursion rule for generating quasi-random sequences. In the Maxwell problem, we used a 5-layer MLP with the $\tanh$ activation for all methods, and we set $\tau=0.75$, $\lambda=64$, and $M=1000$ for the delayed target method. In the Smoluchowski problems, we used a 3-layer MLP with the SiLU activation for all methods, and we set $\tau=0.99$ and $\lambda=1$ for the delayed target method. The detailed hyper-parameter configurations for each problem can be found in the paper's code base.

\subsection{Evaluation Profiles}\label{sec:hdpevalproto}
In high-dimensional problems, we found (1) the choice of the evaluation distribution, (2) the output pre-processing, and (3) the specific deterministic or stochastic performance estimator to be important for the results to be meaningful across different methods and problem dimensions.

Many evaluation distributions may seem reasonable for our unit-charge Poisson problems in Figure~\ref{fig:hidimpoiss} of the main paper. For instance, sampling points uniformly from (1) the $[-1, 1]^d$ cube,
(2) the unit ball centered at zero, (3) the normal distribution, and (4) the training volumes (i.e., the randomly sampled balls used to enforce the divergence theorem) all seem like reasonable choices. While all choices may yield similar results on low-dimensional problems, we found this choice to be more influential in higher-dimensional problems. In particular, sampling points uniformly from the training volumes yields the most meaningful proxy for comparing method performances as it follows the training distribution closely.

The second factor is the output pre-processing before computing the mean-squared error. Having a boundary condition may theoretically guarantee a unique solution to the PDE of interest. That being said, strong enforcement of a boundary condition can have a confounding effect on the evaluation dynamics; note that the boundary condition loss term can interact with the main loss and the induced excess variance in Equation~\eqref{eq:excessvargen}, which may emphasize or hide a method's vulnerability to the excess variances problem. Furthermore, the ground truth solutions' scale can vary between dimensions. For this reason, we avoided enforcing the boundary condition too strongly to allow each method to demonstrate its unconstrained behavior. Moreover, we normalized the model and the ground truth outputs before computing the mean-squared error between them, such that they both have a zero empirical mean and unit empirical variance. We noticed that the training curves for all methods in high-dimensional problems tend to be overly noisy in the absence of this normalization pre-processing. Practically, this normalization step can be implemented by having the models ``calibrated'' through proper output shifting and scaling before evaluation. Since these calibration statistics are only two scalars for each model, this calibration assumption is not an overly unrealistic setup.
\begin{algorithm}[t]
	\caption{The Robust Performance Estimator for the High-Dimensional Poisson Problem}
	\begin{algorithmic}[1]
		\label{alg:eval}
		\REQUIRE The number of radii quantiles $q$ and the number of spherical angles $s$. \\The resulting evaluation sample size will be $e:=qs$.
		\REQUIRE The space dimensionality $d$ and the auxilary sample size $t$. 
		\STATE Sample $t$ points from the training volumes in an i.i.d.\ manner, and denote their radii as $\{r_1, r_2, \cdots, r_t\}$.
		\STATE Find the $(\frac{1}{2q}, \frac{3}{2q}, \cdots, \frac{2q-1}{2q})$ quantiles of the $\{r_1, r_2, \cdots, r_t\}$ population and name them $\{\tilde{r}_1, \cdots, \tilde{r}_q\}$.
		\STATE Sample $z_1, z_2, \cdots, z_s$ from a $d$-dimensional normal distribution in an i.i.d.\ mannaer.
		\STATE Define the evaluation angles:
		\begin{equation}
			\tilde{z}_j := \frac{z_j}{\|z_j\|_2} \qquad \forall 1\leq j\leq s.
		\end{equation}
		\STATE Define the evaluation points grid:
		\begin{equation}
			E := \{\tilde{r}_i \cdot \tilde{z}_j | 1 \leq i \leq q,\; 1 \leq j \leq s \} = \{x_1, x_2, \cdots, x_e\}.
		\end{equation}
		Note the size of the evaluation set $e$ is the product of the $q$ and $s$ sample sizes.
		\STATE Evaluate the model and the ground truth solution on the evaluation set:
		\begin{align}
			\forall 1\leq k\leq e: \qquad a_k := f_{\theta}(x_k), \qquad b_k := f^{*}(x_k).
		\end{align}
		\STATE Normalize the model and ground truth solutions:
		\begin{align}
			\forall 1\leq k\leq e: \qquad \tilde{a}_k &:= \frac{a_k - \EE_{i\sim \text{Unif}[1, e]}[a_i]}{\sqrt{\VV_{i\sim \text{Unif}[1, e]}[a_i]}}, \qquad
			\tilde{b}_k := \frac{b_k - \EE_{i\sim \text{Unif}[1, e]}[b_i]}{\sqrt{\VV_{i\sim \text{Unif}[1, e]}[b_i]}}.
		\end{align}
		\STATE Return the MSE between the $\tilde{a}_k$ and $\tilde{b}_k$ values:
		\begin{align}
			\LL_{e} := \frac{1}{e}\sum_{k=1}^{e} (\tilde{a}_k - \tilde{b}_k)^2.
		\end{align}
	\end{algorithmic}
\end{algorithm}

\renewcommand{\arraystretch}{1.45}
\begin{table}[!h]
    \aboverulesep=0ex
    \belowrulesep=0ex
    \begin{tabular}{|p{0.305\textwidth}|p{0.145\textwidth}|}  
    \toprule
    Hyper-Parameter & Value \\
    \midrule\midrule
    Randomization Seeds & 100 \\
    \hline
    Learning Rate & 0.001 \\
    \hline
    Optimizer & Adam \\
    \hline
    Epoch Function Evaluations & 1000 \\
    \hline
    Training Epochs & 200000 \\
    \hline
    Network Depth & 3-5 Layers \\
    \hline
    Network Width & 64 \\
    \hline
    Network Activation & SiLU or $\tanh$ \\
    \bottomrule
    \end{tabular}  \hspace{0.5mm} \begin{tabular}{|p{0.295\textwidth}|p{0.13\textwidth}|}
    \toprule
    Hyper-Parameter & Value \\
    \midrule\midrule
    Problem Dimensions & 1, 2, and 3 \\
    \hline
    Initial Condition Weight & 1 \\
    \hline
    Initial Condition Time & 0 \\
    \hline
    Initial Condition Points & $\text{Unif}([0,1])$ \\
    \hline
    Time Distribution  & $\text{Unif}([0,1])$ \\
    \hline
    Particle Size Distribution & $\text{Unif}([0,1])$ \\
    \hline
    Ground Truth Integrator & Euler \\
    \hline
    Ground Truth Grid Size & 10000 \\
    \bottomrule
    \end{tabular}\\
    
    \vspace{3mm}
    
    \begin{tabular}{|p{0.305\textwidth}|p{0.145\textwidth}|}
    \toprule
    Hyper-Parameter & Value \\
    \midrule\midrule
    Problem Dimension & 2 \\
    \hline
    Number of Poisson Charges & 3 \\
    \hline
    Integration Volumes & Balls \\
    \hline
    Volume Center Distribution & $\text{Unif}([-1,1])$ \\
    \hline
    Volume Radius Distribution & $\text{Unif}([0.1,1.5])$ \\
    \bottomrule
    \end{tabular}  \hspace{0.5mm} \begin{tabular}{|p{0.295\textwidth}|p{0.13\textwidth}|}
    \toprule
    Hyper-Parameter & Value \\
    \midrule\midrule
    Problem Dimension & 3 \\
    \hline
    Wire Segments & 4 \\
    \hline
    Integration Volumes & 2D Disks \\
    \hline
    Volume Center Distribution & Unit Ball \\
    \hline
    Volume Area Distribution & $\text{Unif}([0,1])$ \\
    \bottomrule
    \end{tabular}  
    
  \caption{A summary of the problem-specific hyper-parameters. \textit{The top left table} represents the common settings used in all experiments. \textit{The top right, bottom left, and bottom right} tables correspond to the Smoluchowski, Poisson, and Maxwell problems, respectively. In the high-dimensional Poisson problems, we defined a single charge at the origin, and the integration volumes were balls where (1) their centers were uniformly distributed inside the unit ball, and (2) their volumes followed a uniform distribution between zero and the volume of a unit ball.}
  \label{tab:hpsmry}
\end{table}

Finally, a stochastic i.i.d.\ sampling procedure of the evaluation points may not yield the most accurate results. Our training volume distribution is rotation invariant; in a spherical coordinate system, the joint distribution of points can be factored into two independent radii and angles distributions:
\begin{equation}\label{eq:evalrphidecomp}
	P(x) = P_{r}(r_x) \cdot P_{\phi}(\phi_x),
\end{equation}
where $r_x=\|x\|_2$ and $\phi_x=\frac{x}{\|x\|_2}$ can express the $x$ evaluation point in a spherical coordinate system. In particular, we found that sampling the point radii and angles independently and forming a grid can yield a robust estimator. Algorithm~\ref{alg:eval} details this procedure. Notice that this process is only applicable to rotation-invariant distributions, that is, distributions where the radii and angles are independent random variables as described in Equation~\eqref{eq:evalrphidecomp}. Our training volumes satisfy this condition. Algorithm~\ref{alg:eval} is statistically consistent, meaning that with $s, q, t\rightarrow \infty$, the estimated $\LL_{e}$ is guaranteed to be accurate in the limit. While choosing a finite $q$ may result in a small bias, it allows our performance estimator to be highly robust to outliers. 

We set both of the $q$ and $s$ grid dimensions to be 500 in Algorithm~\ref{alg:eval}. This results in an evaluation sample size of $2.5\times 10^5$ points for each model. The auxiliary sample size $t$ was set to a large value of $10^4$. To reduce the randomization effects, we matched the random effects in Algorithm~\ref{alg:eval} for all methods; in other words, we used the same $\tilde{r}_i$ and $\tilde{z}_j$ for all evaluations. This allowed us to obtain a robust performance estimate across all methods, training iterations, and problem dimensions.

For the Maxwell-Ampere problem, we sampled the evaluation points in an i.i.d.\ manner from the training volumes. As for pre-processing, we only subtracted their empirical means from the model and ground truth solutions before computing the mean squared error between them (i.e., no scaling was performed). No pre-processing was performed for the Smoluchowski problems as solving this PDE requires the initial conditions to be strongly enforced.

\newcommand{\dimth}{d}
\section{Computational Requirements}
Assuming $\theta$ is $\dimth$-dimensional and we use a deep feed-forward perceptron network, the required resources to run the delayed target method are detailed in Table~\ref{tab:compreq}. As a result, the delayed target method requires the following total computational and dynamic memory requirements per iteration:
\begin{align}
C^{\text{DT}}_{N,\dimth} = (c_5+c_9) N \dimth + c_4 N + (c_3+c'_9 + c_{10}) \dimth + c_6 + c_7 + c_8,
\end{align}
\begin{align}
M^{\text{DT}}_{N,\dimth} = (d_5+d_9) N \dimth + d_4 N + d_3 \dimth + d_6 + d_7 + d_8 + d_{10}.
\end{align}
In comparison, the standard method skips steps 7, 8, and 10, and therefore requires
\begin{align}
C_{N,\dimth} = (c_5+c_9) N \dimth + c_4 N + (c_3 + c'_9) \dimth + c_6,
\end{align}
\begin{align}
M_{N,\dimth} = (d_5+d_9) N \dimth + d_4 N + d_3 \dimth + d_6.
\end{align}
This means that both the standard and delayed target methods have per-step computational cost and dynamic memory usage that is dominated by the same $O(N \dimth)$ terms. Overall we expect the costs to be similar, with up to a doubling of per-step cost for the delayed target method because of the need for two networks (the main and target networks).

\begin{table}[t]
    \centering
	\aboverulesep=0ex
    \belowrulesep=0ex
    \begin{tabular}{|p{0.05\textwidth}|p{0.17\textwidth}|p{0.185\textwidth}|}
	\toprule
        Line & Computational Cost & Memory Requirement \\
		\midrule\midrule
        4 & $c_3\cdot \dimth$ & $d_3\cdot \dimth$ \\ \hline
        5 & $c_4\cdot N$ & $d_4\cdot N$ \\ \hline
        6 & $c_5\cdot N\cdot \dimth$ & $d_5\cdot N\cdot \dimth$ \\ \hline
        7 & $c_6$ & $d_6$ \\
		\bottomrule
    \end{tabular} \hspace{0.5mm} \begin{tabular}{|p{0.05\textwidth}|p{0.17\textwidth}|p{0.185\textwidth}|}
	\toprule
        Line & Computational Cost & Memory Requirement \\
		\midrule\midrule
        8 & $c_7$ & $d_7$ \\ \hline
        9 & $c_8$ & $d_8$ \\ \hline
        10 & $c_9\cdot N\cdot \dimth+c'_9K$ & $d_9\cdot N\cdot \dimth$ \\ \hline
        11 & $c_{10}\dimth$ & $d_{10}$ \\
		\bottomrule
    \end{tabular}

	\caption{The computational requirements of running Algorithm~\ref{alg:brs} of the main paper. The left column denotes the corresponding line in the algorithmic description. The middle and the right columns describe the computational cost and dynamic memory requirements, respectively.}
  \label{tab:compreq}
\end{table}

\section{Related work}

\subsection{Scientific Applications of Integro-Differential PDEs}\label{sec:ipdeapps}

Integro-differential PDEs arise in many areas such as quantum physics~\citep{laskin2000fractional,laskin2002fractional,elgart2007mean,lieb1987chandrasekhar}, visco-elastic fluid dynamics~\citep{constantin2005euler,caffarelli2011regularity,caffarelli2010giorgi,caffarelli2010drift}, nuclear reaction physics~\citep{BERN1994751}, mathematical finance~\citep{nolan1999fitting,ros2014integro}, ecology~\citep{humphries2010environmental,cabre2013influence,reynolds2009levy,viswanathan1996levy}, elasticity and material modeling~\citep{toland1997peierls,lu2005peierls}, particle system evolutions~\citep{chapman1996mean,weinan1994dynamics,giacomin1997phase,carrillo2011global}, aerosol modeling~\citep{wang2022learning}, computed tomography~\citep{wei2019regional}, radiation transfer and wave propagation~\citep{modest2021radiative}, grazing systems and epidemealogy~\citep{lakshmikantham1995theory}, and in the formulation of weak solutions with methods such as variational PINNs~\citep{kharazmi2021hp}. Weak solutions using the divergence or the curl theorems, or the Smoluchowski coagulation equation~\citep{wang2022learning} are a few representative forms we consider as examples for learning from integral losses.

\subsection{Physics-Informed Networks}\label{sec:pinnintro}

The original Physics-Informed Neural Network (PINN) was introduced in~\citet{raissi2019physics}. Later, variational PINNs were introduced in~\citet{kharazmi2019variational}. Variational PINNs introduced the notion of weak solutions using test functions into the original PINNs. This was later followed by hp-VPINNs~\citep{kharazmi2021hp}. In fact, integral forms appear in both VPINNs and hp-VPINNs, and delayed target methods could be used synergistically with these variational models to improve them. The double-sampling trick was originally introduced in reinforcement learning literature~\citep{baird1995residual}. In the context of PINNs,~\citet{guo2022monte} used this technique to address the Monte-Carlo loss estimation problem in fractional PDEs.

Conservative PINNs (or cPINNs for short) were also proposed and used to solve physical systems with conservation laws~\citep{jagtap2020conservative} and~\citet{mao2020physics} examined the application of PINNs to high-speed flows. Many other papers attempted to scale and solve the fundamental problems with PINNs, for example, using domain decomposition techniques~\citep{shukla2021parallel,li2019d3m}, the causality views~\citep{wang2022respecting}, and neural operators~\citep{li2020neural}. Reducing the bias of the estimated training loss is a general topic in machine and reinforcement learning~\citep{sutton1988learning,ghaffari2022fslfirth,arazo2020pseudo}.


\subsection{Bootstrapping Neural Networks}\label{sec:btsnn}

In general, the delayed target strategies and bootstrapping neural models, such as the TD-learning method, have been looked at in multiple contexts such as reinforcement or semi-supervised learning. The TD-learning method is an early example of this family~\citep{sutton1984temporal} and it has been analyzed extensively in prior work~\citep{dayan1992convergence,tsitsiklis1997analysis,baird1995residual,li2008worst,schoknecht2003td}. Time and time again, TD-learning has proven preferable over the ordinary MSE loss minimization (known as the Bellman residual minimization)~\citep{saleh2019deterministic,fujimoto2022should,yin2022experimental,chen2021instrumental}. The deep Q-networks proposed a practical adaptation of this methodology~\citep{mnih2015human}, which has been complemented in the TD3 method~\citep{fujimoto2018addressing}. 

Another example is the recent trend of semi-supervised learning, where teacher-student frameworks result in accuracy improvements of classification models by pseudo-labelling unlabelled examples for training~\citep{hinton2015distilling,pham2021meta,arazo2020pseudo,lee2013pseudo}. While a small number of recent theoretical insights exist on why semi-supervised learning does not produce trivially incorrect solutions~\citep{tian2021understanding}, a wealth of theoretical literature analyzed the ability and shortcomings of TD-learning methods to solve such problems.

\section{Recommendations and Limitations}
From the numerical examples, we consistently see that the delayed taråget method shows superior performance over the other methods. However, this also has limitations, as this method is more temperamental than the standard trainings, and may require careful specification of hyper-parameters such as $\lambda$ and $\tau$, as we showed in Figure~\ref{fig:divergedbstrap} of the main paper. Furthermore, we used large mini-batch sizes in conjunction with small $N$ values. While stochastic gradient descent can often be more effective with small mini-batch sizes, adaptively tuning the target smoothing and regularization weights to always stabilize the main and target parameters in the delayed target method under small mini-batch sizes and highly stochastic problems is another direction for future work. 

Our work solves the problem of learning from integral losses in physics-informed networks. We mostly considered singular and high-variance problems for benchmarking our methods. However, problems with integral losses can have broader applications in solving systems with incomplete observations and limited dataset sizes. This was beyond the scope of our work. Such applications may extend beyond the area of scientific learning and cover diverse applications within machine learning. We rigorously studied the utility of three methods for solving such systems. However, more algorithmic advances may be necessary to make the proposed methods robust and adaptive to the choice of algorithmic and problem-defining hyper-parameters. The delayed target method was shown to be capable of solving challenging problems through its approximate dynamic programming nature. However, we did not provide a systematic approach for identifying bottlenecks in case of failed trainings. Understanding the pathology of the studied methods is certainly a worthwhile future endeavor.

\end{document}